\documentclass{yresearch}
\usepackage{amsmath}
\usepackage{enumerate} 
\usepackage{algorithm}
\usepackage{algpseudocode}
\usepackage{amsfonts}
\usepackage{amsthm}
\newtheorem{theorem}{Theorem}
\newtheorem{lemma}[theorem]{Lemma}
\newtheorem{corollary}[theorem]{Corollary}
\usepackage{newtxtt}
\usepackage{cleveref}
\usepackage{diagbox} 
\usepackage{colortbl}
\usepackage{amssymb}
\usepackage{xspace}
\usepackage{wrapfig}
\usepackage{adjustbox}
\usepackage{tabularx}
\usepackage{booktabs}
\usepackage{mathtools}
\usepackage{tikz}
\usepackage{enumitem}
\usepackage{silence}
\usepackage{dsfont}
\usepackage[table]{xcolor}
\usepackage[dvipsnames]{xcolor}
\usepackage{multirow}
\usepackage{makecell}
\usepackage{xfakebold}

\usepackage{amsmath,amsfonts,bm}

\def\eqref#1{equation~\ref{#1}}

\def\1{\bm{1}}

\DeclareMathAlphabet{\mathsfit}{\encodingdefault}{\sfdefault}{m}{sl}
\SetMathAlphabet{\mathsfit}{bold}{\encodingdefault}{\sfdefault}{bx}{n}

\usepackage{multirow}
\usepackage[normalem]{ulem}
\usepackage{pifont}
\usepackage{placeins}
\usepackage{float}

\useunder{\uline}{\ul}{}
\definecolor{AppleWhite}{RGB}{255,255,255}
\definecolor{ApplePrimaryCoolGray}{RGB}{116,128,139}
\definecolor{AppleCoolGray1}{RGB}{199,209,214}
\definecolor{AppleCoolGray2}{RGB}{147,174,190}
\definecolor{AppleCoolGray3}{RGB}{124,147,160}
\definecolor{AppleCoolGray4}{RGB}{92,102,109}
\definecolor{AppleCoolGray5}{RGB}{78,93,100}
\definecolor{AppleCoolGray6}{RGB}{53,60,65}
\definecolor{AppleBlack}{RGB}{0,0,0}
\definecolor{AppleSecondaryChartGray}{RGB}{168,168,168}
\definecolor{AppleChartGray2}{RGB}{233,233,233}
\definecolor{AppleChartGray3}{RGB}{211,211,211}
\definecolor{AppleChartGray4}{RGB}{190,190,190}
\definecolor{AppleChartGray5}{RGB}{140,140,140}
\definecolor{AppleChartGray6}{RGB}{102,102,102}
\definecolor{AppleChartGray7}{RGB}{64,64,64}
\definecolor{ApplePrimaryChartBlue}{RGB}{84,151,193}
\definecolor{AppleBlue2}{RGB}{212,229,239}
\definecolor{AppleBlue3}{RGB}{169,202,223}
\definecolor{AppleBlue4}{RGB}{127,177,209}
\definecolor{AppleBlue5}{RGB}{71,130,166}
\definecolor{AppleBlue6}{RGB}{55,99,128}
\definecolor{AppleBlue7}{RGB}{45,72,89}
\definecolor{ApplePrimaryChartGreen}{RGB}{83,172,121}
\definecolor{AppleGreen2}{RGB}{212,234,221}
\definecolor{AppleGreen3}{RGB}{169,213,188}
\definecolor{AppleGreen4}{RGB}{126,193,155}
\definecolor{AppleGreen5}{RGB}{58,140,82}
\definecolor{AppleGreen6}{RGB}{39,102,54}
\definecolor{AppleGreen7}{RGB}{29,58,31}
\definecolor{ApplePrimaryChartYellow}{RGB}{253,195,93}
\definecolor{AppleYellow2}{RGB}{254,240,214}
\definecolor{AppleYellow3}{RGB}{254,224,174}
\definecolor{AppleYellow4}{RGB}{254,210,134}
\definecolor{AppleYellow5}{RGB}{230,168,69}
\definecolor{AppleYellow6}{RGB}{191,131,46}
\definecolor{AppleYellow7}{RGB}{153,107,54}
\definecolor{ApplePrimaryChartOrange}{RGB}{250,151,92}
\definecolor{AppleOrange2}{RGB}{254,229,214}
\definecolor{AppleOrange3}{RGB}{252,203,173}
\definecolor{AppleOrange4}{RGB}{252,178,133}
\definecolor{AppleOrange5}{RGB}{227,121,68}
\definecolor{AppleOrange6}{RGB}{191,87,46}
\definecolor{AppleOrange7}{RGB}{143,59,36}
\definecolor{ApplePrimaryChartRed}{RGB}{227,94,105}
\definecolor{AppleRed2}{RGB}{248,215,217}
\definecolor{AppleRed3}{RGB}{241,174,180}
\definecolor{AppleRed4}{RGB}{234,135,143}
\definecolor{AppleRed5}{RGB}{196,63,77}
\definecolor{AppleRed6}{RGB}{153,35,53}
\definecolor{AppleRed7}{RGB}{102,19,43}
\definecolor{ApplePrimaryChartPurple}{RGB}{161,150,204}
\definecolor{ApplePurple2}{RGB}{231,228,242}
\definecolor{ApplePurple3}{RGB}{208,202,229}
\definecolor{ApplePurple4}{RGB}{185,176,217}
\definecolor{ApplePurple5}{RGB}{128,113,171}
\definecolor{ApplePurple6}{RGB}{89,76,128}
\definecolor{ApplePurple7}{RGB}{62,46,101}
\definecolor{AppleCoolGray}{RGB}{116,128,139}
\definecolor{AppleChartGray}{RGB}{168,168,168}
\definecolor{AppleBlue}{RGB}{84,151,193}
\definecolor{AppleGreen}{RGB}{83,172,121}
\definecolor{AppleYellow}{RGB}{253,195,93}
\definecolor{AppleOrange}{RGB}{250,151,92}
\definecolor{AppleRed}{RGB}{227,94,105}
\definecolor{ApplePurple}{RGB}{161,150,204}
\usepackage[table]{xcolor}
\usepackage{wrapfig}

\newcommand{\myparagraph}[1]{\vspace{1pt}\noindent{\bf{#1}}~~}
\newcommand{\titlefont}{\fontsize{19}{20}\selectfont}
\definecolor{myblue}{RGB}{0, 112, 192}
\definecolor{myred}{RGB}{192, 0, 0}

\definecolor{textgray}{HTML}{6E6E73}
\usetikzlibrary{positioning, calc}
\usetikzlibrary{decorations.pathmorphing}

\makeatletter
\patchcmd{\wrong@fontshape}{\@gobbletwo}{}{}{}
\makeatother
\tcbuselibrary{minted}
\setminted[python]{
  linenos,
  breaklines,
  fontsize=\footnotesize,
  xleftmargin=2em
}

\newcommand{\eg}{e.g.\@\xspace}

\newcommand{\etal}{et al.\@\xspace}

\algrenewcommand{\algorithmiccomment}[1]{\hfill{\color{gray}\textit{// #1}}}

\definecolor{eccvblue}{RGB}{0,112,192}

\newcommand{\appitem}[1]{%
  \begingroup
  \hypersetup{linkcolor=eccvblue}%
  \noindent
  \makebox[\linewidth][l]{%
    \makebox[1.7em][l]{\textbf{\ref{#1}}}%
    \textbf{\nameref{#1}}%
    \dotfill%
    \pageref{#1}%
  }\par\vspace{3pt}%
  \endgroup
}

\newcommand{\appsubitem}[1]{%
  \begingroup
  \hypersetup{linkcolor=eccvblue}%
  \noindent
  \makebox[\linewidth][l]{%
    \hspace*{1.2em}%
    \makebox[2.5em][l]{\ref{#1}}%
    \nameref{#1}%
    \dotfill%
    \pageref{#1}%
  }\par\vspace{3pt}%
  \endgroup
}

\makeatletter
\AtBeginDocument{
  \urlstyle{sf}
  
}
\makeatother

\definecolor{light}{RGB}{125, 125, 125}
\crefname{tcb@cnt@pbox}{code}{code}
\Crefname{tcb@cnt@pbox}{Code}{Code}
\crefname{assumption}{assumption}{assumption}
\Crefname{assumption}{Assumption}{Assumptions}

\newtcolorbox[auto counter]{pbox}[2][]{
  colback=white,
  title=Code~\thetcbcounter: #2,
  #1,fonttitle=\sffamily,
  fontupper=\sffamily,
  arc=2pt,
  colframe=bgcolor,
  coltitle=fgcolor,
  colbacktitle=bgcolor,
  toptitle=0.25cm,
  bottomtitle=0.125cm
}

\makeatletter
\newcommand\applefootnote[1]{%
  \begingroup
  \renewcommand\thefootnote{}%
  \renewcommand\@makefntext[1]{\noindent##1}%
  \footnote{#1}%
  \addtocounter{footnote}{-1}%
  \endgroup
}
\makeatother

\definecolor{cverbbg}{gray}{0.90}

\usepackage[stable]{footmisc}

\title{\textit{Scale Where It Matters}: Training-Free Localized Scaling for Diffusion Models}
\author[1]{Qin Ren}
\author[2,6]{Yufei Wang}
\author[3]{Lanqing Guo}
\author[4]{Wen Zhang}
\author[5]{Zhiwen Fan}
\author[1]{Chenyu You}

\affiliation[1]{Stony Brook University}
\affiliation[2]{Nanyang Technological University}
\affiliation[3]{University of Texas at Austin}
\affiliation[4]{Johns Hopkins University}
\affiliation[5]{Texas A\&M University}
\affiliation[6]{SparcAI Research}

\metadata[Github]{\url{https://github.com/Y-Research-SBU/LoTTS}}
\metadata[Correspondence]{\sffamily Chenyu You: \url{chenyu.you@stonybrook.edu}}

\date{\sffamily\today}
\abstract{
Diffusion models have become the dominant paradigm in text-to-image generation, and test-time scaling (TTS) improves sample quality by allocating additional computation at inference. Existing TTS methods, however, resample the {entire} image, while generation quality is often spatially heterogeneous. This leads to unnecessary computation on regions that are already correct, and localized defects remain insufficiently corrected. In this paper, we explore a new direction -- {Localized TTS} -- that adaptively resamples defective regions while preserving high-quality regions, thereby substantially reducing the search space. This raises two challenges: {accurately localizing defects} and {maintaining global consistency}. We propose \textbf{LoTTS}, the first fully training-free framework for localized TTS. For defect localization, LoTTS contrasts cross-/self-attention signals under quality-aware prompts (e.g., ``high-quality'' vs.\ ``low-quality'') to identify defective regions, and then refines them into coherent masks. For consistency, LoTTS perturbs only defective regions and denoises them locally, ensuring that corrections remain confined while the rest of the image remains undisturbed. Extensive experiments on SD2.1, SDXL, and FLUX demonstrate that LoTTS achieves state-of-the-art performance:~it consistently improves both local quality and global fidelity, while reducing GPU cost by $2$--$4\times$ compared to Best-of-$N$ sampling. These findings establish localized TTS as a promising new direction for scaling diffusion models at inference time.
}

\begin{document}
\maketitle

\begingroup

\endgroup
\begin{figure}[t]
\centering
\includegraphics[width=0.98\linewidth]{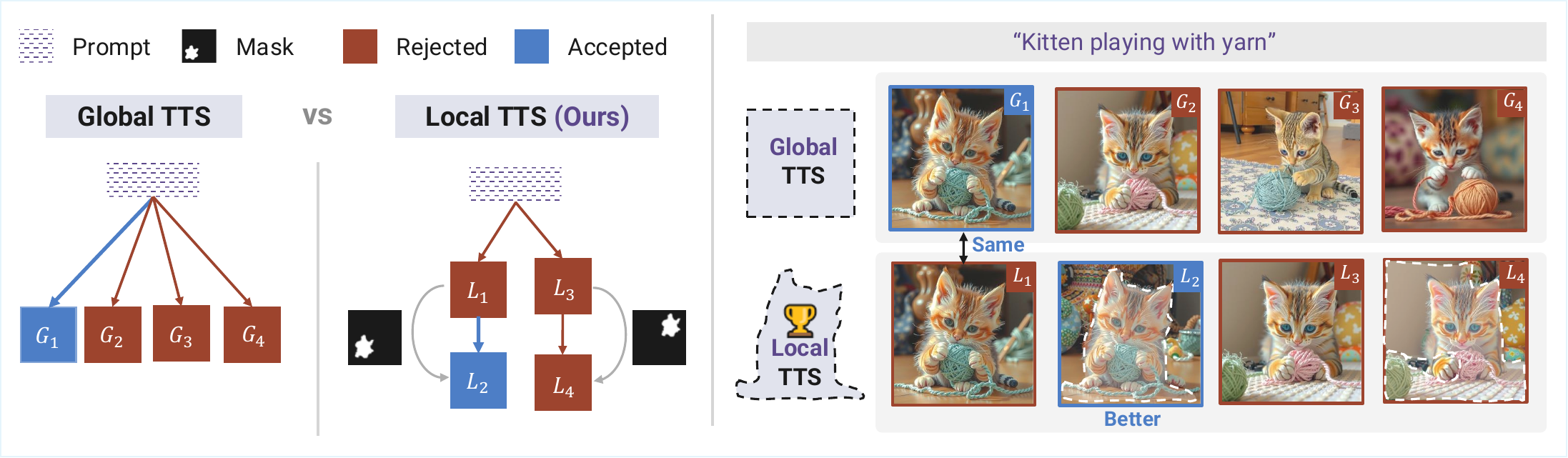}
\vspace{-2mm}
\caption{\textbf{Global vs.\ Local Test-Time Scaling.} Conventional TTS methods perform global search, sampling or perturbing the entire image, which ignores the inherent spatial heterogeneity of image quality and wastes computation on regions that are already good. LoTTS instead performs \emph{localized} refinement: it identifies defective regions using quality-aware masks and selectively resamples only where needed, improving low-quality areas while preserving high-quality content.}
\vspace{-2mm}
\label{fig:main}
\end{figure}

\section{Introduction}\label{sec:intro}

Diffusion models have become the de-facto standard for high-quality image generation, owing to their strong scalability with data, model size, and compute. This scalability has driven remarkable advances in text-to-image synthesis \citep{ho2020denoising,saharia2022photorealistic,rombach2022high,ruiz2023dreambooth,sun2024coma,sun2025ouroboros}, establishing scaling laws as a guiding principle for building more capable models. While most prior work has focused on scaling at training time, recent studies show that test-time scaling (TTS), allocating additional compute during inference, can also significantly improve sample quality and overall performance~\citep{nichol2021glide,esser2024scaling,peebles2023scalable,ma2025scaling,liu2024correcting}. Despite its promise, existing TTS research remains limited in scope, leaving open questions about how to use inference-time compute more effectively.

Existing TTS methods can be broadly grouped into three categories. The first is \textit{denoising step scaling}, which improves quality by increasing the number of sampling steps \citep{song2020score,lu2022dpm}. However, these gains saturate quickly and plateau around 50 steps, with further increases offering negligible benefit. The second is \textit{Best-of-$N$ search}, which generates $N$ samples and selects the best one via a verifier \citep{liu2024correcting,wang2023exploring}. While simple, this brute-force approach treats each candidate as an independent sample from scratch, overlooking the fact that even imperfect images may be substantially improved through local corrections. As a result, potentially promising samples are discarded, and computation is wasted on redundant global search. The third is \textit{trajectory/noise search}, which perturbs the initial noise or explores alternative sampling paths~\citep{xu2023restart,ramesh2025test}. Although more fine-grained, searching over all regions can inadvertently disturb areas that are already of high quality, leading to inefficiency and instability.
Despite their differences, all three categories share a fundamental limitation: \textit{they operate at the full-image level}, as shown in Fig.~\ref{fig:main}. Consequently, they overlook the inherent spatial heterogeneity of image quality and fail to exploit the potential of localized refinement.

This observation naturally motivates an orthogonal direction: \textit{localized TTS}, where only defective regions are resampled while preserving high-quality content \citep{cao2025temporal}, thereby substantially reducing the search space. Yet turning this idea into practice introduces two key challenges. The first is \textit{accurate defect localization}: since the distribution of artifacts is complex and prompt-dependent, reliably identifying regions that truly require correction is non-trivial \citep{liu2024improved,zhang2023perceptual}. The second is \textit{maintaining consistency in local resampling}: as the sampling trajectory is globally defined, locally modifying only a subset of regions may introduce incompatibility, leading to semantic drift, stylistic inconsistency, or boundary artifacts that degrade perceptual quality \citep{song2020score,xu2023restart}.

Our key observation is that diffusion models already encode spatially resolved quality information in their cross-attention maps: by contrasting attention under quality-differentiated prompts (\eg ``high-quality'' vs.\ ``low-quality''), defective regions naturally stand out -- enabling zero-shot defect localization without any external supervision.
Building on this insight, we propose \textbf{LoTTS} ({Prompt-Guided \underline{\textbf{Lo}}calized \underline{\textbf{T}}est-\underline{\textbf{T}}ime \underline{\textbf{S}}caling}), the first fully training-free localized TTS framework that addresses both challenges in a unified pipeline.
For \textit{defect localization}, LoTTS contrasts cross-attention maps under quality-differentiated prompts to produce a quality-aware mask that highlights artifact-prone regions, without relying on any external predictors or annotations.
For \textit{consistency maintenance}, LoTTS injects noise only into the masked regions at an intermediate timestep and performs localized denoising~\citep{meng2021sdedit}, followed by a global harmonization pass to ensure seamless integration. The framework is fully plug-and-play and applies to both diffusion- and flow-based models~\citep{song2020score,lu2022dpm}.
Our main contributions are summarized as follows:
\begin{itemize}[leftmargin=*]
    \item We propose {LoTTS}, the first training-free localized Test-Time Scaling framework. Its core is a prompt-driven defect localization mechanism that contrasts cross-attention signals to produce quality-aware masks without any external supervision, paired with a consistency-preserving resampling strategy that refines only the masked regions while maintaining coherence with the rest of the image.
    \item We provide theoretical analysis showing that localized resampling achieves a higher expected quality gain than global resampling under stated conditions, formally justifying the localized TTS paradigm.
    \item We conduct extensive experiments on SD2.1, SDXL, and FLUX, showing that LoTTS consistently achieves state-of-the-art performance across multiple human-preference and automated evaluation metrics, while reducing GPU cost by $2$--$4\times$ relative to Best-of-$N$ sampling.  
\end{itemize}

\begin{figure*}[t]
\centering
\includegraphics[width=0.90\linewidth]{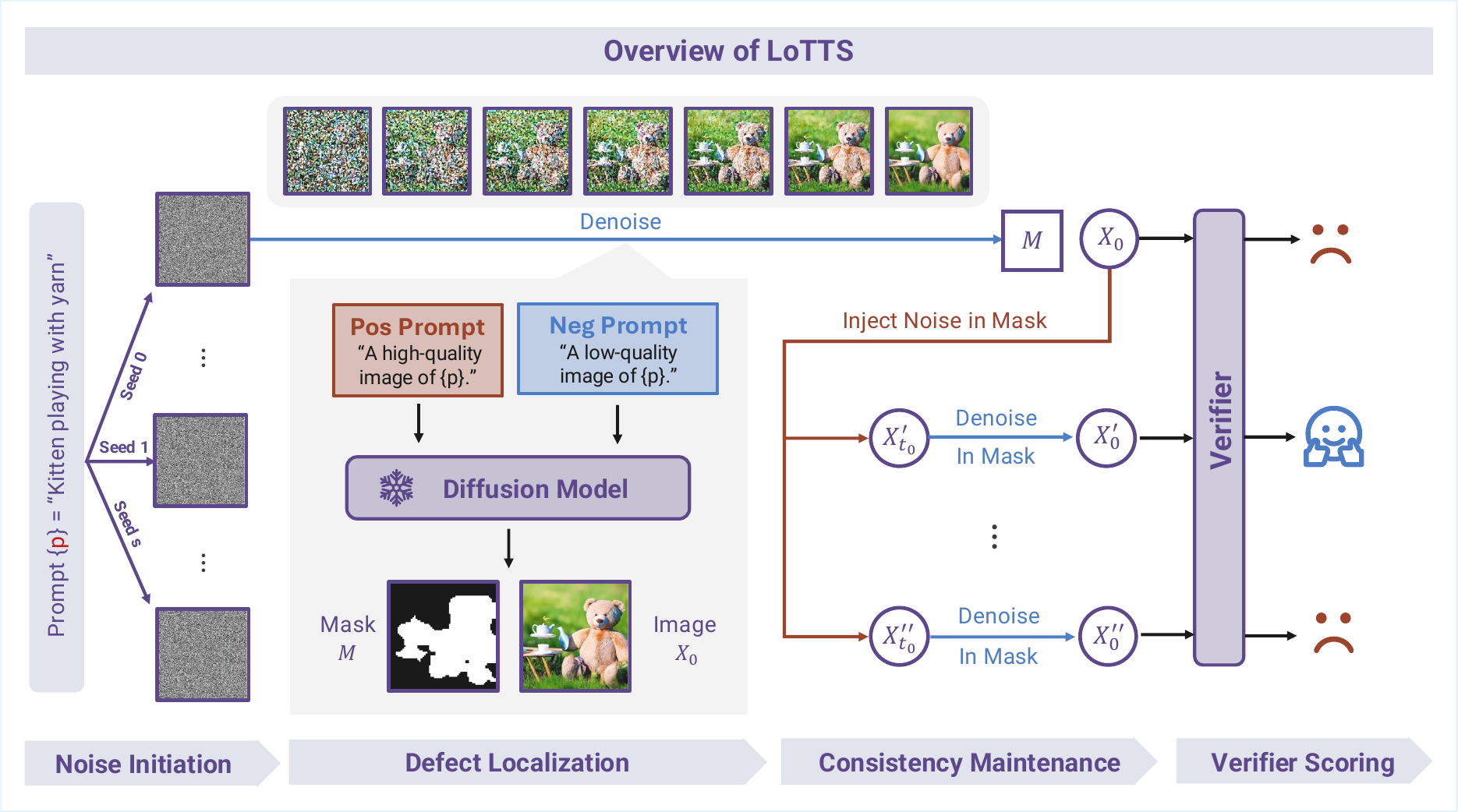}
\vspace{-2mm}
\caption{\textbf{Overview of LoTTS.}
Given a text prompt, LoTTS first generates candidate images from different noise seeds.
It then localizes defective regions using high-/low-quality prompt contrast and constructs a quality-aware mask.
Noise is injected only inside the masked regions, followed by localized denoising with spatial and temporal consistency.
A verifier finally selects the best refined sample.}
\vspace{-3mm}
\label{fig:overview}
\end{figure*}

\section{Related Work}

\myparagraph{Image Generation and Evaluation.} Image generation has advanced from GANs~\citep{goodfellow2020generative,karras2019style} and autoregressive models~\citep{van2016pixel,van2016conditional} to flow-based and diffusion models. GANs achieve high fidelity but are unstable, while autoregressive models capture dependencies yet remain slow. Early flows~\citep{kingma2018glow} provide exact likelihoods but scale poorly, whereas recent flow matching and rectified flow methods~\citep{lipman2022flow,albergo2023stochastic,kim2025inference} connect closely to diffusion and offer efficient alternatives. Diffusion models~\citep{ho2020denoising,rombach2022high} now dominate, delivering state-of-the-art text-to-image synthesis~\citep{rombach2022high,podell2023sdxl}. In parallel, evaluation methods have evolved: earlier work often collapsed quality into a single global score~\citep{wang2004image,salimans2016improved,heusel2017gans}, while spatial indicators relied on texture- or frequency-based heuristics~\citep{yu2019attributing,durall2020watch} or classifier- and VLM-based predictors~\citep{liu2021aligning,zhang2023perceptual,liu2024improved}, typically requiring supervision and external datasets. In contrast, LoTTS leverages the diffusion model's inherent attention to localize defects automatically, enabling training-free, localized refinement.

\myparagraph{Defect Localization.} 
A substantial body of work aims to localize defects in generated images. Early methods analyze GAN or super-resolution artifacts through spatial or frequency cues~\citep{yu2019attributing, durall2020watch,dzanic2020fourier,you2025uncovering}. More recent approaches train supervised models to predict perceptual-error or artifact heatmaps~\citep{zhang2023perceptual,li2024g,ren2025otsurv}, or pursue defect detection via uncertainty estimation~\citep{kou2023bayesdiff,liu2025together,zhang2025supervise} and diffusion-trajectory analysis~\citep{cao2025temporal}.
In contrast to methods that rely on external predictors, supervised training, or heavy post-hoc analysis, LoTTS derives defect localization directly from the diffusion model itself. Our masks are computed from attention responses to positive and negative prompts, leveraging the model's inherent cross-modal alignment. This makes LoTTS fully training-free and inherently consistent with the generative semantics of the T2I model.

\myparagraph{Generation Quality Enhancement.} Image editing in modern T2I systems is typically implemented as a post-processing stage. The cascade paradigm, consisting of a generator followed by a refiner, has become a standard design, as in IF~\citep{genIF,you2022megan}, SDXL~\citep{podell2023sdxl} and SD Cascade~\citep{pernias2023wurstchen}, where an additional pass refines the entire image to enhance global fidelity. However, such refiners usually require separate training and introduce non-negligible computational overhead. In contrast, SDEdit~\citep{meng2021sdedit} shows that diffusion models can perform localized edits via noise--denoise updates within user-specified masks, suggesting the potential of training-free localized refinement. This localized resampling mechanism inspires our approach, which extends SDEdit's manual editing to automated quality-aware refinement at test time.

\myparagraph{Test-Time Scaling in Vision.} Some test-time scaling methods have been proposed to enhance diffusion model generation by allocating more computation at inference. Early work simply increased denoising steps, but improvements saturate quickly beyond a certain number of function evaluations (NFE) \citep{karras2022elucidating,song2020denoising,song2020score}. Recent studies therefore explore alternative directions, such as \textit{Best-of-$N$ search}, where multiple candidates are generated from different noise seeds and a verifier selects the best one \citep{liu2024correcting,wang2023exploring}, searching over noise initializations \citep{song2020score,xu2023restart}, optimizing sampling trajectories \citep{ramesh2025test} with verifier feedback \citep{song2020denoising,karras2022elucidating,liu2022pseudo,lu2022dpm,salimans2022progressive,you2024calibrating}, or adopting evolutionary~\citep{he2025scaling} and tree-search methods \citep{yoon2025fast}. Unlike these approaches, which all operate on the \emph{entire} image and require full regeneration, our LoTTS performs localized TTS by concentrating on low-quality regions for greater efficiency.

\section{Preliminaries}
\myparagraph{Diffusion Models.}
Diffusion models transform a simple source distribution, \eg a standard Gaussian, into a target data distribution $p_0$. In diffusion models~\citep{sohl2015deep,ho2020denoising}, the forward process gradually corrupts clean data with noise, as:
\begin{equation}
\bm{x}_t = \alpha(t) \bm{x}_0 + \sigma(t) \epsilon, \quad \epsilon \sim \mathcal{N}(0,I),
\end{equation}
where $\alpha(t)$ and $\sigma(t)$ denote the noise schedule, and $t\in [0,T]$. To recover data from its diffused representation, diffusion models generally rely on Stochastic Differential Equation (SDE)-based sampling during inference~\citep{song2020score,song2020denoising}, which introduces stochasticity at every denoising step:

\begin{equation}
  \bm{x}_{t-1} = \alpha(t{-}1)\,\hat{\bm{x}}_0 + \sqrt{\sigma(t{-}1)^2-\tilde{\sigma}_t^2}\,\epsilon_\theta(\bm{x}_t,t) + \tilde{\sigma}_t\,\epsilon_t,
\end{equation}
where $\hat{\bm{x}}_0 = (\bm{x}_t - \sigma(t)\,\epsilon_\theta(\bm{x}_t,t))/{\alpha(t)}$ is the predicted clean image,
$\tilde{\sigma}_t$ controls the stochasticity of sampling,
and $\epsilon_t \sim \mathcal{N}(0,I)$.

\myparagraph{Flow Models.}
Flow models~\citep{lipman2022flow,albergo2023stochastic} parameterize the velocity field $u_t \in \mathbb{R}^d$ and generate samples by solving the Flow Ordinary Differential Equation (ODE)~\citep{song2020score} backward from $t=T$ to $t=0$: 
\begin{equation}
  \label{eq:flow-ode}
  d\bm{x}_t = u_t(\bm{x}_t)\,dt,
\end{equation}
where $u_t(\bm{x}_t)$ denotes the velocity field learned by the flow model, 
and $dt$ is an infinitesimal step along the reverse-time ODE. 
This deterministic dynamics evolves $\bm{x}_t$ continuously in time, 
producing identical outcomes for the same input and limiting the applicability of test-time scaling methods~\citep{kim2025inference}, 
which require stochasticity to explore diverse trajectories.
To address this limitation, recent studies propose that the deterministic Flow-ODE could be reformulated into an equivalent SDE~\citep{albergo2023stochastic,ma2024sit,patel2024exploring,kim2025inference,singh2024stochastic}. The resulting stochastic process can be written as:
\begin{equation}
d\bm{x}_t = \left(u_t(\bm{x}_t) - \tfrac{\sigma(t)^2}{2}\nabla \log p_t(\bm{x}_t)\right) dt + \sigma(t) d\bm{w},
\end{equation}
where the score function $\nabla \log p_t(\bm{x}_t)$ can be estimated from the velocity field $u_t$~\citep{singh2024stochastic}, and the Brownian motion term $d\bm{w}$ introduces stochasticity at each sampling step. This enables LoTTS to naturally apply to flow models.

\section{Method}\label{sec:method}

\subsection{Overview}

\myparagraph{Motivation and Challenges.}
Existing test-time scaling methods, such as Best-of-$N$ or trajectory search, apply uniform resampling to the entire image, overlooking spatial heterogeneity in quality and thereby wasting computation or even degrading well-formed regions. This motivates a more targeted form of TTS that selectively resamples only defective regions, reducing the search space and improving both efficiency and stability. However, realizing this paradigm raises two key challenges:
(1) \textit{Defect Localization}: how to generate reliable resample masks that identify low-quality regions, and  
(2) \textit{Consistency Maintenance}: how to perform localized resampling within these regions without disrupting the rest of the image.

\myparagraph{Overview of LoTTS.}
To address these challenges, we introduce \textbf{LoTTS} -- \textit{Localized Test-Time Scaling} -- a fully training-free framework that operationalizes localized resampling for diffusion models.
Figure~\ref{fig:overview} illustrates the LoTTS pipeline. 
Given a text prompt, LoTTS begins by sampling several noise seeds and generating multiple initial images via standard diffusion denoising. 
For each initial image, it detects artifact-prone regions using contrastive attention derived from high-quality and low-quality auxiliary prompts, producing a binary mask that highlights potential defects. 
LoTTS then injects noise exclusively within the masked regions at an intermediate timestep~$t_0$ and performs localized denoising to refine these areas while keeping the unmasked content unchanged, thereby maintaining global semantic coherence and avoiding boundary artifacts. 
All refined variants are subsequently evaluated by an external verifier, and the candidate with the highest score is chosen as the final output.

Pseudocode is given in Appendix~\ref{app:algo}, 
and Appendix~\ref{app:TheoreticalAnalysis} proves that 
localized resampling achieves a higher expected quality gain than global resampling under mild conditions.

\subsection{Defect Localization}

We leverage the intrinsic attention signals of diffusion models to localize defects: cross-attention with a ``low-quality'' prompt highlights artifact-prone regions, and contrasting it with a ``high-quality'' prompt makes those defects stand out. To obtain stable and semantically meaningful masks, we further propagate and reconcile these signals.

\begin{figure}[t]
\begin{center}
\includegraphics[width=0.98\linewidth]{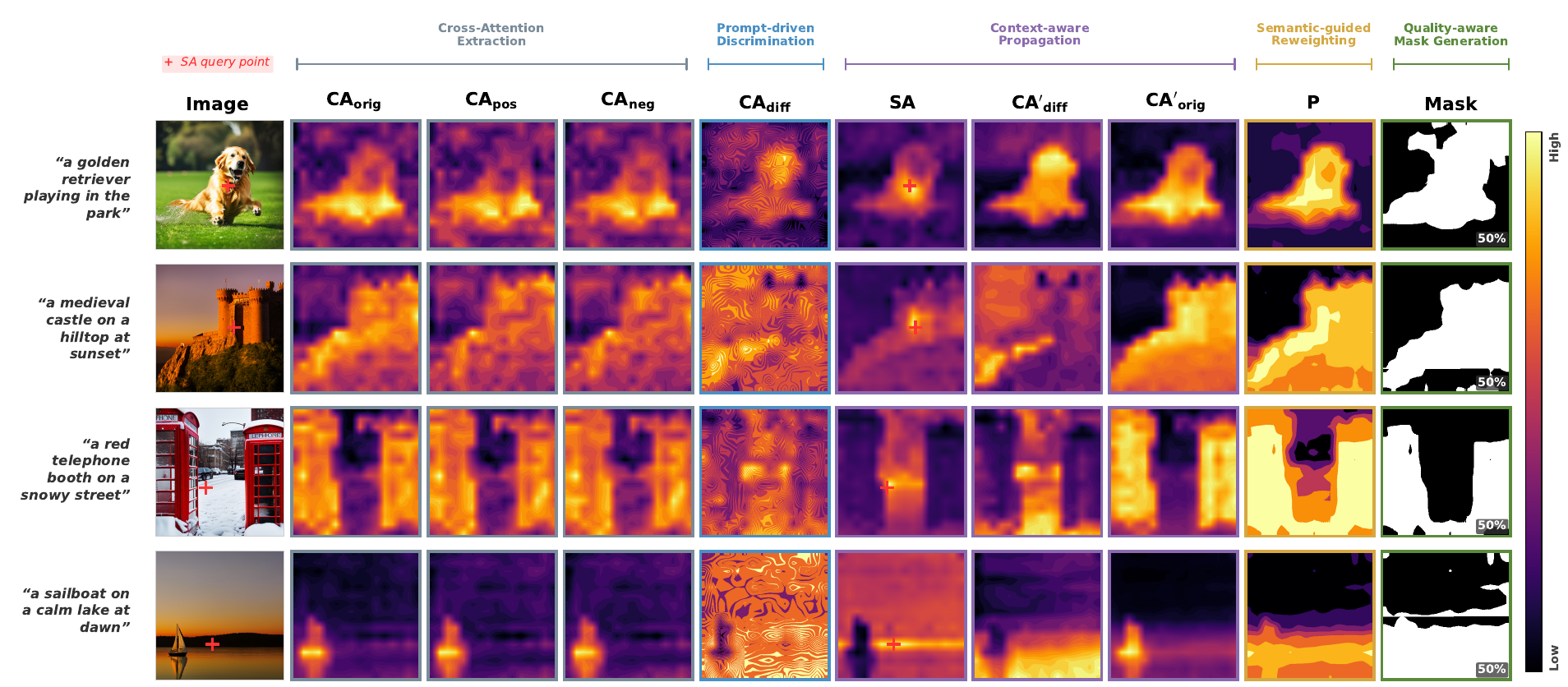}
\vspace{-0mm}
\caption{\textbf{Defect Localization Overview.}  LoTTS localizes defects from cross-attention maps through four stages:  prompt-driven discrimination, context-aware propagation,  semantic-guided reweighting, and quality-aware mask generation.}
\vspace{-3mm}
\label{fig:attn} 
\end{center}
\end{figure}

As illustrated in Fig.~\ref{fig:attn}, starting from extracted cross-attention maps, the process consists of four stages: 
(1) \textit{Prompt-driven Discrimination}: take the difference between high-/low-quality cross-attention maps to highlight candidate defect regions. 
(2) \textit{Context-aware Propagation}: refine these signals by propagating across spatially similar positions, mitigating noise, and enforcing local coherence. 
(3) \textit{Semantic-guided Reweighting}: combine with original prompt attention to suppress irrelevant background. 
(4) \textit{Quality-aware Mask Generation}: binarize by percentile to control the resampling ratio.

\myparagraph{Prompt-driven Discrimination.}
Cross-attention maps reflect how different prompts attend to image regions: 
a ``low-quality'' prompt tends to focus on artifact-prone areas, while a ``high-quality'' prompt attends to cleaner regions. 
By contrasting the two, we can highlight potential defects.  

Following the common setting~\citep{liu2024towards,helbling2025conceptattention}, we obtain a spatial cross-attention vector 
$\mathrm{CA} \in \mathbb{R}^S$ by averaging over tokens, heads, and selected layers, 
where $S = H_s \times W_s$ denotes the number of spatial positions. 
For a given prompt $p$, we construct three variants: a \textit{positive} prompt 
(``A high-quality image of \{$p$\}''), a \textit{negative} prompt 
(``A low-quality image of \{$p$\}''), and the \textit{origin} prompt (the original $p$). 
This yields three attention vectors 
$\mathrm{CA}_\mathrm{pos}, \mathrm{CA}_\mathrm{neg}, \mathrm{CA}_\mathrm{orig} \in \mathbb{R}^S$.  
We then define a contrastive cross-attention map as:
\begin{equation}
    \mathrm{CA}_{\mathrm{diff}} = 
    \mathrm{CA}_\mathrm{neg} - \mathrm{CA}_\mathrm{pos},
\end{equation}
which emphasizes spatial locations where the negative prompt receives higher attention than the positive prompt. 
The origin vector $\mathrm{CA}_\mathrm{orig}$ will later serve as a foreground prior in our aggregation step.

\myparagraph{Context-aware Propagation.}
While the contrastive attention map $\mathrm{CA}_{\mathrm{diff}}$ highlights defect-prone regions, 
it is often noisy and fragmented, with neighboring pixels showing inconsistent scores. 
Intuitively, spatially or semantically similar regions should share similar quality signals. 
To enforce such coherence, we propagate the attention scores using a self-attention matrix derived from the image queries $Q$:
\begin{equation}
    \mathrm{CA}' = \mathrm{SA} \times \mathrm{CA}, \qquad 
    \mathrm{SA} = \mathrm{Softmax}\!\left(\frac{Q Q^\top}{\sqrt{d}}\right).
\end{equation}
This operation smooths the raw attention map by diffusing scores across related spatial positions. 
Applying it to both $\mathrm{CA}_{\mathrm{diff}}$ and $\mathrm{CA}_{\mathrm{orig}}$ yields 
refined maps $\mathrm{CA}'_{\mathrm{diff}}$ and $\mathrm{CA}'_{\mathrm{orig}}$ 
that are more stable and spatially coherent.

\myparagraph{Semantic-guided Reweighting.}
Although the contrastive map $\mathrm{CA}'_{\mathrm{diff}}$ can reveal defect-prone areas, 
it often assigns high scores to background regions with little semantic content 
(\eg large sky areas). 
Since meaningful defects should lie within the semantic foreground, 
we combine the contrastive map with the original attention map 
$\mathrm{CA}'_{\mathrm{orig}}$, which serves as a soft foreground mask:
\begin{equation}
    P = \mathrm{CA}'_{\mathrm{diff}} + \lambda \,\mathrm{CA}'_{\mathrm{orig}},
    \label{eq:Semantic_guided_Reweighting}
\end{equation}
where $\lambda$ balances the defect signal and the foreground prior. 
The resulting map $P$ emphasizes semantically relevant regions, 
leading to more reliable defect localization.

\myparagraph{Quality-Aware Mask Generation.}
Finally, we convert the aggregated quality map into a binary mask 
by keeping the top $r$ proportion of spatial positions with the highest defect scores, 
ensuring that resampling is both targeted and controllable. 
Formally, the mask $M \in \{0,1\}^S$ is given by:
\begin{equation}
    M = \mathbb{I}\!\left(P > \mathrm{Perc}(P,\,1-r)\right),
\end{equation}
where $\mathrm{Perc}(P,\,1-r)$ is the $(1-r)$-quantile of the quality map $P$, 
and $r \in (0,1)$ controls the fraction of resampled regions. 
The mask is then reshaped into the spatial grid $\bm{M} \in \{0,1\}^{H_s \times W_s}$ 
to guide localized refinement. We compute the attention mask at $t=0$, matching the final image.

\begin{figure}[t]
  \begin{center}
  \includegraphics[width=0.98\linewidth]{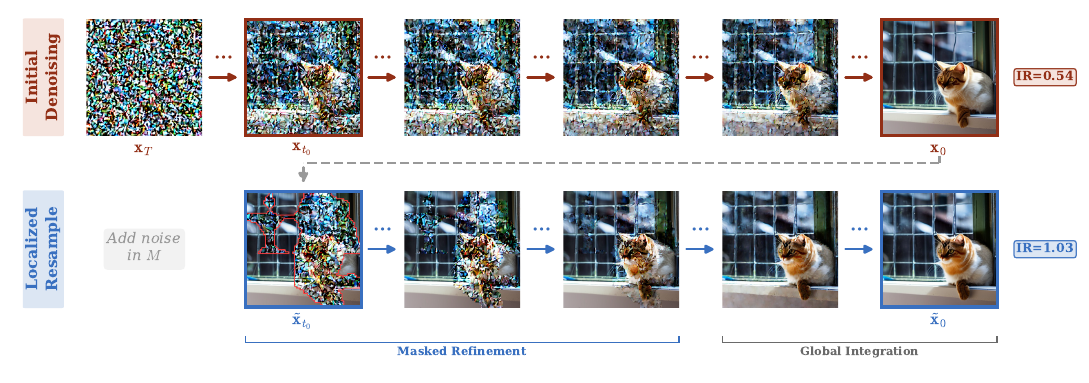}
  \vspace{-3mm}
  \caption{\textbf{Localized Resampling Process.} \textit{Initial Denoising}: standard denoising from $\bm{x}_T$ produces $\bm{x}_0$ with localized artifacts (IR\,=\,0.54). \textit{Localized Resample}: LoTTS injects noise within the defect mask $\bm{M}$ at step $t_0$, then performs \textit{Masked Refinement} followed by \textit{Global Integration}, yielding $\tilde{\bm{x}}_0$ with improved quality (IR\,=\,1.03).}
  \vspace{-2mm}
  \label{fig:resample_process}
  \end{center}
  \end{figure}

\subsection{Consistency Maintenance}

With reliable defect masks from the previous step, the next challenge is how to resample 
the identified regions without disrupting the rest of the image. 
Restricting updates only to masked regions often introduces boundary artifacts or semantic drift. 
To address these issues, LoTTS maintains spatial, temporal, and holistic consistency during refinement, as illustrated in Fig.~\ref{fig:resample_process}.

\myparagraph{Spatial Consistency.}
Resampling in diffusion models typically begins by perturbing the latent representation. 
If perturbation is applied only within the mask, the noise distribution becomes inconsistent 
with surrounding regions, creating visible seams. 
We avoid this by injecting comparable noise into both masked and unmasked areas, 
which balances noise levels and ensures smooth transitions across boundaries. 
Formally, given the clean latent $\bm{x}_0$ and binary mask $\bm{M}$, 
we initialize the perturbed latent at timestep $t_0$ as:
\begin{equation}
  \bm{x}_{t_0} = \alpha(t_0)\bm{x}_0 + \sigma(t_0)((1-\bm{M}) \odot \bm{z}_{\mathrm{bg}} + \bm{M} \odot \bm{z}_{\mathrm{mask}}),
\end{equation}
where $\bm{z}_{\mathrm{bg}}\!\sim\!\mathcal{N}(0,I)$ and $\bm{z}_{\mathrm{mask}}\!\sim\!\mathcal{N}(0,I)$.

\myparagraph{Temporal Consistency.}
A second challenge is preserving the global semantics of the image. 
Restarting from pure noise discards structure and forces the model to regenerate the entire scene. 
Instead, following Meng \etal~\citep{meng2021sdedit}, we resume denoising from an intermediate step $t_0$, 
so that global content is retained while localized corrections remain possible. 
The masked reverse update is:
\begin{equation}
  \begin{split}
  \bm{x}_{t-\Delta t}
  = &\,(1-\bm{M}) \odot \big(\alpha(t{-}\Delta t)\bm{x}_0 + \sigma(t{-}\Delta t)\bm{z}\big) \\
    &+ \bm{M} \odot \Big(\bm{x}_{t}
       - \Delta t\,\epsilon_\theta(\bm{x}_t,t)
       + \sigma(t{-}\Delta t)\bm{z}\Big),
  \end{split}
\end{equation}
where $\epsilon_\theta(\bm{x}_t,t)$ denotes the predicted noise, 
$\sigma(t)$ is the noise schedule at step $t$, 
$\bm{z}\sim\mathcal{N}(0,I)$, and $\Delta t$ is the reverse step size. This keeps unmasked regions faithful while allowing masked regions to be selectively refined.

\myparagraph{Holistic Consistency.}
Even after localized refinement, residual inconsistencies may remain around mask boundaries.
To restore smooth transitions, we remove the mask and apply a few global denoising steps, allowing the model to harmonize style and structure across the entire image.

\section{Experiments and Analysis}\label{sec:exp}

\begin{figure}[th]
  \centering
  \includegraphics[width=0.98\linewidth]{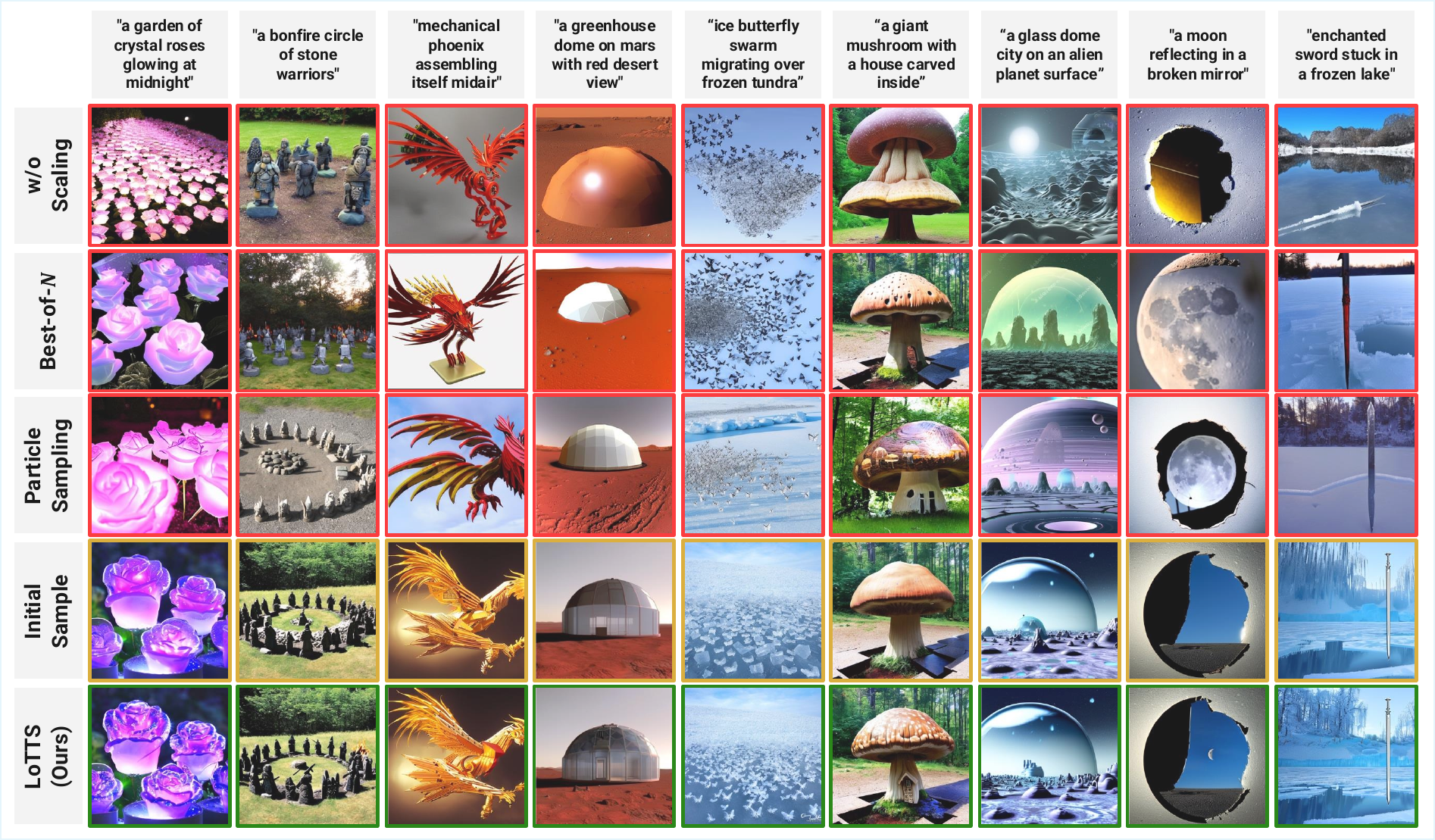}
  \vspace{-2mm}
  \caption{\textbf{Qualitative results on challenging text-to-image prompts.} Compared to Resampling, Best-of-$N$, and Particle Sampling, LoTTS better follows complex prompts. The 
  \textcolor{green}{green} borders indicate high-quality generations, \textcolor{red}{red} mark lower-quality ones, and \textcolor{brown}{brown} shows the initial image used as the starting point of localized refinement.}
  \label{fig:qualitative_comparison}
\end{figure}

\begin{table}[t]
  \centering
  \caption{\textbf{Quantitative results on three benchmarks (Pick-a-Pic, DrawBench, COCO2014).} Across human-aligned metrics (HPS, AES, Pick, IR) and automated metrics (FID, CLIP), LoTTS achieves the best overall performance across Resampling, Best-of-$N$, and Particle Sampling baselines.}
  \vspace{-2mm}
  \label{tab:quantitative_results}
  \resizebox{0.90\linewidth}{!}{%
  \begin{tabular}{c lcccc cccc cc}
  \toprule
  \multirow{2}{*}{Model} & \multirow{2}{*}{Method} 
  & \multicolumn{4}{c}{Pick-a-Pic} & \multicolumn{4}{c}{DrawBench} & \multicolumn{2}{c}{COCO2014} \\
  \cmidrule(lr){3-6} \cmidrule(lr){7-10} \cmidrule(lr){11-12}
  & & HPS$\uparrow$ & AES$\uparrow$ & Pick$\uparrow$ & IR$\uparrow$
    & HPS$\uparrow$ & AES$\uparrow$ & Pick$\uparrow$ & IR$\uparrow$
    & FID$\downarrow$ & CLIP$\uparrow$ \\
  \midrule
  \multirow{4}{*}{SD2.1} 
  & Resampling         & 20.44 & 5.377 & 20.32 & 0.236 & 21.34 & 5.456 & 20.23 & 0.244 & 15.33 & 0.201 \\
  & Best-of-$N$        & 21.56 & 5.534 & 21.04 & 0.470 & 22.45 & 5.589 & 20.59 & 0.446 & 13.21 & 0.252 \\
  & Particle Sampling  & 23.44 & \textbf{5.980} & 21.30 & 0.530 & 22.19 & 5.790 & 21.23 & 0.520 & 12.34 & 0.260 \\
  \rowcolor{purple!15}
  & \textbf{LoTTS (Ours)} & \textbf{24.52} & 5.805 & \textbf{21.32} & \textbf{0.680} & \textbf{23.29} & \textbf{5.911} & \textbf{21.47} & \textbf{0.698} & \textbf{10.89} & \textbf{0.263} \\
  \midrule
  \multirow{4}{*}{SDXL} 
  & Resampling         & 23.44 & 6.011 & 21.18 & 0.680 & 23.84 & 6.034 & 21.09 & 0.657 &  9.56 & 0.234 \\
  & Best-of-$N$        & 24.54 & 6.198 & 22.01 & 0.790 & 25.27 & 6.238 & 22.23 & 0.756 &  8.34 & 0.268 \\
  & Particle Sampling  & 25.33 & 6.235 & 22.05 & 0.865 & 26.46 & 6.233 & 22.31 & 0.844 &  7.99 & 0.271 \\
  \rowcolor{purple!15}
  & \textbf{LoTTS (Ours)} & \textbf{28.23} & \textbf{6.304} & \textbf{22.30} & \textbf{1.102} & \textbf{28.90} & \textbf{6.321} & \textbf{22.38} & \textbf{1.111} & \textbf{7.33} & \textbf{0.297} \\
  \midrule
  \multirow{4}{*}{FLUX} 
  & Resampling         & 29.34 & 6.298 & 22.07 & 1.038 & 29.28 & 6.223 & 22.05 & 1.100 &  7.01 & 0.282 \\
  & Best-of-$N$        & 30.23 & 6.299 & 22.89 & 1.235 & 30.46 & 6.290 & 22.33 & 1.221 &  6.34 & 0.306 \\
  & Particle Sampling  & 31.56 & \textbf{6.532} & \textbf{23.31} & 1.450 & 32.28 & 6.523 & 22.90 & 1.445 &  6.02 & 0.332 \\
  \rowcolor{purple!15}
  & \textbf{LoTTS (Ours)} & \textbf{33.33} & 6.501 & 23.04 & \textbf{1.605} & \textbf{33.90} & \textbf{6.890} & \textbf{23.21} & \textbf{1.623} & \textbf{5.31} & \textbf{0.351} \\
  \bottomrule
  \end{tabular}}
\end{table}

\myparagraph{Benchmarks.} 
We evaluate LoTTS on three benchmarks: Pick-a-Pic~\citep{pick}, DrawBench~\citep{saharia2022photorealistic}, and COCO2014~\citep{lin2014microsoft}. Performance is measured by four human-aligned metrics, HPS(v2)~\citep{wu2023human}, PickScore~\citep{pick}, ImageReward (IR)~\citep{ir}, and Aesthetic Score (AES)~\citep{aes_laion}, together with FID~\citep{heusel2017gans} and CLIP Score~\citep{radford2021learning} on COCO. Following prior work~\citep{kim2025test,singhal2025general}, all methods use the same verifier (IR) to select the best candidate, ensuring a fair comparison.  

\myparagraph{Baselines.} Experiments are conducted on SD2.1~\citep{rombach2022high}, SDXL~\citep{podell2023sdxl}, and Flux.1-schnell~\citep{flux2024}. SD2.1 and SDXL are diffusion-based, while Flux.1-schnell is a rectified flow-based model. We compare LoTTS with representative sampling and search methods under matched NFE budgets, including vanilla Resampling (standard sampling with a single random seed and no verifier selection), Best-of-$N$, and Particle Sampling~\citep{kim2025test,singhal2025general}.  

\myparagraph{Implementation Details.} 
Unless otherwise specified, LoTTS uses a mask ratio of $r=50\%$, an attention reweighting coefficient $\lambda=0.5$, a sampler-dependent noise-injection step $t_0$ (25/15/1 for SD2.1/SDXL/FLUX), and a default sample count of $N=9$ per prompt, obtained from $s=3$ initial images and $k=2$ localized refinement candidates for each. More implementation details are provided in Appendix~\ref{app:repro}.

\begin{figure}[t]
\centering
\includegraphics[width=0.98\linewidth]{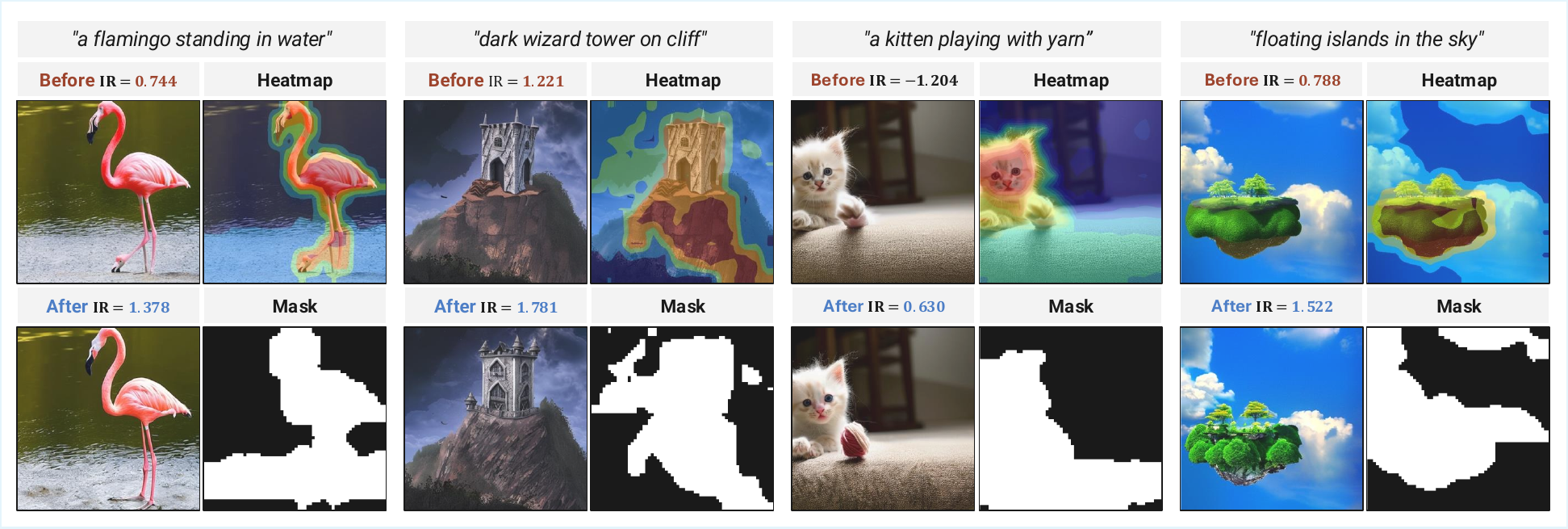}
\vspace{-2mm}
\caption{\textbf{Localized Refinement on SD2.1.} LoTTS corrects diverse localized artifacts (\eg distorted hands and faces). Heatmaps show per-region artifact scores, and binary masks indicate the regions selected for refinement.}
\label{fig:before_after_comparison}
\end{figure}

\subsection{Main Results}
\label{sec:experiments_main_results}
\myparagraph{Quantitative Performance.} Tab.~\ref{tab:quantitative_results} shows that LoTTS consistently outperforms Resampling, Best-of-$N$, and Particle Sampling~\citep{kim2025test,singhal2025general} under matched NFE budgets across SD2.1, SDXL, and FLUX, achieving clear improvements on both human-aligned metrics (HPS, AES, Pick, IR) and automated metrics. Notably, the gains extend beyond the verifier metric (IR) to independent metrics such as HPS, PickScore, FID, and CLIP, confirming genuine quality improvement rather than verifier overfitting. The consistent gains across three benchmarks and three architectures (U-Net-based SD2.1/SDXL and Transformer-based FLUX) confirm that LoTTS generalizes well and is architecture-agnostic.

\myparagraph{Human Evaluation.} We conduct a blind pairwise preference study on SD2.1 with 100 prompts. Five human annotators independently choose the preferred image from each pair without knowing the method identity, yielding 500 total judgments. LoTTS is preferred over Best-of-$N$ in 86\% of cases, consistent with the automatic metric trends.

\myparagraph{Qualitative Comparison.}
Figure~\ref{fig:qualitative_comparison} shows that unscaled results often contain severe artifacts, while Best-of-$N$ and Particle Sampling still exhibit issues such as inconsistent lighting and local distortions. Starting from the initial sample (brown border), LoTTS performs targeted local refinements while preserving the global layout, producing more stable and higher-fidelity images.

\myparagraph{Local Refinement Capability.}
Fig.~\ref{fig:before_after_comparison} shows several examples of localized refinement by LoTTS. The flamingo case corrects a duplicated bird head near the legs, while the kitten example introduces the missing yarn mentioned in the prompt. For the wizard tower and floating islands, refinement mainly enhances structural details and visual clarity. Heatmaps highlight artifact-prone regions, and binary masks indicate the areas selected for localized resampling, leading to consistent improvements in ImageReward (IR).

\myparagraph{More Results.}
Additional qualitative results on SDXL and FLUX, as well as failure cases, are provided in Appendix~\ref{app:qual} and~\ref{app:fail}.

\subsection{Ablations}
\label{sec:experiments_ablations}

\begin{figure}[t]
  \centering
  \includegraphics[width=0.7\linewidth]{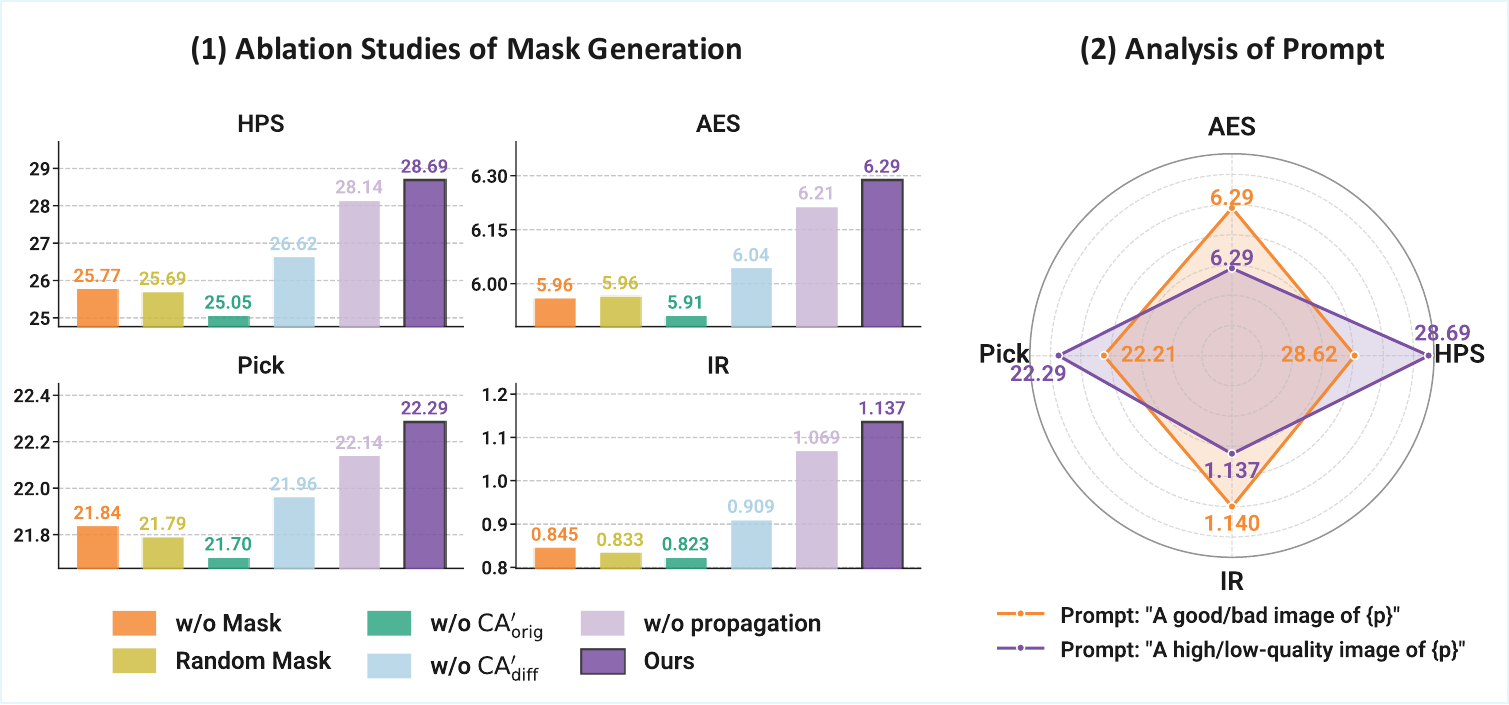}
  \vspace{-2mm}
  \caption{\textbf{(1) Ablation of Mask Generation.} Removing any component consistently degrades all four metrics. \textbf{(2) Prompt Design.} Radar chart comparing two prompt strategies: ``Good/Bad'' (``a good/bad image of \{p\}'') and ``High/Low-Quality'' (``a high/low-quality image of \{p\}''). The nearly overlapping profiles show that LoTTS is robust to prompt phrasing.}
  \label{fig:ablation_MaskGeneration}
\end{figure}

\myparagraph{Mask Generation.}
As summarized in Fig.~\ref{fig:ablation_MaskGeneration}(1), removing any component -- whether the discriminative difference map $\mathrm{CA}_{\mathrm{diff}}^{\prime}$, the propagated original attention $\mathrm{CA}_{\mathrm{orig}}^{\prime}$, or the context-aware propagation -- consistently degrades all four metrics. The variant without $\mathrm{CA}_{\mathrm{orig}}^{\prime}$ performs the worst overall, indicating the importance of the strong semantic foreground prior. The no-mask and random-mask baselines also underperform, further confirming the need for accurate, structure-aware masks.

\begin{figure}[t]
    \centering
    \includegraphics[width=0.75\linewidth]{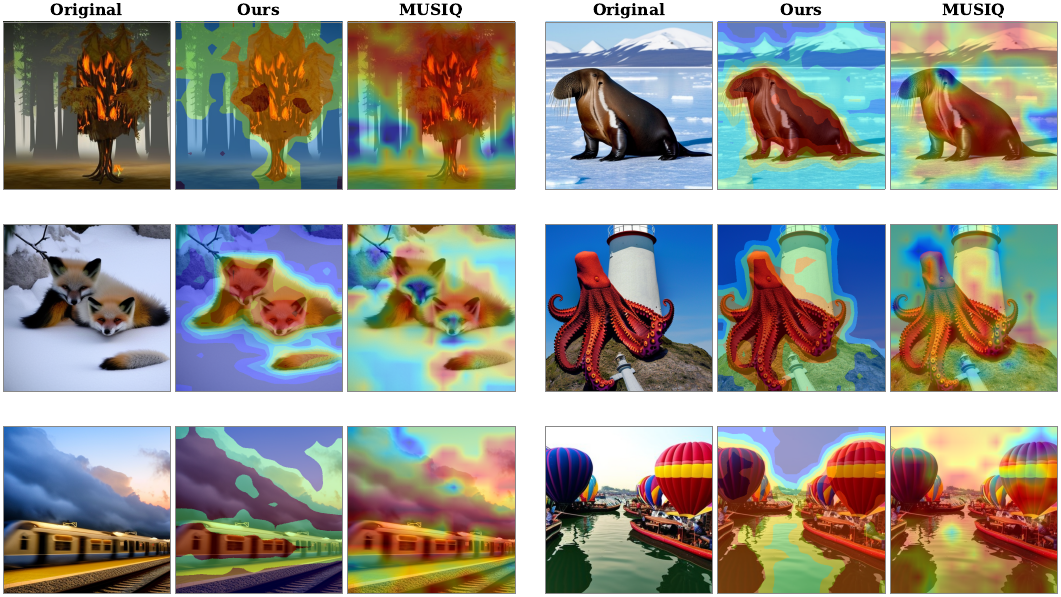}
    \vspace{-2mm}
    \caption{\textbf{Comparison with MUSIQ~\citep{ke2021musiq} quality maps.} Each triplet shows the original image, our attention heatmap, and the MUSIQ heatmap (red = lower quality). Both methods consistently highlight similar low-quality regions.}
    \label{fig:mask_comparison}
\end{figure}

\myparagraph{External Validation of Mask Quality.}
To verify that our masks reflect genuine quality signals, we compare them with pixel-level quality heatmaps from MUSIQ~\citep{ke2021musiq}, an independent NR-IQA model, computed via sliding windows (patch 64, stride 32). As shown in Fig.~\ref{fig:mask_comparison}, both methods consistently highlight similar low-quality regions, despite ours being entirely training-free. This confirms that diffusion cross-attention encodes meaningful perceptual quality information that LoTTS can exploit without any additional training.

\myparagraph{Quality Prompt Design.} As shown in Fig.~\ref{fig:ablation_MaskGeneration}(2), different auxiliary prompts (\eg ``a good/bad image of \{p\}'') yield nearly identical metric profiles, indicating that LoTTS is robust to prompt phrasing and relies on quality contrast rather than specific wording.

\begin{figure}[t]
    \centering
    \includegraphics[width=0.9\linewidth]{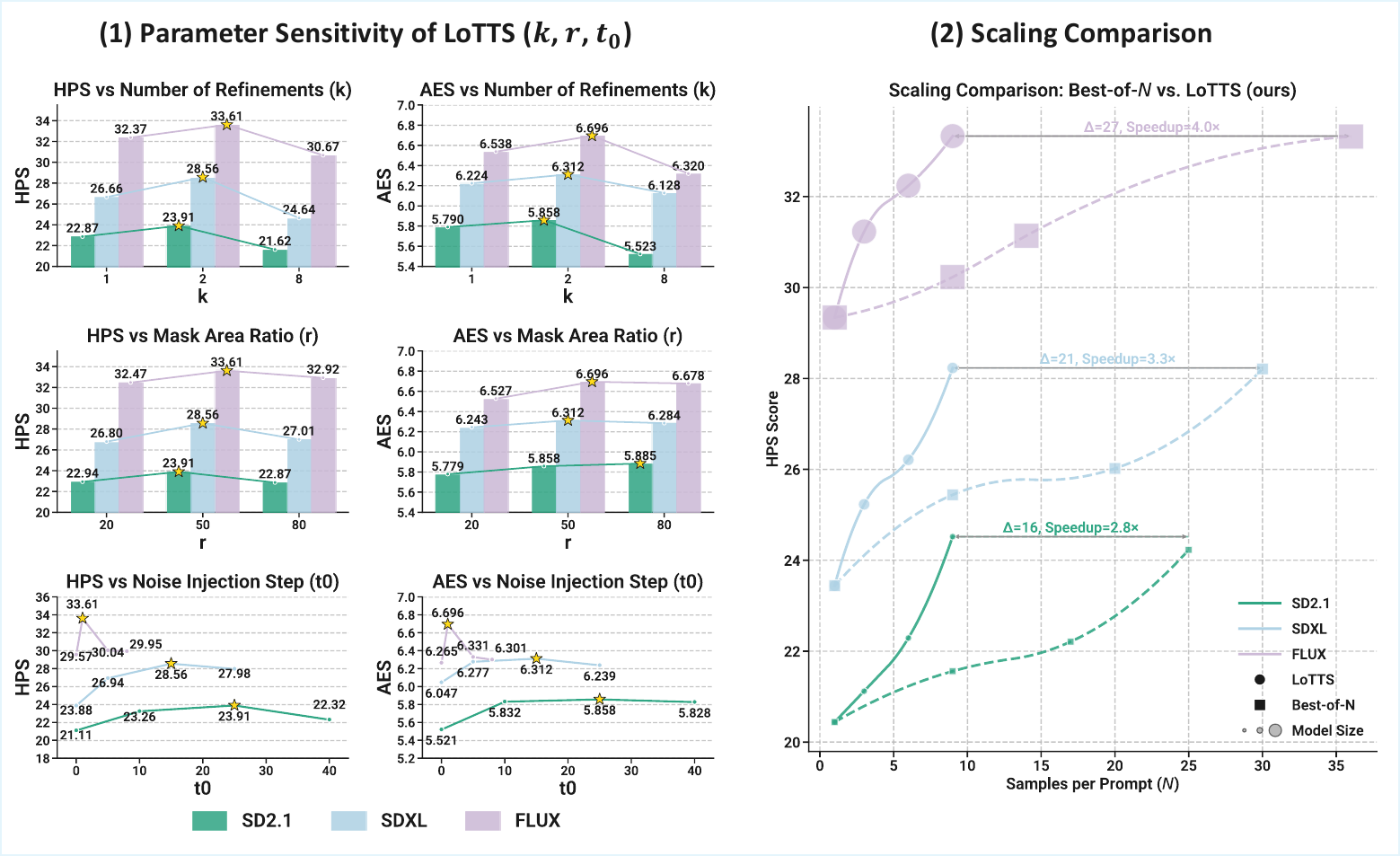}
    \vspace{-2mm}
    \caption{\textbf{(1) Parameter sensitivity analysis of LoTTS.} Varying localized refinement iterations ($k$), resample area ratio ($r$), and denoising start step ($t_{0}$) shows that LoTTS maintains stable improvements across HPS and AES metrics. \textbf{(2) Scaling Comparison:} LoTTS vs.~Best-of-$N$ Sampling. LoTTS reaches the same HPS with far fewer samples, yielding significant speedups.}
    \label{fig:ParameterAnalysis}
    \label{fig:scaling}
\end{figure}

\begin{table}[t]
\centering
\small
\caption{\textbf{Wall-clock latency comparison.} Per-image latency (seconds) and total time to reach comparable quality. LoTTS achieves similar per-image cost but significantly lower total runtime.}
\vspace{-1mm}
\label{tab:latency}
\resizebox{0.7\linewidth}{!}{%
\begin{tabular}{l cc cc cc}
\toprule
& \multicolumn{2}{c}{\textbf{SD2.1}} & \multicolumn{2}{c}{\textbf{SDXL}} & \multicolumn{2}{c}{\textbf{FLUX}} \\
\cmidrule(lr){2-3} \cmidrule(lr){4-5} \cmidrule(lr){6-7}
\textbf{Method} & \textbf{Per-img} & \textbf{Total} & \textbf{Per-img} & \textbf{Total} & \textbf{Per-img} & \textbf{Total} \\
\midrule
Best-of-$N$       & 0.613 & 13.80 & 0.294 & 7.050 & 0.154 & 3.888 \\
Particle Sampling  & 0.616 &  8.31 & 0.298 & 4.760 & 0.157 & 2.750 \\
\rowcolor{purple!15}
\textbf{LoTTS (Ours)} & 0.618 & \textbf{5.01} & 0.298 & \textbf{2.16} & 0.158 & \textbf{1.04} \\
\midrule
\multicolumn{1}{r}{Speedup vs.~Best-of-$N$} & -- & \textbf{2.75$\times$} & -- & \textbf{3.26$\times$} & -- & \textbf{3.76$\times$} \\
\bottomrule
\end{tabular}}
\end{table}

\myparagraph{Parameter Sensitivity ($k$, $r$, $t_0$).}
As shown in Fig.~\ref{fig:ParameterAnalysis}(1), LoTTS is robust across a wide range of hyperparameters. Performance peaks at $k{=}2$ refinements, as two attempts provide sufficient stochastic diversity to fix most artifacts. A moderate mask ratio $r{=}50\%$ works best: smaller masks miss defects while larger ones overwrite clean regions. The noise injection step $t_0$ balances preservation and refinability -- mid-range steps suit SD2.1/SDXL, while FLUX benefits from earlier injection due to its rapid convergence.

\myparagraph{Scaling Efficiency.} As shown in Fig.~\ref{fig:scaling}(2), LoTTS reaches the same HPS as Best-of-$N$ with $2.8\times$--$3.3\times$ fewer samples on SD2.1 and SDXL, and up to $4\times$ fewer on FLUX, with larger savings on stronger models.

\myparagraph{Wall-clock Latency.} Tab.~\ref{tab:latency} translates the above sample savings into wall-clock time: LoTTS achieves comparable per-image latency while requiring far fewer samples to reach the target quality, resulting in $2.75\times$--$3.76\times$ lower total runtime.

\myparagraph{More Ablations.}
Further ablations on $\lambda$ and the consistency components, along with extended parameter analyses, are provided in Appendix~\ref{app:ablation} and~\ref{app:parameter_analysis}.

\subsection{Discussion}

\begin{figure}[t]
    \centering
    \includegraphics[width=0.95\linewidth]{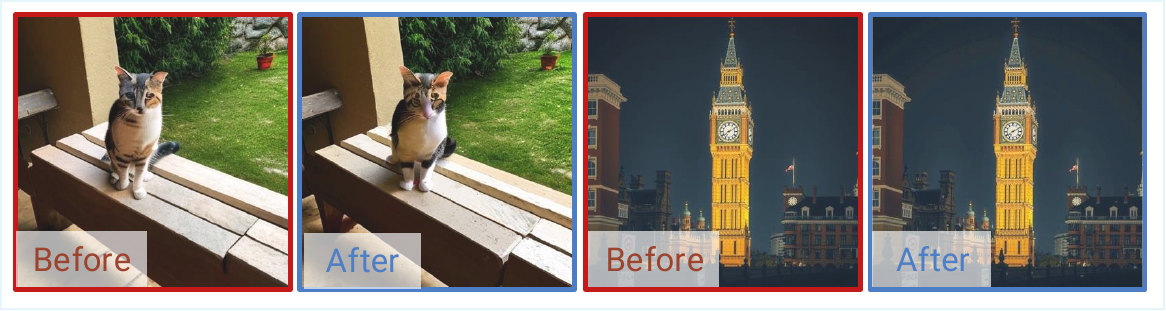}
    \vspace{-3mm}
    \caption{\textbf{Failure cases of LoTTS.} Left: local refinement introduces a facial distortion in the cat image. Right: the image has minimal visible change after refinement.}
    \vspace{-5mm}
    \label{fig:failure_case_main}
\end{figure}

As shown in Fig.\ref{fig:failure_case_main}, LoTTS occasionally produces failures. We observe that LoTTS failures mainly fall into three categories. First, \textit{imperfect masks}:  extremely subtle defects or high-frequency regions may cause missed or unstable attention signals. This challenge is not unique to our training-free LoTTS -- supervised defect-detection models exhibit similar limitations -- but importantly, our Appendix~\ref{app:TheoreticalAnalysis} shows that localized TTS can outperform global resampling even with imperfect masks. Second, \textit{global defects}: issues involving perspective, spatial relations, or global semantics cannot be fixed through local resampling alone, pointing toward future work on adaptive local--global TTS strategies. Third, \textit{initial images at quality extremes}: low-quality images may be too fragile to refine locally, while near-perfect images offer little room for improvement, suggesting that early-stopping mechanisms may further improve LoTTS. Overall, these cases point to natural opportunities for extending localized TTS, and LoTTS marks an early step toward more adaptive scaling methods.

\section{Conclusion}\label{sec:conclusion}

We proposed LoTTS, a training-free framework that extends test-time scaling from global resampling to localized refinement. Unlike conventional methods that apply sampling uniformly, LoTTS leverages defect-aware masks and consistency constraints to focus computation where it matters most. This design not only improves image quality and efficiency but also shows that scaling can be made adaptive to the spatial heterogeneity of generative outputs. Beyond the specific models tested, we believe the principle of region-aware scaling can benefit a broader class of generative architectures. Our analysis further establishes that localized scaling provably outperforms global sampling under mild conditions, and we hope LoTTS opens up promising directions for integrating fine-grained, spatially adaptive control into test-time algorithms.

\bibliographystyle{plainnat}
\bibliography{main}

\appendix
\clearpage
\section*{Table of Contents}

\appitem{app:repro}
\appitem{app:algo}
\appsubitem{app:algo_1}
\appsubitem{app:algo_2}
\appsubitem{app:algo_3}
\appsubitem{app:algo_4}
\appitem{app:qual}
\appsubitem{app:qual_1}
\appsubitem{app:qual_2}
\appitem{app:ablation}
\appsubitem{app:mask_generation}
\appsubitem{app:prompt_design}
\appsubitem{app:lambda_ablation}
\appsubitem{app:consistency}
\appitem{app:parameter_analysis}
\appsubitem{app:parameter_sensitivity}
\appsubitem{app:scaling_comparison}
\appitem{app:fail}
\appsubitem{app:fail_1}
\appsubitem{app:fail_2}
\appitem{app:TheoreticalAnalysis}
\appsubitem{app:TheoreticalAnalysis_1}
\appsubitem{app:TheoreticalAnalysis_2}
\appsubitem{app:TheoreticalAnalysis_3}
\appsubitem{app:TheoreticalAnalysis_4}

\clearpage

\begin{figure}[t]
\centering
\includegraphics[width=0.98\linewidth]{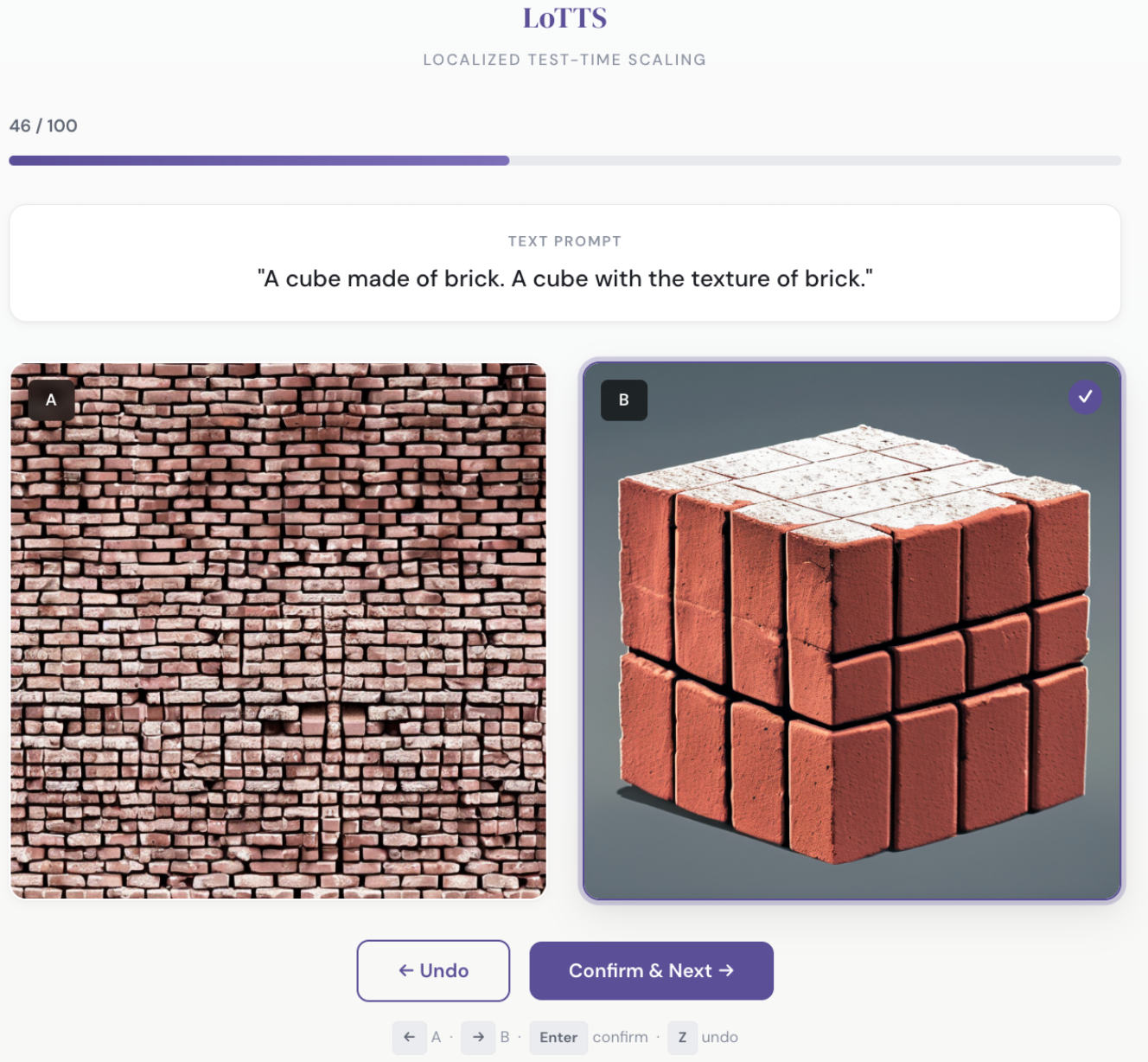}
\vspace{-5pt}
\caption{\textbf{Human evaluation interface.} Annotators perform blind pairwise
preference judgments between LoTTS and Best-of-$N$ outputs without knowing method identity.}
\label{fig:user_study_interface}
\end{figure}

\section{Implementation Details}
\label{app:repro}

\myparagraph{Hardware and software.}
All experiments were run on NVIDIA A100 (80GB) GPUs; most runs use a single GPU unless otherwise stated. Mixed precision is enabled via \texttt{torch.cuda.amp.autocast} (fp16) unless the metric implementation requires fp32.

\myparagraph{Checkpoints and preprocessing.}
We use the official public checkpoints for SD2.1, SDXL, and FLUX.1-schnell. 
Images are generated at $512{\times}512$ for SD2.1 and at $1024{\times}1024$ for SDXL and FLUX. 
For latent diffusion backbones, we use the default VAE shipped with each checkpoint. 
Evaluation images are fed to each metric with the metric's \emph{own} preprocessing. 
Unless noted, we do not run SDXL-Refiner to avoid confounding the effect of local resampling.

\myparagraph{Default sampler and schedule.}
For SD2.1, we use SDE-DPM-Solver$++$ sampling with 50 NFEs, and CFG scale $=7.5$. For SDXL, we use SDE-DPM-Solver$++$ sampling with 30 NFEs, and CFG scale $=5.5$. For FLUX, we use modified SDE-DPM-Solver$++$ sampling with 10 NFEs, and no CFG. 
For LoTTS, we inject noise to reach an intermediate step $t_0$ (Sec.~\ref{sec:method}), 
then perform local masked refinement followed by a short global integration phase. 
We keep the original prompt and CFG weight unchanged during refinement.

\begin{table}[t]
\centering
\small
\caption{\textbf{Default LoTTS hyperparameters.}}
\label{tab:default_hparams}
\vspace{-2mm}
\begin{tabular}{lccc}
\toprule
 & \textbf{SD\,2.1} & \textbf{SDXL} & \textbf{FLUX} \\
\midrule
\textbf{Res.} & $512{\times}512$ & $1024{\times}1024$ & $1024{\times}1024$ \\
\textbf{Sampler} & 
\makecell{SDE-DPM-\\Solver$++$} &
\makecell{SDE-DPM-\\Solver$++$} &
\makecell{Modified SDE-\\DPM-Solver$++$} \\
\textbf{Steps} & 50 & 30 & 10 \\
\textbf{CFG} & 7.5 & 5.5 & None \\
\textbf{\boldmath $\lambda$} & 0.50 & 0.50 & 0.50 \\
\textbf{\boldmath $t_0$} & 25 & 15 & 1 \\
\textbf{\boldmath $t_g$} & 5 & 3 & 1 \\
\textbf{\boldmath $k$} & 2 & 2 & 2 \\
\textbf{\boldmath $r\;(\%)$} & 50 & 50 & 50 \\
\bottomrule
\end{tabular}
\end{table}

\myparagraph{Attention hook and mask resolution.}
For SD2.1, we hook U-Net cross-attention tensors at the $16{\times}16$ spatial resolution blocks (for $512{\times}512$ images). For SDXL, we hook U-Net cross-attention tensors at the $64{\times}64$ spatial resolution blocks (for $1024{\times}1024$ images).  For FLUX, we extract Transformer cross-attention tensors from the $64{\times}64$ resolution of the last 10 blocks (for $1024{\times}1024$ images), and generate attention maps following the method of \citep{helbling2025conceptattention}. The attention maps are averaged across heads, mean-pooled across the feature dimension, and then averaged across tokens to obtain a single spatial map per prompt type (pos/neg/origin).
Self-attention maps are normalized to $[0,1]$. 
The quality map $P$ is computed via Eq.~\eqref{eq:Semantic_guided_Reweighting} with $\lambda=0.5$ by default. 
We compute attention and the final mask at $t{=}0$ (last step), where maps and image latents are already stable. 
Masks are kept in latent resolution during resampling; only for visualization we bilinearly upsample to pixel space.

\myparagraph{Datasets.} For Pick-a-Pic, we follow the official prompt pool for preference-oriented evaluation. For DrawBench, we use all categories and report aggregate metrics. For COCO2014, we use the standard COCO2014-30k split.

\myparagraph{Metrics.}
\label{app:metrics}
We report HPSv2, PickScore, ImageReward, and AES using the authors’ official weights and preprocessing. 
CLIP score and FID are computed with standard settings: 
CLIP uses ViT-L/14 unless otherwise stated; 
FID uses Inception-V3 features with $2048$-D activations. ImageReward (IR) score is used as the global selection score for Best-of-$N$, LoTTS, and other search-based baselines.

\myparagraph{Baselines and compute budgets.}
\label{app:baselines}
We compare with Resample, Best-of-$N$ and Particle Sampling under matched NFEs. 
For “Best-of-$N$” baselines, we sample $N$ candidates \emph{from scratch} and select by the same metric used for reporting to avoid selection bias. 
For LoTTS we use $k=2$ localized refinement iterations by default.

\myparagraph{Reproduction checklist.}
(1) Use the same checkpoint and VAE; 
(2) match resolution, sampler, steps, CFG; 
(3) compute the mask at $t{=}0$ with the same prompt templates; 
(4) keep $\lambda$, $r$, $t_0$, $t_g$, $k$ as in Tab.~\ref{tab:default_hparams}; 
(5) fix seeds and determinism flags; 
(6) run the same metric code and preprocessing.

\myparagraph{Human Evaluation Protocol.}
We conduct a blind pairwise preference study as described in Sec.~\ref{sec:exp}.
Fig.~\ref{fig:user_study_interface} shows the annotation interface: each annotator
views two images side-by-side without method labels and selects the preferred one.
Five annotators independently evaluated 100 prompts, yielding 500 judgments total.

\begin{table}[t]
\centering
\small
\caption{\textbf{Unified notation for Algorithms~\ref{alg:mask}, \ref{alg:resampling}, and \ref{alg:dfs}.}}
\label{tab:algo_notation}
\vspace{-2mm}
\begin{tabular*}{0.9\textwidth}{l @{\extracolsep{\fill}} p{0.7\textwidth}}
\toprule
\textbf{Symbol} & \textbf{Description} \\
\midrule
$p$ & original user prompt \\
$p_{\text{pos}}, p_{\text{neg}}$ & quality-aware prompts (``A high-quality image of $\{p\}$'', ``A low-quality image of $\{p\}$'') \\
$x$ & current image passed through DFS recursion \\
$x_{\mathrm{in}}, x_{\mathrm{out}}$ & input/output image of \texttt{LocalizedResample} \\
$v_{\mathrm{out}}$ & verifier score of $x_{\mathrm{out}}$ \\
$x_{\text{best}}, v_{\text{best}}$ & current best image and score during DFS search \\
$x^\star$ & final best refined image returned by DFS \\
$M$ & binary resample mask from Algorithm~\ref{alg:mask} (latent resolution) \\
$r$ & area ratio used in quantile thresholding \\
$\mathcal{L}$ & set of backbone layers used to extract cross-attention maps \\
$\lambda$ & foreground weighting coefficient in mask reweighting \\
$T$ & total number of diffusion steps (scheduler alignment) \\
$t_0$ & intermediate re-noise step for localized resampling \\
$t_g$ & global integration step \\
$S$ & number of global seeds at DFS depth 0 (root) \\
$k$ & number of localized refinements per global seed \\
$D$ & DFS maximum depth ($D{=}2$) \\
$\Delta t$ & scheduler step size used for discretization \\
$\alpha(t), \sigma(t)$ & noise schedule for both forward and reverse updates \\
$\epsilon_\theta(\cdot)$ & noise predictor (with CFG if enabled) \\
$V(\cdot)$ & global verifier producing a scalar score \\
$\text{SampleBase}$ & base sampler generating an initial global sample conditioned on $p$ \\
$\mathrm{CrossAttn}$ & cross-attention maps from backbone layers $\mathcal{L}$ during a forward pass with latent $x$ and prompt $p$ \\
\bottomrule
\end{tabular*}
\end{table}

\section{Algorithmic Details}
\label{app:algo}

This section presents the full algorithmic components of \textbf{LoTTS}.  
Algorithm~\ref{alg:mask} extracts a binary resample mask $M$ from prompt-conditioned 
cross-attention maps.  
Algorithm~\ref{alg:resampling} integrates this mask into a localized resampling module 
that refines only defect-prone regions while preserving background structure.  
Algorithm~\ref{alg:dfs} combines these components into a depth-first search (DFS) that 
explores global diversity and localized refinement, ultimately selecting the highest-scoring image.

\subsection{Notation and Preliminaries}
\label{app:algo_1}

Tab.~\ref{tab:algo_notation} summarizes all symbols used in 
Algorithms~\ref{alg:mask}, \ref{alg:resampling}, and \ref{alg:dfs}.  
The goal is to unify notation so that every variable appearing in the pseudocode 
is well-defined.  
We explicitly distinguish quantities related to 
user prompts,  
mask generation,  
localized refinement, and  
DFS-level search dynamics.

\subsection{Defect Localization Procedure}
\label{app:algo_2}

Algorithm~\ref{alg:mask} details the \textbf{MaskGen} module, whose goal is to identify 
regions likely to contain visual defects.  
The key intuition is that low-quality prompts activate areas where the model is 
susceptible to producing artifacts, whereas high-quality prompts suppress them.  
By contrasting their cross-attention maps and propagating information through 
self-attention layers, MaskGen amplifies quality-sensitive regions while preserving 
semantic coherence.  
Quantile thresholding is then applied to select only the top-$r$ most suspicious patches.  
The resulting binary mask $M$ provides spatial guidance for localized refinement.

\begin{algorithm}
\caption{MaskGen: Defect Localization}
\label{alg:mask}
\begin{algorithmic}[1]
\Require Current image $x$, prompt $p$, positive prompt $p_{\text{pos}}$, negative prompt $p_{\text{neg}}$, 
         selected layers $\mathcal{L}$, weight $\lambda$, area ratio $r$
\Ensure Binary mask $M$

\State \textbf{Prompt-driven Discrimination}
\State $\mathrm{CA}_{\mathrm{pos}} \gets \mathrm{CrossAttn}(p_{\text{pos}}, x, \mathcal{L})$
\State $\mathrm{CA}_{\mathrm{neg}} \gets \mathrm{CrossAttn}(p_{\text{neg}}, x, \mathcal{L})$
\State $\mathrm{CA}_{\mathrm{orig}} \gets \mathrm{CrossAttn}(p, x, \mathcal{L})$
\State $\mathrm{CA}_{\mathrm{diff}} \gets \mathrm{CA}_{\mathrm{neg}} - \mathrm{CA}_{\mathrm{pos}}$

\State \textbf{Context-aware Propagation}
\State $\mathrm{CA}'_{\mathrm{diff}} \gets \mathrm{SelfAttnProp}(\mathrm{CA}_{\mathrm{diff}})$
\State $\mathrm{CA}'_{\mathrm{orig}} \gets \mathrm{SelfAttnProp}(\mathrm{CA}_{\mathrm{orig}})$

\State \textbf{Semantic-guided Reweighting}
\State $P \gets \mathrm{CA}'_{\mathrm{diff}} + \lambda \cdot \mathrm{CA}'_{\mathrm{orig}}$

\State \textbf{Mask Generation}
\State $M \gets \mathbb{I}[P > \mathrm{Perc}(P, 1-r)]$
\State \Return $M$
\end{algorithmic}
\end{algorithm}

\subsection{Localized Resampling Procedure}
\label{app:algo_3}

Algorithm~\ref{alg:resampling} defines \textbf{LocalizedResample}, which refines only the 
defective regions indicated by mask $M$.  
The process begins by re-noising the entire image to an intermediate timestep $t_0$, 
ensuring that both masked and unmasked regions lie on the diffusion trajectory.  
During the refinement phase ($t_0 \!\rightarrow\! t_g$), masked regions perform reverse 
denoising steps guided by $\epsilon_\theta$, while unmasked regions follow a scheduled 
forward-noise injection that preserves background content.  
A final integration phase ($t_g \!\rightarrow\! 0$) restores global coherence using 
standard reverse diffusion.  
The verifier $V$ evaluates the refined output, providing a scalar score used in DFS.

\begin{algorithm}
\caption{LocalizedResample: Attention-Guided Resampling (single refinement)}
\label{alg:resampling}
\begin{algorithmic}[1]
\Require Input image $x_{\mathrm{in}}$ (at $t{=}0$), prompt $p$, total steps $T$,
         re-noise step $t_0$, integration step $t_g$, step size $\Delta t$, mask $M$, diffusion model, verifier $V$
\Ensure Refined image $x_{\mathrm{out}}$, score $v_{\mathrm{out}}$

\State $x_{\text{anchor}} \gets x_{\mathrm{in}}$

\State \textbf{Global re-noise to $t_0$}
\State Sample $z_{\text{bg}}, z_{\text{mask}} \sim \mathcal{N}(0,I)$
\State $x_{t_0} \gets \alpha(t_0)x_{\text{anchor}} 
       + \sigma(t_0)(M\odot z_{\text{mask}} + (1-M)\odot z_{\text{bg}})$

\State \textbf{Masked refinement ($t_0 \rightarrow t_g$)}
\For{$t = t_0,\, t_0-\Delta t,\, \dots,\, t_g$}
    \State Sample $z_t \sim \mathcal{N}(0,I)$
    \State $x_{t-\Delta t} \gets (1-M)\odot(\alpha(t{-}\Delta t)x_{\text{anchor}} + \sigma(t{-}\Delta t)z_t)
            + M\odot(x_t - \Delta t\,\epsilon_\theta(x_t,t;p) + \sigma(t{-}\Delta t)z_t)$
\EndFor

\State \textbf{Global integration ($t_g \rightarrow 0$)}
\For{$t = t_g,\, t_g-\Delta t,\, \dots,\, 0$}
    \State Sample $z_t \sim \mathcal{N}(0,I)$
    \State $x_{t-\Delta t} \gets x_t - \Delta t\,\epsilon_\theta(x_t,t;p) + \sigma(t)z_t$
\EndFor

\State $x_{\mathrm{out}} \gets x_0$
\State $v_{\mathrm{out}} \gets V(x_{\mathrm{out}})$
\State \Return $x_{\mathrm{out}},\, v_{\mathrm{out}}$

\Statex {\footnotesize \textit{Note: $T$ is included for scheduler compatibility; discretization uses $\Delta t$.}}
\end{algorithmic}
\end{algorithm}

\subsection{Overall Algorithm of LoTTS}
\label{app:algo_4}

Algorithm~\ref{alg:dfs} presents the full \textbf{LoTTS} pipeline expressed as a 
depth-first search with depth $D{=}2$.  
Depth~0 explores global diversity by sampling $S$ initial images from scratch via 
$\text{SampleBase}$.  
At depth~1, each global sample undergoes $k$ localized refinements guided by 
Algorithm~\ref{alg:resampling}.  
Every candidate is scored by the verifier $V$, and DFS continually tracks the highest 
score seen so far.  
This hierarchical search effectively combines global variability with targeted local 
improvements, achieving strong test-time scaling.

\begin{algorithm}
\caption{Overall Algorithm of LoTTS (DFS with depth $D{=}2$)}
\label{alg:dfs}
\begin{algorithmic}[1]
\Require Prompt $p$, DFS depth $D{=}2$,
         global seeds $S$, localized refinements $k$,
         total steps $T$, re-noise $t_0$, integration $t_g$, step size $\Delta t$, verifier $V$
\Ensure Best refined image $x^\star$

\Function{DFS}{$p, d, x, x_{\text{best}}, v_{\text{best}}$}
    \If{$x \neq \varnothing$}
        \State $v \gets V(x)$
        \If{$v > v_{\text{best}}$}
            \State $x_{\text{best}} \gets x$
            \State $v_{\text{best}} \gets v$
        \EndIf
    \EndIf
    \If{$d = D$} \State \Return $(x_{\text{best}}, v_{\text{best}})$ \EndIf

    \If{$d = 0$} \Comment{root $\rightarrow$ global seeds}
        \For{$i=1$ \textbf{to} $S$}
            \State $x_{\mathrm{in}} \gets \text{SampleBase}(p)$
            \State $(x_{\text{best}}, v_{\text{best}}) 
                   \gets \Call{DFS}{p,1,x_{\mathrm{in}},x_{\text{best}},v_{\text{best}}}$
        \EndFor
        \State \Return $(x_{\text{best}}, v_{\text{best}})$

    \Else \Comment{$d=1$: global sample $\rightarrow k$ localized refinements}
        \State $M \gets \text{MaskGen}(x,p)$
        \For{$j=1$ \textbf{to} $k$}
            \State $(x_{\mathrm{out}}, v_{\mathrm{out}}) \gets 
                   \text{LocalizedResample}(x,p,T,t_0,t_g,\Delta t,M,V)$
            \State $(x_{\text{best}}, v_{\text{best}}) 
                   \gets \Call{DFS}{p,2,x_{\mathrm{out}},x_{\text{best}},v_{\text{best}}}$
        \EndFor
        \State \Return $(x_{\text{best}}, v_{\text{best}})$
    \EndIf
\EndFunction

\State $(x^\star,\,\_) \gets \Call{DFS}{p,0,\varnothing,\varnothing,-\infty}$
\State \Return $x^\star$
\end{algorithmic}
\end{algorithm}

\begin{figure*}[ht]
\centering
\includegraphics[width=0.85\linewidth]{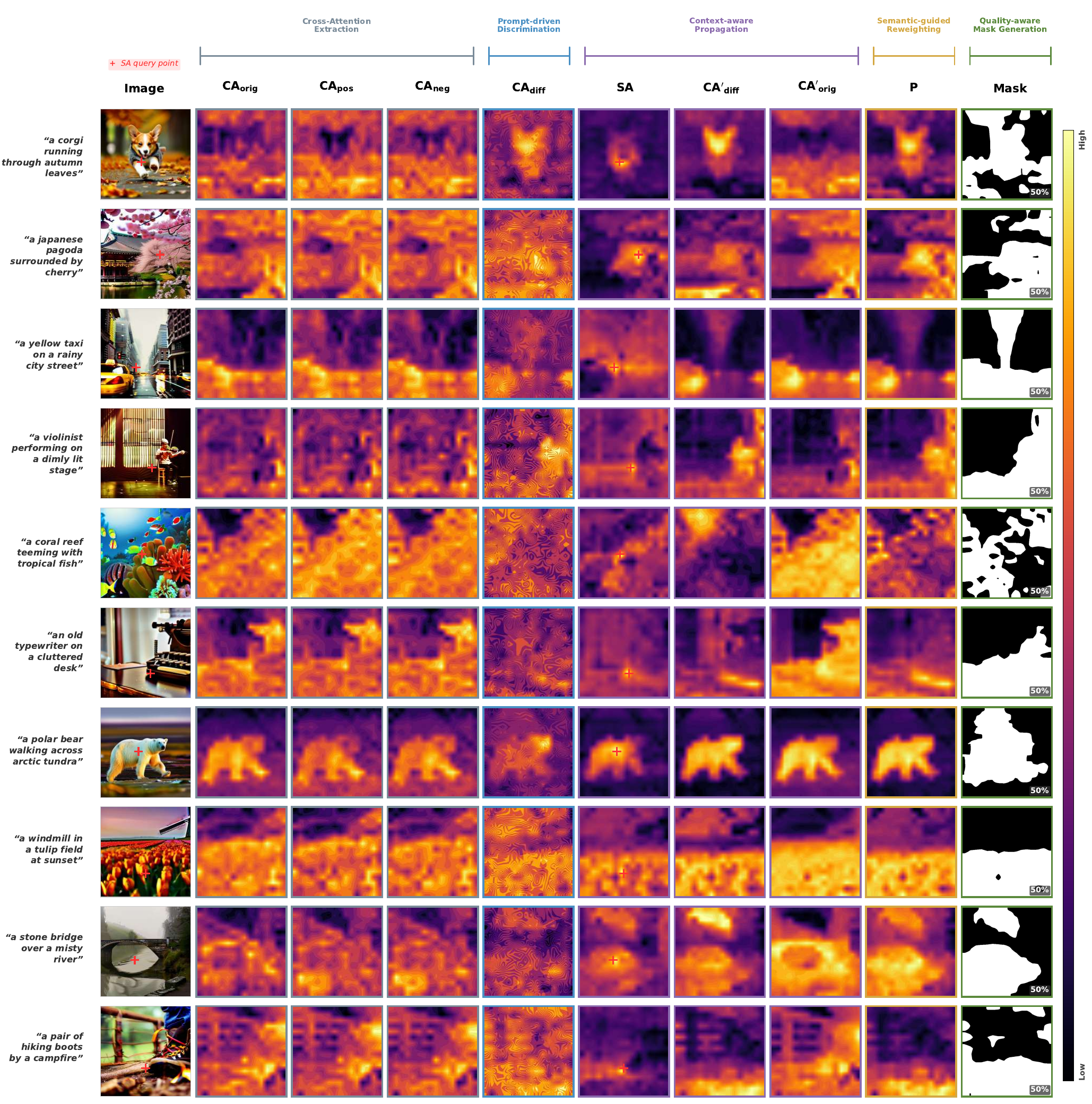}
\vspace{-3mm}
\caption{\textbf{Additional examples of mask generation in SD2.1.} LoTTS produces reliable quality-aware masks across diverse prompts, effectively localizing low-quality regions for refinement.}
\label{fig:additional_mask_examples}
\end{figure*}

\section{Qualitative Results}
\label{app:qual}

We provide additional qualitative comparisons to highlight LoTTS's local refinement capabilities.  

\subsection{Mask Generation Examples}
\label{app:qual_1}

\begin{figure*}[t]
\centering
\includegraphics[width=0.85\linewidth]{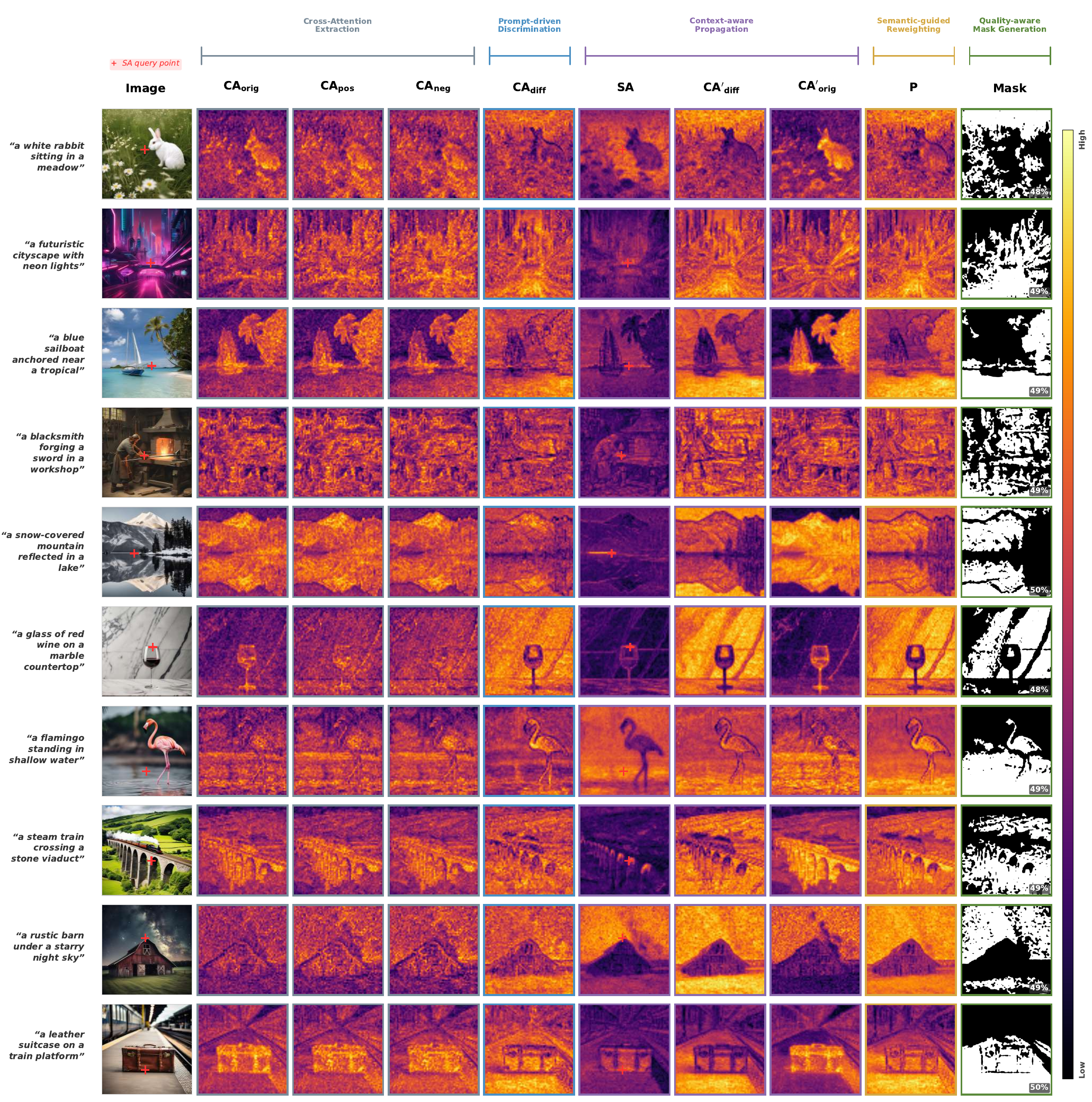}
\vspace{-3mm}
\caption{\textbf{Additional examples of mask generation in SDXL.} Compared to SD2.1, LoTTS produces sharper and more spatially precise masks in SDXL, accurately delineating defect boundaries for targeted refinement.}
\label{fig:additional_mask_examples_sdxl}
\end{figure*}

\begin{figure*}[t]
\centering
\includegraphics[width=0.85\linewidth]{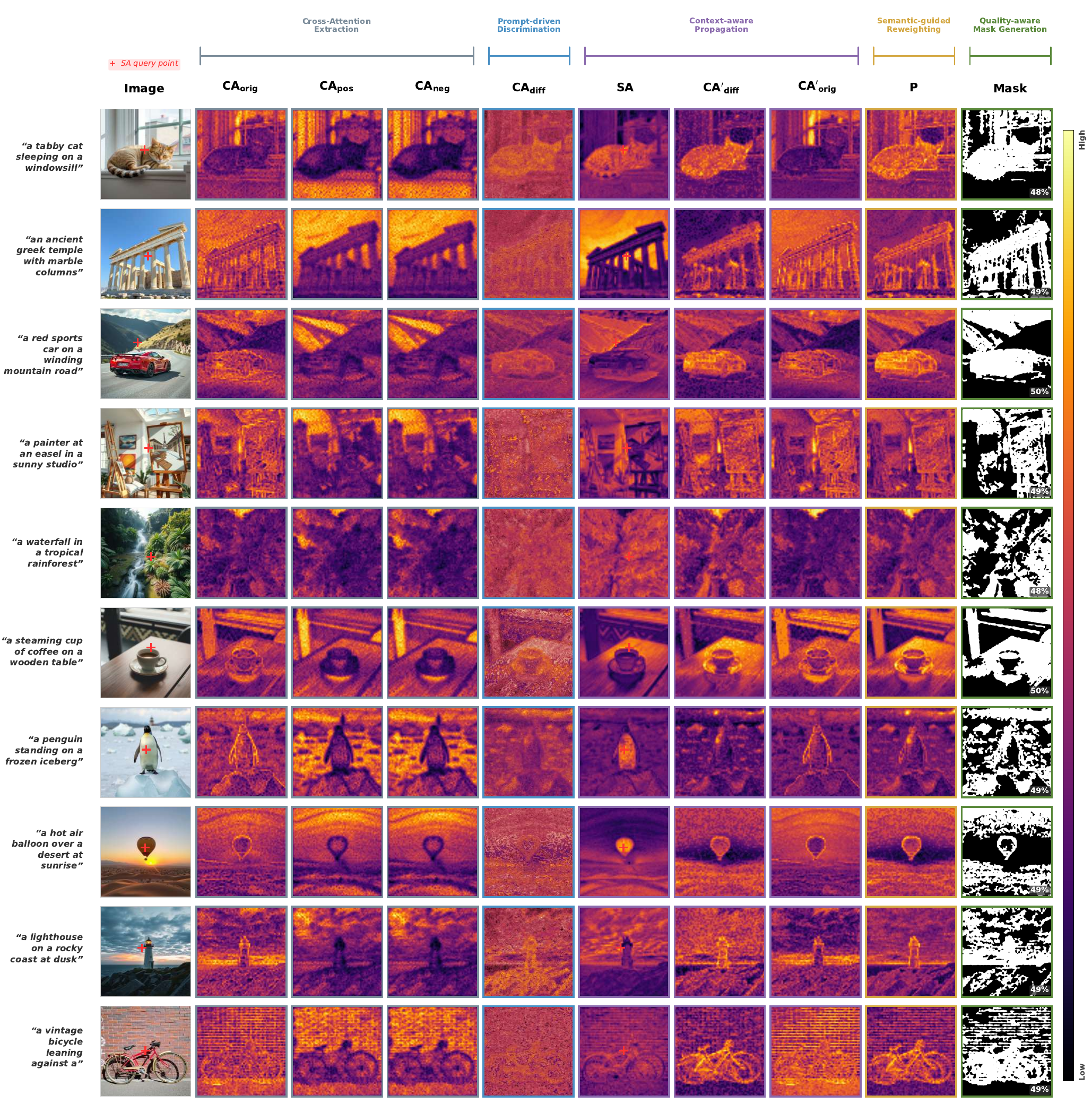}
\vspace{-3mm}
\caption{\textbf{Additional examples of mask generation in FLUX.} LoTTS produces finer and more detailed masks in FLUX, consistently localizing subtle degradations 
under varied prompts and enabling more precise localized resampling.}
\label{fig:additional_mask_examples_flux}
\end{figure*}

Fig.~\ref{fig:additional_mask_examples}, Fig.~\ref{fig:additional_mask_examples_sdxl}, and Fig.~\ref{fig:additional_mask_examples_flux}
provide a detailed visualization of the mask generation process in SD2.1, SDXL, and FLUX.
Across all models, LoTTS exhibits several consistent behaviors that highlight
its ability to identify and isolate low-quality regions.

First, the original cross-attention map ($\mathrm{CA}_{\mathrm{orig}}$) captures
semantic foreground structure but provides limited information about actual
degradations. The positive and negative prompt attention maps ($\mathrm{CA}_{\mathrm{pos}}$,
$\mathrm{CA}_{\mathrm{neg}}$) reveal complementary patterns: $\mathrm{CA}_{\mathrm{pos}}$
tends to emphasize well-formed regions aligned with the prompt, while
$\mathrm{CA}_{\mathrm{neg}}$ highlights visually inconsistent or low-confidence areas.
Taking their contrast produces $\mathrm{CA}_{\mathrm{diff}}$, which reliably exposes
localized artifacts such as missing limbs, distorted geometry, texture noise,
or incorrect shading.

Next, the propagated attention $\mathrm{CA}_{\mathrm{orig}}^{\prime}$ spreads semantic
information through self-attention layers, sharpening object boundaries and
reinforcing regions that are coherent with the global context. When combined
($\mathrm{CA}_{\mathrm{diff}}^{\prime} + \lambda\,\mathrm{CA}_{\mathrm{orig}}^{\prime}$), the resulting
mask becomes both defect-aware and semantically grounded: irrelevant background
areas are suppressed, while problematic regions remain dominant.

As shown in Fig.~\ref{fig:additional_mask_examples}, SD2.1 masks tend to be smoother
and highlight larger defective zones.
SDXL (Fig.~\ref{fig:additional_mask_examples_sdxl}) operates at $1024{\times}1024$ resolution, producing masks that are noticeably sharper and more spatially precise than SD2.1, accurately delineating defect boundaries while retaining strong semantic grounding.
FLUX (Fig.~\ref{fig:additional_mask_examples_flux})
pushes this further, producing the finest-grained masks that capture subtle local degradations (e.g., thin
structure distortions, glass reflections, or geometric irregularities), reflecting its higher
visual capacity.
Across diverse prompts (human bodies, animals, food, vehicles, and indoor scenes),
LoTTS consistently localizes error-prone regions, demonstrating its robustness.

\subsection{Localized Refinement Examples}
\label{app:qual_2}

\begin{figure}[t]
\centering
\includegraphics[width=0.9\linewidth]{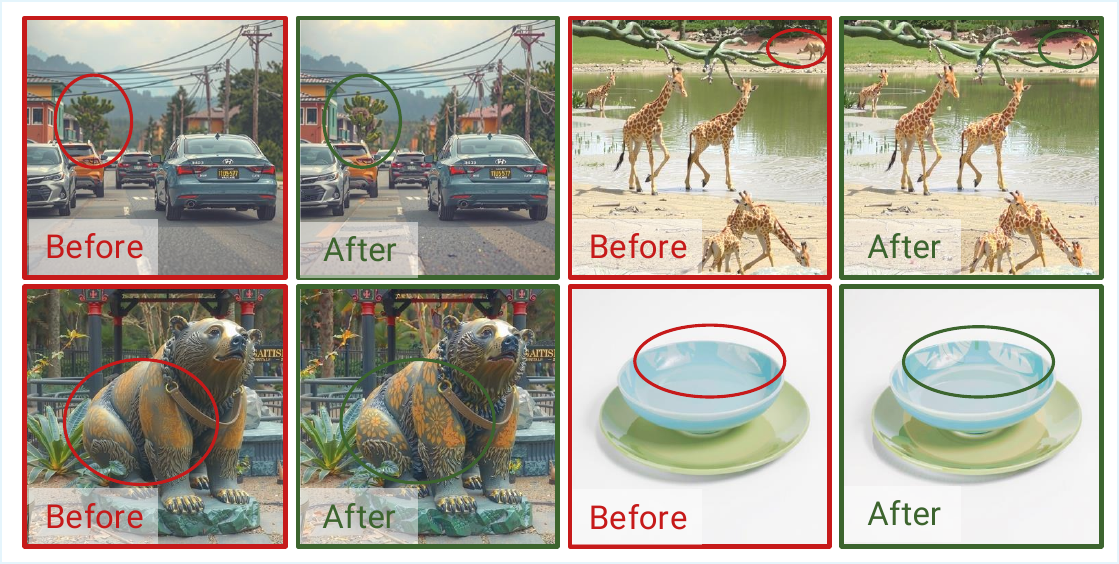}
\vspace{-2mm}
\caption{\textbf{Localized Refinement Examples in FLUX.} LoTTS achieves consistent local refinements across diverse prompts, enhancing structural integrity, semantics, and perceptual quality. The red and green circles highlight corrected regions.}
\label{fig:before_after_comparison_flux}
\end{figure}

Fig.~\ref{fig:before_after_comparison_flux} presents additional before–after
examples from FLUX, illustrating how LoTTS performs targeted corrections on
locally degraded regions while preserving the global layout and scene semantics.
In the upper-left example, a tree in the background appears blurry and lacks
structure; after refinement, LoTTS restores clear branches and edges without
altering the surrounding scene. In the upper-right example, the original image
contains an animal with an implausible and distorted body shape. LoTTS corrects
the malformed geometry, recovering a coherent and recognizable giraffe with
consistent proportions.
The lower-left example highlights LoTTS’s ability to enhance fine textures.
The bear statue originally exhibits smeared and noisy patterns on its torso.
After refinement, LoTTS produces sharper textures and more aesthetically
coherent shading, improving the realism of the material. Finally, in the
lower-right example, the bowl shows washed-out colors and softened boundaries.
LoTTS sharpens the rim, restores the color transitions, and produces a cleaner
overall appearance while keeping the composition unchanged.
Across all cases, the refined results maintain the original structure and
content, modifying only the regions that exhibit degradation. The red and green circles
in Fig.~\ref{fig:before_after_comparison_flux} indicate the specific areas
improved by LoTTS, demonstrating its ability to perform precise, localized
enhancements without introducing unintended global changes.

\begin{table}[t]
\centering
\caption{\textbf{Ablation study of mask generation strategies on Pick-a-Pic and DrawBench.} Results for SD2.1, SDXL, and FLUX show that removing or altering components (e.g., w/o Mask, Random Mask, w/o $\mathrm{CA}_{\mathrm{orig}}^{\prime}$, w/o $\mathrm{CA}_{\mathrm{diff}}^{\prime}$, or w/o Propagation) degrades performance, confirming the effectiveness of our full design.}
\label{tab:mask_ablation}
\resizebox{0.85\textwidth}{!}{%
\begin{tabular}{c lcccc cccc}
\toprule
\multirow{2}{*}{Model} & \multirow{2}{*}{Mask Strategy}
& \multicolumn{4}{c}{Pick-a-Pic} & \multicolumn{4}{c}{DrawBench} \\
\cmidrule(lr){3-6} \cmidrule(lr){7-10}
& & HPS$\uparrow$ & AES$\uparrow$ & Pick$\uparrow$ & IR$\uparrow$
  & HPS$\uparrow$ & AES$\uparrow$ & Pick$\uparrow$ & IR$\uparrow$ \\
\midrule
\multirow{6}{*}{SD2.1}
& w/o Mask         & 21.21 & 5.245 & 21.14 & 0.452 & 22.33 & 5.591 & 20.66 & 0.453 \\
& Random Mask      & 20.59 & 5.325 & 21.01 & 0.457 & 22.34 & 5.580 & 20.67 & 0.452 \\
& w/o $\mathrm{CA}_{\mathrm{orig}}^{\prime}$    & 20.43 & 5.223 & 20.87 & 0.444 & 21.22 & 5.570 & 20.44 & 0.445 \\
& w/o $\mathrm{CA}_{\mathrm{diff}}^{\prime}$         & 21.31 & 5.303 & 20.95 & 0.458 & 22.53 & 5.605 & 21.01 & 0.500 \\
& w/o Propagation       & 23.51 & 5.705 & 21.12 & 0.600 & 22.89 & 5.811 & 21.36 & 0.688 \\
\rowcolor{purple!15}
& \textbf{Ours}    & \textbf{24.52} & \textbf{5.805} & \textbf{21.32} & \textbf{0.680} 
                   & \textbf{23.29} & \textbf{5.911} & \textbf{21.47} & \textbf{0.698} \\
\midrule
\multirow{6}{*}{SDXL}
& w/o Mask         & 24.20 & 6.120 & 22.04 & 0.780 & 25.31 & 6.244 & 22.31 & 0.788 \\
& Random Mask      & 24.19 & 6.115 & 22.01 & 0.755 & 25.65 & 6.231 & 22.32 & 0.778 \\
& w/o $\mathrm{CA}_{\mathrm{orig}}^{\prime}$    & 24.00 & 6.100 & 22.00 & 0.748 & 24.55 & 6.222 & 22.28 & 0.766 \\
& w/o $\mathrm{CA}_{\mathrm{diff}}^{\prime}$         & 26.00 & 6.201 & 22.05 & 0.850 & 26.35 & 6.272 & 22.30 & 0.901 \\
& w/o Propagation       & 28.11 & 6.290 & 22.11 & 1.001 & 28.30 & 6.301 & 22.35 & 1.100 \\
\rowcolor{purple!15}
& \textbf{Ours}    & \textbf{28.23} & \textbf{6.304} & \textbf{22.30} & \textbf{1.102} 
                   & \textbf{28.90} & \textbf{6.321} & \textbf{22.38} & \textbf{1.111} \\
\midrule
\multirow{6}{*}{FLUX}
& w/o Mask         & 30.25 & 6.324 & 22.67 & 1.241 & 31.32 & 6.231 & 22.20 & 1.357 \\
& Random Mask      & 30.15 & 6.313 & 22.55 & 1.211 & 31.23 & 6.225 & 22.18 & 1.344 \\
& w/o $\mathrm{CA}_{\mathrm{orig}}^{\prime}$    & 30.00 & 6.131 & 22.45 & 1.200 & 30.13 & 6.220 & 22.17 & 1.333 \\
& w/o $\mathrm{CA}_{\mathrm{diff}}^{\prime}$         & 31.55 & 6.345 & 22.64 & 1.315 & 32.00 & 6.530 & 22.81 & 1.433 \\
& w/o Propagation       & 33.23 & 6.399 & 22.88 & 1.501 & 32.80 & 6.780 & 23.01 & 1.523 \\
\rowcolor{purple!15}
& \textbf{Ours}    & \textbf{33.33} & \textbf{6.501} & \textbf{23.04} & \textbf{1.605} 
                   & \textbf{33.90} & \textbf{6.890} & \textbf{23.21} & \textbf{1.623} \\
\bottomrule
\end{tabular}}
\end{table}

\section{Ablations}
\label{app:ablation}

\subsection{Mask Generation}
\label{app:mask_generation}

\begin{figure*}[t]
\centering
\includegraphics[width=0.95\linewidth]{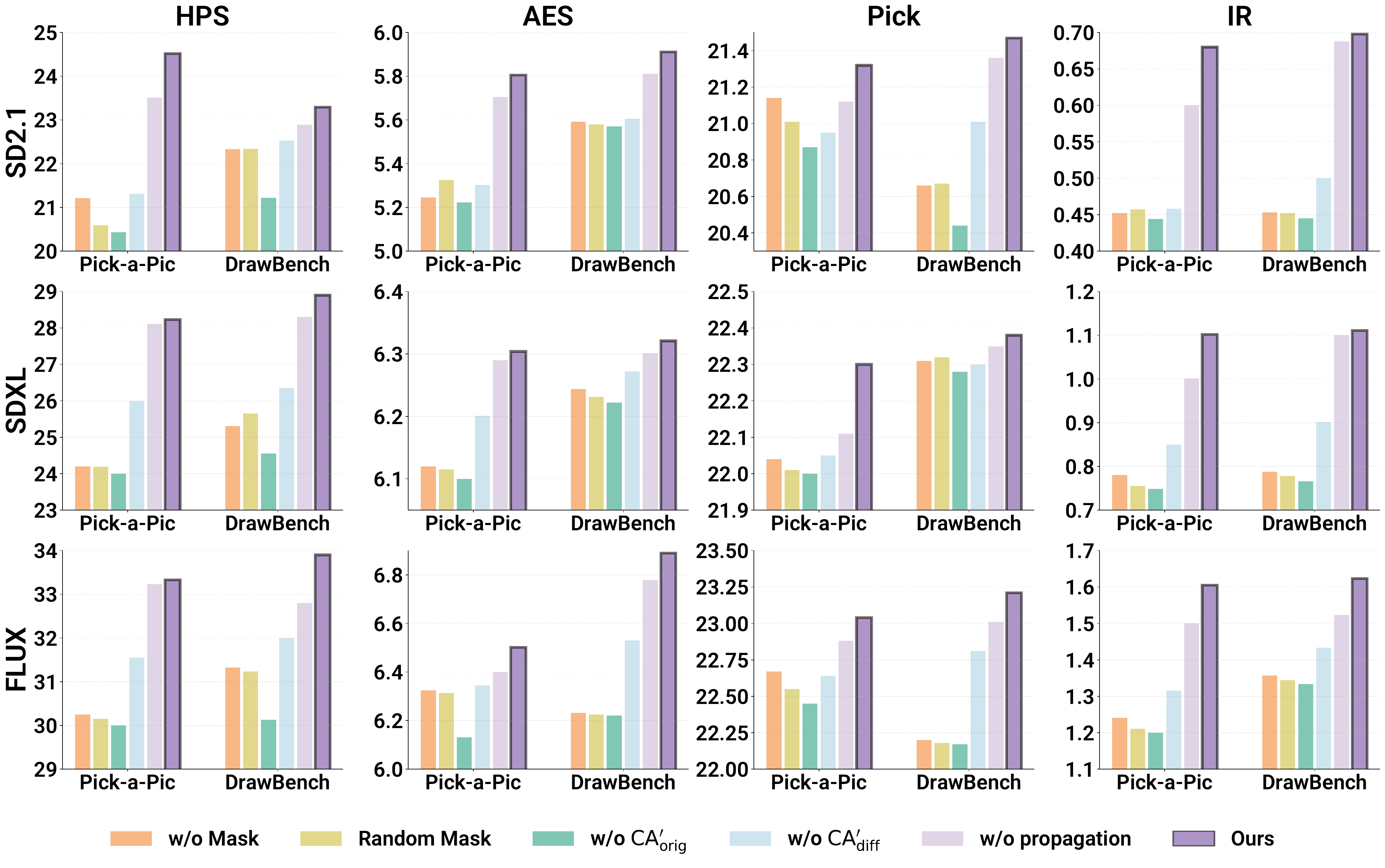}
\vspace{-3mm}
\caption{\textbf{Visualization of mask strategy ablations.} Quantitative results on Pick-a-Pic and DrawBench across SD2.1, SDXL, and FLUX for HPS, AES, Pick, and IR metrics. Compared variants include w/o Mask, Random Mask, w/o $\mathrm{CA}_{\mathrm{orig}}^{\prime}$, w/o $\mathrm{CA}_{\mathrm{diff}}^{\prime}$, and w/o Propagation. LoTTS consistently outperforms these ablated versions, highlighting the importance of each mask generation component.}
\label{fig:2_bar_all}
\end{figure*}

We conduct detailed ablations to understand the role of each component in our
mask-generation pipeline. Tab.~\ref{tab:mask_ablation} and
Fig.~\ref{fig:2_bar_all} summarize results on SD2.1, SDXL, and FLUX. Across
all settings, modifying or removing any component degrades performance, but the
severity of degradation follows a clear pattern that reflects how each factor
affects mask quality and refinement behavior.

\myparagraph{Worst case: removing $\mathrm{CA}_{\mathrm{orig}}^{\prime}$.}
The largest drop occurs when $\mathrm{CA}_{\mathrm{orig}}^{\prime}$ is removed.
Without the original cross-attention prior, the mask loses its semantic
foreground constraint and often activates on background regions rather than
meaningful objects. This severely reduces mask precision and leads the model to
refine irrelevant areas, producing the lowest scores across nearly all metrics.
This confirms that $\mathrm{CA}_{\mathrm{orig}}^{\prime}$ is essential for
locating semantically valid regions where refinement should be applied.

\myparagraph{Random mask: too fragmented and semantically inconsistent.}
The second-worst performance comes from using a random mask. Unlike the
``w/o Mask'' baseline, the random mask still enforces locality, but it does so
in an unstructured and highly fragmented manner. Because foreground and
background regions are randomly mixed, the mask fails to form coherent spatial
blocks, making refinement unstable and preventing the model from effectively
correcting artifacts. This produces masks with both low precision and low
recall, explaining why the random baseline consistently underperforms.

\myparagraph{No mask: global resampling is unnecessarily destructive.}
Removing the mask (``w/o Mask'') yields the next level of degradation. In this
case, refinement becomes global: the model resamples the entire image,
including regions that were already high quality. Although this avoids the
fragmentation issue of random masks, the lack of locality causes semantic drift
and over-editing, which lowers HPS, AES, and Pick metrics. This demonstrates
that locality is crucial for preserving global semantics while improving
specific defective regions.

\myparagraph{Removing $\mathrm{CA}_{\mathrm{diff}}^{\prime}$: no defect-specific cue.}
Mask quality improves when keeping $\mathrm{CA}_{\mathrm{orig}}^{\prime}$ while
removing $\mathrm{CA}_{\mathrm{diff}}^{\prime}$. However, without the differential
signal, the mask becomes insensitive to quality differences and focuses only on
semantic saliency. As a result, some artifacts are missed, lowering recall
and reducing the improvement rate. This explains why performance is better than
random or no-mask baselines, but still clearly below our full design.

\myparagraph{Removing Propagation: masks lose spatial continuity.}
The best-performing ablation variant is ``w/o Propagation''. The mask still
captures the correct semantic and defective regions due to the two
cross-attention signals, but lacks spatial smoothness. As a result, the mask
becomes fragmented and refinement occurs in slightly noisy regions. Performance
drops moderately, indicating that Propagation improves the spatial stability of
the refinement.

\myparagraph{Summary.}
Across all variants, the degradation trend follows:
\[
\begin{aligned}
\text{w/o } \mathrm{CA}_{\mathrm{orig}}^{\prime} 
&< \text{Random Mask} 
< \text{w/o Mask} \\
&< \text{w/o } \mathrm{CA}_{\mathrm{diff}}^{\prime} 
< \text{w/o Propagation} 
< \text{Ours}.
\end{aligned}
\]
This ordering reflects the relative contributions of semantic grounding from
$\mathrm{CA}_{\mathrm{orig}}^{\prime}$, defect sensitivity from
$\mathrm{CA}_{\mathrm{diff}}^{\prime}$, and spatial coherence from
Propagation. Together, these factors enable LoTTS to produce precise, contiguous,
and defect-aware masks that yield the strongest overall refinement performance.

\subsection{Prompt Design}
\label{app:prompt_design}

As shown in Tab.~\ref{tab:prompt_ablation_1}, this ablation investigates how different prompt formulations influence the effectiveness of LoTTS. 
We compare two strategies: (1) ``A good/bad image of \{p\}'', and (2) ``A high/low-quality image of \{p\}'', 
where \{p\} denotes the original text prompt. 
These two designs differ subtly in semantics: the former focuses on the overall desirability of the image, 
while the latter emphasizes rendering fidelity and fine-grained visual quality.
Across all three model families (SD2.1, SDXL, and FLUX), both prompt strategies consistently improve generation quality, 
with only mild differences between them. 
For SD2.1 and FLUX, the two prompts perform similarly across all metrics, 
with the ``high/low-quality'' formulation yielding slightly higher Pick and IR scores. 
For SDXL, however, the ``good/bad'' prompt generally achieves better HPS and AES on Pick-a-Pic, 
suggesting that SDXL is more responsive to value-oriented descriptors than to explicit quality modifiers.
Overall, the results indicate that LoTTS is robust to the specific choice of prompt template: 
both designs effectively provide the contrastive supervision needed for preference-aware optimization. 
The observed differences are small and model-specific, implying mild lexical sensitivity but no strong dependence on prompt wording.

\begin{table}[t]
\centering
\caption{\textbf{Ablation on Prompt Strategies for LoTTS.}
We evaluate two prompt designs, where $i{=}0$ corresponds to 
``A good/bad image of \{p\}'', and $i{=}1$ corresponds to 
``A high/low-quality image of \{p\}'', where \{p\} denotes the original text prompt. 
Results are reported on Pick-a-Pic and DrawBench for SD2.1, SDXL, and FLUX.}
\label{tab:prompt_ablation_1}
\resizebox{0.75\linewidth}{!}{%
\begin{tabular}{c ccccc cccc}
\toprule
\multirow{2}{*}{Model} & \multirow{2}{*}{${i}$} 
& \multicolumn{4}{c}{Pick-a-Pic} & \multicolumn{4}{c}{DrawBench} \\
\cmidrule(lr){3-6} \cmidrule(lr){7-10}
& & HPS$\uparrow$ & AES$\uparrow$ & Pick$\uparrow$ & IR$\uparrow$
  & HPS$\uparrow$ & AES$\uparrow$ & Pick$\uparrow$ & IR$\uparrow$ \\
\midrule
\multirow{2}{*}{SD2.1}  
& 0 & 24.23 & \textbf{5.821} & 21.31 & 0.679 & 23.27 & \textbf{5.923} & 21.33 & \textbf{0.701} \\
& 1 & \textbf{24.52} & 5.805 & \textbf{21.32} & \textbf{0.680} & \textbf{23.29} & 5.911 & \textbf{21.47} & 0.698 \\
\midrule
\multirow{2}{*}{SDXL}  
& 0 & \textbf{28.45} & \textbf{6.312} & 22.27 & \textbf{1.105} & 28.89 & \textbf{6.318} & 22.21 & \textbf{1.131} \\
& 1 & 28.23 & 6.304 & \textbf{22.30} & 1.102 & \textbf{28.90} & 6.321 & \textbf{22.38} & 1.111 \\
\midrule
\multirow{2}{*}{FLUX}  
& 0 & 33.11 & \textbf{6.505} & 23.00 & 1.604 & 33.78 & 6.889 & 23.15 & 1.620 \\
& 1 & \textbf{33.33} & 6.501 & \textbf{23.04} & \textbf{1.605} & \textbf{33.90} & \textbf{6.890} & \textbf{23.21} & \textbf{1.623} \\
\bottomrule
\end{tabular}}
\end{table}

\subsection{Attention Reweighting Coefficient \texorpdfstring{$\lambda$}{lambda}}
\label{app:lambda_ablation}

We ablate the semantic-guided reweighting coefficient $\lambda$ in Eq.~\eqref{eq:Semantic_guided_Reweighting}, which balances the contrastive defect signal $\mathrm{CA}'_{\mathrm{diff}}$ and the original foreground prior $\mathrm{CA}'_{\mathrm{orig}}$.
Tab.~\ref{tab:lambda_ablation} reports results across SD2.1, SDXL, and FLUX on Pick-a-Pic and DrawBench.

When $\lambda{=}0$, the mask relies solely on the contrastive signal and loses its semantic foreground constraint, often activating on background regions. This is functionally equivalent to the ``w/o $\mathrm{CA}_{\mathrm{orig}}^{\prime}$'' variant in Tab.~\ref{tab:mask_ablation} and yields the weakest performance across all models and metrics.
Performance peaks at $\lambda{=}0.5$ across all three models, striking the best balance between defect sensitivity and foreground grounding.
When $\lambda$ is further increased to $1.0$, the foreground prior begins to dominate, causing the mask to spread across the entire foreground rather than focusing on defective regions. This leads to a modest but consistent decline: for example, SD2.1 HPS on Pick-a-Pic drops from 24.52 ($\lambda{=}0.5$) to 24.01 ($\lambda{=}1.0$), while SDXL IR on DrawBench decreases from 1.111 to 1.055.
Notably, $\lambda{=}1.0$ still significantly outperforms $\lambda{=}0$ (e.g., SD2.1 HPS: 24.01 vs.\ 20.43), confirming that even an over-weighted foreground prior is far more beneficial than having none at all.
Overall, the results demonstrate that LoTTS is robust to the choice of $\lambda$, with the degradation from the optimal $\lambda{=}0.5$ to $\lambda{=}1.0$ being much smaller than the gap between $\lambda{=}0$ and $\lambda{=}0.5$.

\begin{table}[t]
\centering
\caption{\textbf{Ablation on attention reweighting coefficient $\lambda$.} Results on Pick-a-Pic and DrawBench for SD2.1, SDXL, and FLUX.}
\label{tab:lambda_ablation}
\resizebox{0.75\textwidth}{!}{%
\begin{tabular}{c c cccc cccc}
\toprule
\multirow{2}{*}{Model} & \multirow{2}{*}{$\lambda$} 
& \multicolumn{4}{c}{Pick-a-Pic} & \multicolumn{4}{c}{DrawBench} \\
\cmidrule(lr){3-6} \cmidrule(lr){7-10}
& & HPS$\uparrow$ & AES$\uparrow$ & Pick$\uparrow$ & IR$\uparrow$
  & HPS$\uparrow$ & AES$\uparrow$ & Pick$\uparrow$ & IR$\uparrow$ \\
\midrule
\multirow{3}{*}{SD2.1}  
& 0    & 20.43 & 5.223 & 20.87 & 0.444 & 21.22 & 5.570 & 20.44 & 0.445 \\
& 0.50 & \textbf{24.52} & \textbf{5.805} & \textbf{21.32} & \textbf{0.680} & \textbf{23.29} & \textbf{5.911} & \textbf{21.47} & \textbf{0.698} \\
& 1.00 & 24.01 & 5.799 & 21.29 & 0.643 & 23.10 & 5.835 & 21.33 & 0.655 \\
\midrule
\multirow{3}{*}{SDXL}  
& 0    & 24.00 & 6.100 & 22.00 & 0.748 & 24.55 & 6.222 & 22.28 & 0.766 \\
& 0.50 & \textbf{28.23} & \textbf{6.304} & \textbf{22.30} & \textbf{1.102} & \textbf{28.90} & \textbf{6.321} & \textbf{22.38} & \textbf{1.111} \\
& 1.00 & 27.56 & 6.299 & 22.22 & 1.023 & 26.55 & 6.300 & 22.30 & 1.055 \\
\midrule
\multirow{3}{*}{FLUX}  
& 0    & 30.00 & 6.131 & 22.45 & 1.200 & 30.13 & 6.220 & 22.17 & 1.333 \\
& 0.50 & \textbf{33.33} & \textbf{6.501} & \textbf{23.04} & \textbf{1.605} & \textbf{33.90} & \textbf{6.890} & \textbf{23.21} & \textbf{1.623} \\
& 1.00 & 32.66 & 6.449 & 22.89 & 1.599 & 33.01 & 6.655 & 23.01 & 1.578 \\
\bottomrule
\end{tabular}}
\end{table}

\subsection{Consistency Components}
\label{app:consistency}

We ablate the three consistency mechanisms introduced in Sec.~\ref{sec:method} to evaluate their individual contributions to refinement quality.
Tab.~\ref{tab:consistency_ablation} reports results for SD2.1, SDXL, and FLUX on Pick-a-Pic and DrawBench.

\myparagraph{Temporal Consistency (largest degradation).}
Removing temporal consistency means starting the refinement from pure noise ($t{=}T$) rather than the intermediate step $t_0$. This forces the model to regenerate the masked region from scratch, destroying the global semantic context and causing the refined content to be semantically inconsistent with the surrounding image. This variant shows the largest degradation across all models: for example, SD2.1 HPS drops from 24.52 to 20.33 on Pick-a-Pic and IR drops from 0.680 to 0.235, falling back to the level of a single unrefined generation. The pattern is consistent for SDXL and FLUX, confirming that preserving global structure via partial denoising is critical.

\myparagraph{Spatial Consistency (substantial degradation).}
Removing spatial consistency means injecting noise only within the mask while keeping the background latent clean at step $t_0$. This creates a discontinuous noise distribution at mask boundaries, producing visible seams and edge artifacts. The resulting quality drop is substantial across all metrics, with HPS and IR suffering the most (e.g., SDXL HPS drops from 28.23 to 24.99, IR drops from 1.102 to 0.894 on Pick-a-Pic), as boundary artifacts degrade both perceptual preference and overall image quality.

\myparagraph{Holistic Consistency (moderate degradation).}
Removing holistic consistency means skipping the global integration phase (setting $t_g{=}0$, i.e., no unmasked denoising steps after refinement). Without these final harmonization steps, subtle boundary inconsistencies in lighting, color, and texture persist. The degradation is the smallest among the three components but remains consistent: for instance, SD2.1 IR drops from 0.680 to 0.650 on Pick-a-Pic. This demonstrates that even a few global denoising steps effectively smooth residual artifacts and contribute to final quality.

\begin{table}[t]
\centering
\caption{\textbf{Ablation on consistency components.} Results on Pick-a-Pic and DrawBench for SD2.1, SDXL, and FLUX.}
\label{tab:consistency_ablation}
\resizebox{0.85\textwidth}{!}{%
\begin{tabular}{c lcccc cccc}
\toprule
\multirow{2}{*}{Model} & \multirow{2}{*}{Variant}
& \multicolumn{4}{c}{Pick-a-Pic} & \multicolumn{4}{c}{DrawBench} \\
\cmidrule(lr){3-6} \cmidrule(lr){7-10}
& & HPS$\uparrow$ & AES$\uparrow$ & Pick$\uparrow$ & IR$\uparrow$
  & HPS$\uparrow$ & AES$\uparrow$ & Pick$\uparrow$ & IR$\uparrow$ \\
\midrule
\multirow{4}{*}{SD2.1}
& w/o Temporal    & 20.33 & 5.298 & 20.24 & 0.235 & 21.32 & 5.423 & 20.21 & 0.240 \\
& w/o Spatial   & 21.36 & 5.390 & 20.69 & 0.388 & 21.93 & 5.584 & 20.58 & 0.288 \\
& w/o Holistic   & 24.24 & 5.799 & 21.22 & 0.650 & 23.09 & 5.899 & 21.40 & 0.687 \\
\rowcolor{purple!15}
& \textbf{Ours}  & \textbf{24.52} & \textbf{5.805} & \textbf{21.32} & \textbf{0.680} & \textbf{23.29} & \textbf{5.911} & \textbf{21.47} & \textbf{0.698} \\
\midrule
\multirow{4}{*}{SDXL}
& w/o Temporal    & 24.21 & 6.178 & 21.89 & 0.789 & 25.01 & 6.220 & 22.01 & 0.742 \\
& w/o Spatial   & 24.99 & 6.225 & 21.94 & 0.894 & 25.55 & 6.256 & 22.45 & 0.789 \\
& w/o Holistic   & 28.05 & 6.299 & 22.18 & 1.008 & 28.59 & 6.305 & 22.35 & 1.105 \\
\rowcolor{purple!15}
& \textbf{Ours}  & \textbf{28.23} & \textbf{6.304} & \textbf{22.30} & \textbf{1.102} & \textbf{28.90} & \textbf{6.321} & \textbf{22.38} & \textbf{1.111} \\
\midrule
\multirow{4}{*}{FLUX}
& w/o Temporal    & 30.12 & 6.288 & 22.01 & 1.030 & 29.19 & 6.215 & 22.02 & 1.084 \\
& w/o Spatial   & 30.89 & 6.301 & 22.33 & 1.088 & 29.58 & 6.225 & 22.56 & 1.146 \\
& w/o Holistic   & 33.01 & 6.488 & 23.03 & 1.599 & 33.49 & 6.823 & 23.11 & 1.598 \\
\rowcolor{purple!15}
& \textbf{Ours}  & \textbf{33.33} & \textbf{6.501} & \textbf{23.04} & \textbf{1.605} & \textbf{33.90} & \textbf{6.890} & \textbf{23.21} & \textbf{1.623} \\
\bottomrule
\end{tabular}}
\end{table}

\section{Parameter Analysis}
\label{app:parameter_analysis}

\subsection{Hyperparameter Sensitivity}
\label{app:parameter_sensitivity}

\begin{figure*}[ht]
\centering
\includegraphics[width=0.95\textwidth]{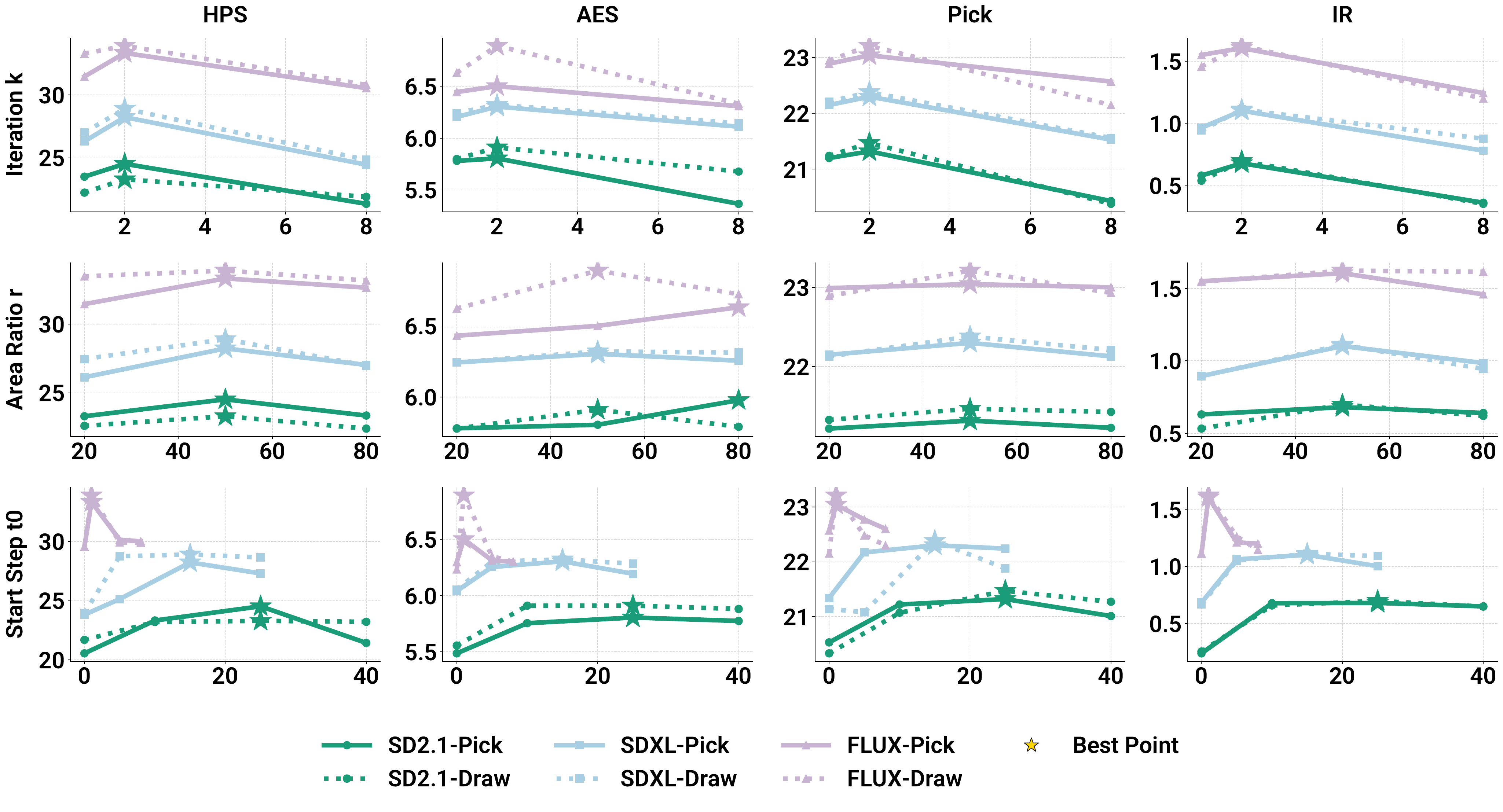}
\vspace{-5pt}
\caption{\textbf{Hyperparameter analysis of LoTTS.} 
Performance on Pick-a-Pic and DrawBench with respect to (top) 
the number of refinements $k$, (middle) mask area ratio $r$, 
and (bottom) noise injection step $t_0$, across SD2.1, SDXL, and FLUX. 
Stars mark the best-performing configurations.}
\label{fig:2_parameters}
\end{figure*}

We analyze the sensitivity of LoTTS to its key hyperparameters: number of
refinements $k$, mask area ratio $r$, and noise injection step $t_0$.
Tab.~\ref{tab:quantitative_k}, Tab.~\ref{tab:quantitative_r}, Tab.~\ref{tab:quantitative_t0} and
Fig.~\ref{fig:2_parameters} summarize the results.

\myparagraph{Number of refinements $k$.}
Across SD2.1, SDXL, and FLUX, the best performance consistently occurs at
$k=2$. In all experiments, we fix the final number of images per prompt at $N=9$,
so changing $k$ redistributes this fixed sampling budget between the number of
distinct initial samples and the number of local refinements applied to each of
them.
When $k$ increases, more budget is spent on local resampling around each
starting point, but fewer distinct starting points are available. Although each
local refinement is stochastic, the local generation space around a fixed
starting point is inherently limited: additional local samples do not provide
substantially better refinement candidates. In contrast, reducing the number of
initial samples significantly weakens global diversity, which is often very
important for discovering high-quality candidates under a fixed budget.
With $k=2$, this balance is optimal: there are enough distinct initial samples
to maintain global diversity, and enough local variations to explore meaningful
refinement modes. In comparison, $k=1$ offers insufficient local exploration,
while $k=8$ collapses global diversity by relying on a single starting point,
leading to notably worse Best-of-$9$ performance. Thus, $k=2$ provides the most
effective trade-off under a fixed sampling budget.

\begin{table}[t]
\centering
\caption{\textbf{Quantitative results w.r.t. number of refinements $k$.} Results on Pick-a-Pic and DrawBench for SD2.1, SDXL, and FLUX.}
\label{tab:quantitative_k}
\resizebox{0.75\textwidth}{!}{%
\begin{tabular}{c c cccc cccc}
\toprule
\multirow{2}{*}{Model} & \multirow{2}{*}{$k$} 
& \multicolumn{4}{c}{Pick-a-Pic} & \multicolumn{4}{c}{DrawBench} \\
\cmidrule(lr){3-6} \cmidrule(lr){7-10}
& & HPS$\uparrow$ & AES$\uparrow$ & Pick$\uparrow$ & IR$\uparrow$
  & HPS$\uparrow$ & AES$\uparrow$ & Pick$\uparrow$ & IR$\uparrow$ \\
\midrule
\multirow{3}{*}{SD2.1}  
& 1 & 23.50 & 5.781 & 21.20 & 0.581 & 22.23 & 5.799 & 21.24 & 0.540 \\
      & 2 & \textbf{24.52} & \textbf{5.805} & \textbf{21.32} & \textbf{0.680} & \textbf{23.29} & \textbf{5.911} & \textbf{21.47} & \textbf{0.698} \\
      & 8 & 21.34 & 5.367 & 20.43 & 0.363 & 21.90 & 5.678 & 20.38 & 0.354 \\
\midrule
\multirow{3}{*}{SDXL}  
& 1 & 26.31 & 6.205 & 22.15 & 0.965 & 27.01 & 6.243 & 22.21 & 0.944 \\
      & 2 & \textbf{28.23} & \textbf{6.304} & \textbf{22.30} & \textbf{1.102} & \textbf{28.90} & \textbf{6.321} & \textbf{22.38} & \textbf{1.111} \\
      & 8 & 24.45 & 6.111 & 21.53 & 0.781 & 24.84 & 6.144 & 21.56 & 0.877 \\
\midrule
\multirow{3}{*}{FLUX}  
& 1 & 31.46 & 6.445 & 22.89 & 1.550 & 33.28 & 6.631 & 22.95 & 1.457 \\
& 2 & \textbf{33.33} & \textbf{6.501} & \textbf{23.04} & \textbf{1.605} & \textbf{33.90} & \textbf{6.890} & \textbf{23.21} & \textbf{1.623} \\
& 8 & 30.54 & 6.308 & 22.57 & 1.244 & 30.81 & 6.332 & 22.15 & 1.200 \\
\bottomrule
\end{tabular}}
\end{table}

\myparagraph{Mask area ratio $r$.}
Tab.~\ref{tab:quantitative_r} shows that a moderate mask area ratio (about $50\%$) consistently yields the best performance across SD2.1,
SDXL, and FLUX. This trend reflects a fundamental trade-off between defect
coverage and structural preservation. When the ratio is small (e.g., $r=20\%$),
the mask typically captures only the most salient defect regions while ignoring
secondary or subtle artifacts. As a result, many imperfections remain
unrefined, lowering both the improvement rate (IR) and aesthetic gains (HPS,
AES). This under-refinement behavior is evident in SD2.1, SDXL, and FLUX, where
$r=20\%$ systematically underperforms the $50\%$ setting across nearly all
metrics.
Increasing the ratio to a moderate level ($r=50\%$) substantially improves all
metrics. At this setting, the mask covers not only primary defects but also
neighboring regions that contribute to local inconsistency, enabling more
comprehensive correction. This leads to strong gains in HPS, AES, and IR across
all three models. 
However, when the mask becomes too large ($r=80\%$), performance begins to decline. Excessively large masks start to overwrite high-
quality regions, particularly detailed textures or well-formed structures,
making the refinement process resemble global resampling. 
Interestingly, on Pick-a-Pic, FLUX obtains the highest AES at $r=80\%$, but its HPS, Pick, and
IR metrics do not surpass the $50\%$ setting, indicating that although aesthetic
smoothness may improve, semantic fidelity and user preference do not benefit
from excessive mask coverage.
Overall, the optimal range around $50\%$ aligns with the observation that
localized refinement should balance two competing goals: (1) covering enough
area to correct both major and subtle defects, while (2) preserving globally
coherent structure and avoiding unnecessary alteration of high-quality content.
This trade-off consistently favors a moderate ratio across all tested models and
datasets.

\begin{table}[t]
\centering
\caption{\textbf{Quantitative results w.r.t. mask area ratio $r$.} Results on Pick-a-Pic and DrawBench for SD2.1, SDXL, and FLUX.}
\label{tab:quantitative_r}
\resizebox{0.75\textwidth}{!}{%
\begin{tabular}{c c cccc cccc}
\toprule
\multirow{2}{*}{Model} & \multirow{2}{*}{$r$} 
& \multicolumn{4}{c}{Pick-a-Pic} & \multicolumn{4}{c}{DrawBench} \\
\cmidrule(lr){3-6} \cmidrule(lr){7-10}
& & HPS$\uparrow$ & AES$\uparrow$ & Pick$\uparrow$ & IR$\uparrow$
  & HPS$\uparrow$ & AES$\uparrow$ & Pick$\uparrow$ & IR$\uparrow$ \\
\midrule
\multirow{3}{*}{SD2.1}
& 20 & 23.29 & 5.780 & 21.22 & 0.630 & 22.59 & 5.778 & 21.33 & 0.532 \\
& 50 & \textbf{24.52} & 5.805 & \textbf{21.32} & \textbf{0.680} & \textbf{23.29} & \textbf{5.911} & \textbf{21.47} & \textbf{0.698} \\
      & 80 & 23.34 & \textbf{5.978} & 21.23 & 0.640 & 22.39 & 5.791 & 21.43 & 0.620 \\
\midrule
\multirow{3}{*}{SDXL}
& 20 & 26.13 & 6.244 & 22.15 & 0.895 & 27.46 & 6.243 & 22.13 & 0.894 \\
& 50 & \textbf{28.23} & \textbf{6.304} & \textbf{22.30} & \textbf{1.102} & \textbf{28.90} & \textbf{6.321} & \textbf{22.38} & \textbf{1.111} \\
& 80 & 27.03 & 6.256 & 22.13 & 0.985 & 26.98 & 6.313 & 22.21 & 0.944 \\
\midrule
\multirow{3}{*}{FLUX}
& 20 & 31.46 & 6.432 & 22.99 & 1.550 & 33.48 & 6.623 & 22.89 & 1.545 \\
& 50 & \textbf{33.33} & 6.501 & \textbf{23.04} & \textbf{1.605} & \textbf{33.90} & \textbf{6.890} & \textbf{23.21} & \textbf{1.623} \\
& 80 & 32.66 & \textbf{6.632} & 23.00 & 1.459 & 33.18 & 6.724 & 22.93 & 1.615 \\
\bottomrule
\end{tabular}}
\end{table}

\myparagraph{Noise injection step $t_0$.}
Tab.~\ref{tab:quantitative_t0} shows that the optimal noise injection step
varies noticeably across SD2.1, SDXL, and FLUX, reflecting the different
generative dynamics of their underlying architectures. For the U-Net–based
diffusion models SD2.1 and SDXL, injecting noise at an early-to-mid denoising
step produces the strongest results. Specifically, SD2.1 achieves its best
performance at $t_0=25$ out of 50 steps, while SDXL peaks at $t_0=15$ out of
30 steps. At these intermediate stages, the model retains enough semantic
structure from the original sample to preserve the global layout while still
having sufficient freedom to update defect regions. This balance leads to clear
gains in HPS, AES, and IR. The results in Tab.~\ref{tab:quantitative_t0}
further show that both too-early injection ($t_0=0$) and too-late injection
($t_0=40$ for SD2.1, $t_0=25$ for SDXL) result in inferior performance:
injecting noise too early limits the model's ability to make meaningful local corrections, while injecting too late disrupts global structure.
In contrast, FLUX exhibits a markedly different pattern due to its rectified
flow architecture. Unlike U-Net diffusion, FLUX rapidly forms stable global
structures at the beginning of its generation trajectory. As a result, even an
extremely small noise step such as $t_0=1$ provides enough flexibility to refine
local defects without disturbing the overall composition. This is reflected in
its consistently strong performance at $t_0=1$, achieving the best HPS, AES,
Pick, and IR across both datasets. Larger injection steps, however, degrade
performance: injecting noise at $t_0=5$ or $8$ overwrites too much of the
already-converged global structure, causing the refined image to deviate from
the intended semantics.
Taken together, these observations highlight that the optimal noise injection
stage is tightly coupled to the generative process of each architecture.
U-Net–based diffusion models favor early-to-mid stages where both structure and
flexibility coexist, whereas rectified flow models like FLUX require only a
minimal noise injection due to their fast convergence properties. LoTTS
naturally adapts to these dynamics by allowing $t_0$ to control the balance
between preserving global coherence and enabling local refinement.

\begin{table}[t]
\centering
\caption{\textbf{Quantitative results w.r.t. noise injection step $t_0$.} Results on Pick-a-Pic and DrawBench for SD2.1, SDXL, and FLUX.}
\label{tab:quantitative_t0}
\resizebox{0.75\textwidth}{!}{%
\begin{tabular}{c c cccc cccc}
\toprule
\multirow{2}{*}{Model} & \multirow{2}{*}{$t_0$} 
& \multicolumn{4}{c}{Pick-a-Pic} & \multicolumn{4}{c}{DrawBench} \\
\cmidrule(lr){3-6} \cmidrule(lr){7-10}
& & HPS$\uparrow$ & AES$\uparrow$ & Pick$\uparrow$ & IR$\uparrow$
  & HPS$\uparrow$ & AES$\uparrow$ & Pick$\uparrow$ & IR$\uparrow$ \\
\midrule
\multirow{4}{*}{SD2.1} 
& 0  & 20.54 & 5.487 & 20.53 & 0.238 & 21.68 & 5.556 & 20.33 & 0.254 \\
& 10 & 23.32 & 5.755 & 21.22 & 0.678 & 23.19 & 5.910 & 21.07 & 0.658 \\
& 25 & \textbf{24.52} & \textbf{5.805} & \textbf{21.32} & \textbf{0.680} & \textbf{23.29} & \textbf{5.911} & \textbf{21.47} & \textbf{0.698} \\
& 40 & 21.42 & 5.775 & 21.01 & 0.650 & 23.21 & 5.881 & 21.27 & 0.648 \\
\midrule
\multirow{4}{*}{SDXL} 
& 0  & 23.81 & 6.041 & 21.34 & 0.683 & 23.95 & 6.054 & 21.14 & 0.667 \\
& 5  & 25.13 & 6.254 & 22.17 & 1.062 & 28.75 & 6.301 & 21.08 & 1.051 \\
& 15 & \textbf{28.23} & \textbf{6.304} & \textbf{22.30} & \textbf{1.102} & \textbf{28.90} & \textbf{6.321} & \textbf{22.38} & \textbf{1.111} \\
& 25 & 27.30 & 6.194 & 22.24 & 1.002 & 28.65 & 6.283 & 21.88 & 1.091 \\
\midrule
\multirow{4}{*}{FLUX} 
& 0  & 29.55 & 6.298 & 22.57 & 1.111 & 29.59 & 6.233 & 22.15 & 1.117 \\
& 1  & \textbf{33.33} & \textbf{6.501} & \textbf{23.04} & \textbf{1.605} & \textbf{33.90} & \textbf{6.890} & \textbf{23.21} & \textbf{1.623} \\
& 5  & 30.15 & 6.312 & 22.77 & 1.211 & 29.93 & 6.349 & 22.48 & 1.263 \\
& 8  & 30.01 & 6.301 & 22.60 & 1.200 & 29.88 & 6.301 & 22.30 & 1.145 \\
\bottomrule
\end{tabular}}
\end{table}

\myparagraph{Visualization.}
Fig.~\ref{fig:2_parameters} visualizes the behavior of LoTTS under different
hyperparameter settings. Across all models and metrics, the curves vary smoothly
and exhibit clear optima, indicating that LoTTS is stable and well-conditioned.
The best-performing points form broad, well-defined basins rather than isolated
spikes, showing that LoTTS does not rely on fragile hyperparameter choices and
is robust across a wide operating range.

\subsection{Scaling Comparison with Best-of-\texorpdfstring{$N$}{N}}
\label{app:scaling_comparison}

We evaluate how LoTTS scales with the total number of generated images $N$ and
compare it against a pure global Best-of-$N$ baseline. In both cases, $N$
represents the total number of images generated per prompt. For Best-of-$N$,
this is simply the number of globally sampled candidates. For LoTTS, however,
the same budget $N$ is redistributed between the number of initial samples and
the number of local refinements performed around each initial sample. For
example, $N=1$ corresponds to pure global sampling, $N=3$ corresponds to $1$ initial sample producing $1+k=3$ total images (the original plus $k{=}2$ local refinements), $N=6$ corresponds to $2$
initial samples each producing $1+k=3$ images, and $N=9$ corresponds to the
default LoTTS setting of $3$ initial samples each producing $1+k=3$ images.

\myparagraph{Scaling behavior of LoTTS.}
Tab.~\ref{tab:lotts_scaling} shows that LoTTS exhibits overall consistent
performance gains as the total number of generated images $N$ increases from 1
to 9. Even with extremely small budgets, such as $N=3$, LoTTS already achieves
notable improvements over pure global sampling ($N=1$). For example, on SD2.1,
HPS increases from 20.44 at $N=1$ to 21.12 at $N=3$, accompanied by an IR gain
from 0.236 to 0.413. As $N$ continues increasing, these gains become more
substantial: at $N=6$, SD2.1 reaches HPS 22.29 and IR 0.531; and at $N=9$,
LoTTS achieves its strongest performance with HPS 24.52 and IR 0.680. A similar pattern appears in SDXL and FLUX. On SDXL, HPS improves from 23.44
($N=1$) → 25.23 ($N=3$) → 26.21 ($N=6$) → 28.23 ($N=9$), with IR rising from
0.680 to 1.102. FLUX shows even stronger scaling: HPS increases from 29.34
($N=1$) to 31.23 ($N=3$), then to 32.24 ($N=6$), and ultimately reaches 33.33
at $N=9$. The IR metric follows the same trend, improving from 1.038 to 1.605.
These consistent improvements across all models and metrics indicate that each
additional global starting point introduced by a larger budget $N$ provides new
semantic modes for LoTTS to refine. Since every starting point is further expanded
into multiple locally refined variants, LoTTS is able to generate a rich set of
high-quality candidates around these diverse global modes.

\begin{table}[ht]
\centering
\caption{\textbf{Quantitative results of LoTTS w.r.t. sample count $N$.} Results on Pick-a-Pic and DrawBench for SD2.1, SDXL, and FLUX.}
\label{tab:lotts_scaling}
\resizebox{0.75\textwidth}{!}{%
\begin{tabular}{c c cccc cccc}
\toprule
\multirow{2}{*}{Model} & \multirow{2}{*}{$N$} 
& \multicolumn{4}{c}{Pick-a-Pic} & \multicolumn{4}{c}{DrawBench} \\
\cmidrule(lr){3-6} \cmidrule(lr){7-10}
& & HPS$\uparrow$ & AES$\uparrow$ & Pick$\uparrow$ & IR$\uparrow$
  & HPS$\uparrow$ & AES$\uparrow$ & Pick$\uparrow$ & IR$\uparrow$ \\
\midrule
\multirow{4}{*}{SD2.1}   
& 1 & 20.44 & 5.377 & 20.32 & 0.236 & 21.34 & 5.456 & 20.23 & 0.244 \\
& 3 & 21.12 & 5.532 & 20.56 & 0.413 & 21.88 & 5.623 & 20.54 & 0.398 \\
& 6 & 22.29 & 5.712 & 20.99 & 0.531 & 21.45 & 5.807 & 21.00 & 0.591 \\
& 9 & \textbf{24.52} & \textbf{5.805} & \textbf{21.32} & \textbf{0.680} & \textbf{23.29} & \textbf{5.911} & \textbf{21.47} & \textbf{0.698} \\
\midrule
\multirow{4}{*}{SDXL}   
& 1 & 23.44 & 6.011 & 21.18 & 0.680 & 23.84 & 6.034 & 21.09 & 0.657 \\
      & 3 & 25.23 & 6.124 & 21.55 & 0.712 & 25.11 & 6.123 & 21.55 & 0.701 \\
      & 6 & 26.21 & 6.153 & 21.89 & 0.813 & 26.23 & 6.241 & 21.98 & 0.812 \\
      & 9 & \textbf{28.23} & \textbf{6.304} & \textbf{22.30} & \textbf{1.102} & \textbf{28.90} & \textbf{6.321} & \textbf{22.38} & \textbf{1.111} \\
\midrule
\multirow{4}{*}{FLUX}   
& 1 & 29.34 & 6.298 & 22.07 & 1.038 & 29.28 & 6.223 & 22.05 & 1.100 \\
      & 3 & 31.23 & 6.370 & 22.53 & 1.203 & 30.45 & 6.350 & 22.47 & 1.321 \\
      & 6 & 32.24 & 6.434 & 22.88 & 1.523 & 32.23 & 6.591 & 23.01 & 1.521 \\
      & 9 & \textbf{33.33} & \textbf{6.501} & \textbf{23.04} & \textbf{1.605} & \textbf{33.90} & \textbf{6.890} & \textbf{23.21} & \textbf{1.623} \\
\bottomrule
\end{tabular}}
\end{table}

\myparagraph{Scaling behavior of Best-of-$N$.}
Tab.~\ref{tab:bestofn_scaling} shows that the Best-of-$N$ baseline consistently improves
as the number of globally sampled images increases. This trend is expected:
drawing more samples enlarges the global search space, increasing the likelihood
of encountering rare but high-quality generations. As $N$ grows, the tails of the
model's output distribution are explored more thoroughly, naturally raising the
probability of selecting an image with stronger aesthetics or semantic alignment.
The numerical results reflect this behavior clearly. For \textbf{SD2.1}, HPS improves
from 20.44 at $N{=}1$ to 21.56 at $N{=}9$, then to 22.21 at $N{=}17$, ultimately reaching
24.23 at $N{=}25$. \textbf{SDXL} follows a similar progression, increasing from 23.44 to
24.54, then to 26.02, and finally to 28.21 when $N{=}30$. \textbf{FLUX} exhibits the same
scaling pattern, with HPS rising from 29.34 at $N{=}1$ to 30.23 at $N{=}9$, then to 31.14
at $N{=}14$, and ultimately reaching 33.32 at $N{=}36$.
The other metrics, AES, Pick, and IR, show comparable monotonic gains, though
with diminishing increments as $N$ becomes large. Overall, these results confirm
that Best-of-$N$ behaves like a global search procedure: as more samples are drawn,
the chance of finding a superior candidate increases, leading to smooth and
predictable scaling trends across models and evaluation metrics.

\begin{table}[ht]
\centering
\caption{\textbf{Quantitative results of Best-of-$N$ baseline w.r.t. sample count $N$.} Results on Pick-a-Pic and DrawBench for SD2.1, SDXL, and FLUX.}
\label{tab:bestofn_scaling}
\resizebox{0.75\textwidth}{!}{%
\begin{tabular}{c c cccc cccc}
\toprule
\multirow{2}{*}{Model} & \multirow{2}{*}{$N$} 
& \multicolumn{4}{c}{Pick-a-Pic} & \multicolumn{4}{c}{DrawBench} \\
\cmidrule(lr){3-6} \cmidrule(lr){7-10}
& & HPS$\uparrow$ & AES$\uparrow$ & Pick$\uparrow$ & IR$\uparrow$
  & HPS$\uparrow$ & AES$\uparrow$ & Pick$\uparrow$ & IR$\uparrow$ \\
\midrule
\multirow{4}{*}{SD2.1}   
& 1  & 20.44 & 5.377 & 20.32 & 0.236 & 21.34 & 5.456 & 20.23 & 0.244 \\
& 9  & 21.56 & 5.534 & 21.04 & 0.470 & 22.45 & 5.589 & 20.59 & 0.446 \\
& 17 & 22.21 & 5.755 & 21.11 & 0.524 & 23.17 & 5.817 & 21.10 & 0.602 \\
& 25 & \textbf{24.23} & \textbf{5.825} & \textbf{21.26} & \textbf{0.681} & \textbf{23.33} & \textbf{5.915} & \textbf{21.50} & \textbf{0.701} \\
\midrule
\multirow{4}{*}{SDXL}   
& 1  & 23.44 & 6.011 & 21.18 & 0.680 & 23.84 & 6.034 & 21.09 & 0.657 \\
& 9  & 24.54 & 6.198 & 22.01 & 0.790 & 25.27 & 6.238 & 22.23 & 0.756 \\
& 20 & 26.02 & 6.201 & 22.11 & 0.833 & 26.31 & 6.233 & 22.30 & 0.832 \\
& 30 & \textbf{28.21} & \textbf{6.302} & \textbf{22.21} & \textbf{1.108} & \textbf{28.88} & \textbf{6.330} & \textbf{22.42} & \textbf{1.113} \\
\midrule
\multirow{4}{*}{FLUX}   
& 1  & 29.34 & 6.298 & 22.07 & 1.038 & 29.28 & 6.223 & 22.05 & 1.100 \\
& 9  & 30.23 & 6.299 & 22.89 & 1.235 & 30.46 & 6.290 & 22.33 & 1.221 \\
& 14 & 31.14 & 6.343 & 22.93 & 1.521 & 32.90 & 6.599 & 22.98 & 1.533 \\
& 36 & \textbf{33.32} & \textbf{6.531} & \textbf{23.09} & \textbf{1.615} & \textbf{33.88} & \textbf{6.891} & \textbf{23.22} & \textbf{1.620} \\
\bottomrule
\end{tabular}}
\end{table}

\myparagraph{Scaling comparison.}
As illustrated in Fig.~\ref{fig:scaling}(2) of the main paper, aligning both methods by the same total
sampling budget $N$ reveals a clear and consistent trend: LoTTS delivers
substantially higher-quality results under the same cost and reaches strong
performance far earlier than Best-of-$N$. While Best-of-$N$ relies solely on
global exploration, LoTTS allocates part of its budget to targeted local
refinement guided by attention-derived masks, generating multiple high-quality
variants around semantically meaningful regions. These local refinements allow
LoTTS to recover strong candidates that Best-of-$N$ would only encounter after
significantly larger global sampling effort. Across SD2.1, SDXL, and FLUX,
LoTTS with only $N=9$ matches or surpasses Best-of-$N$ results that require
25--36 global samples, corresponding to a 2--4$\times$ improvement in sample
efficiency. This demonstrates that LoTTS extracts far more value from each
generated image than traditional global resampling strategies.

\section{Discussion}
\label{app:fail}

This supplementary discussion expands upon the brief analysis in 
Sec.~\ref{sec:exp} and provides a unified perspective connecting 
empirical failure modes (illustrated in Fig.~\ref{fig:failure_cases}) 
with the theoretical guarantees and boundary conditions formalized in 
Appendix~\ref{app:TheoreticalAnalysis}.  
We organize the discussion into two parts: \emph{Failure Analysis} and 
\emph{Directions for Improvement}.

\begin{figure}[t]
\centering
\includegraphics[width=0.9\linewidth]{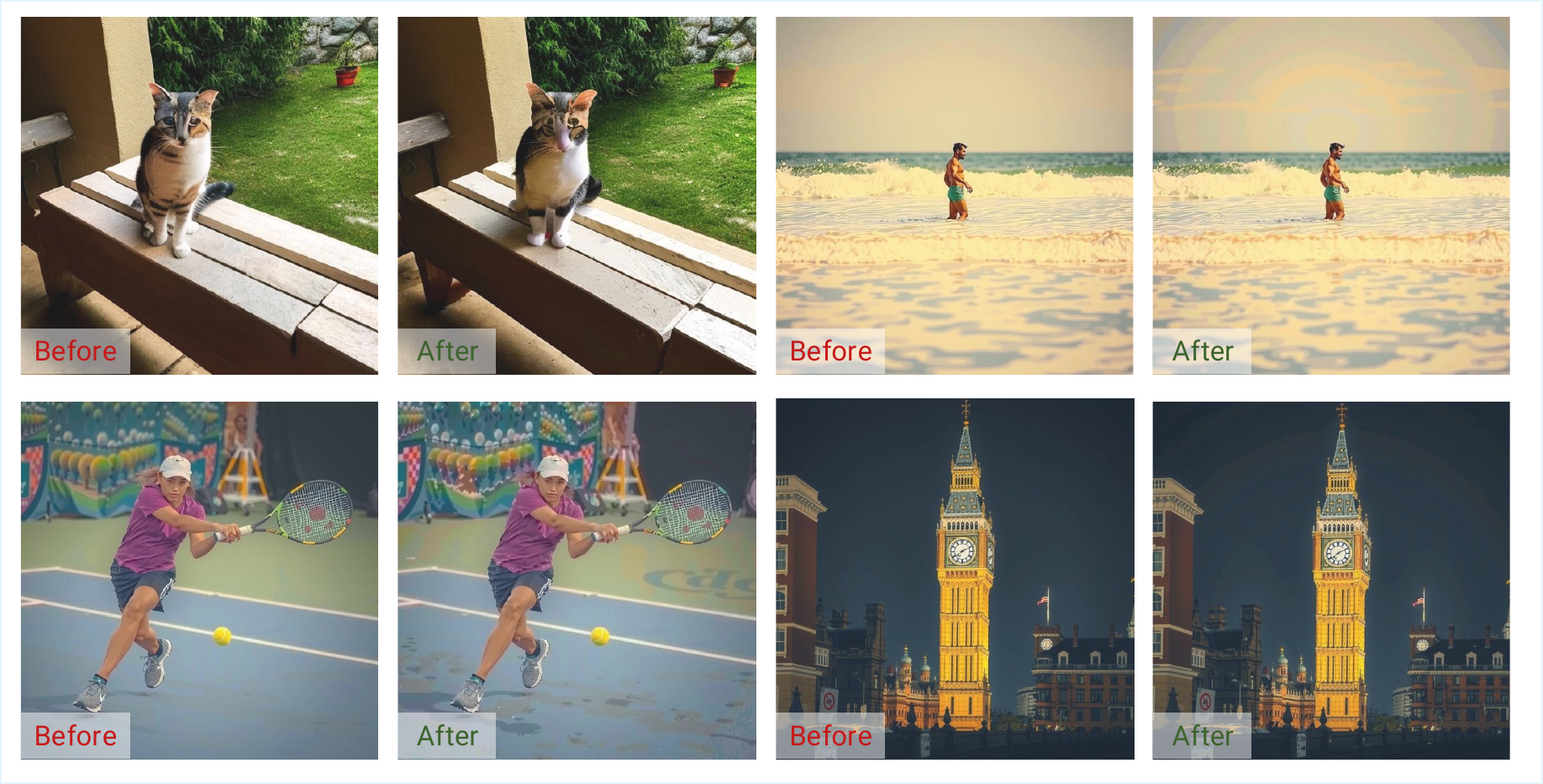}
\vspace{-5pt}
\caption{\textbf{Failure cases of LoTTS.} While LoTTS reduces many
local artifacts, a few failures remain in boundary geometry, in
scenes requiring global coherence or containing low-texture regions (e.g., sky, court floor), and in cases with only subtle defects
where refinements bring limited improvements.}
\label{fig:failure_cases}
\end{figure}

\subsection{Failure Analysis}
\label{app:fail_1}

\myparagraph{Imperfect masks and unstable attention signals.}
A major source of failure arises from attention-based masks missing true
defects or selecting unstable high-frequency regions. Subtle artifacts,
micro-textures, or weak structural inconsistencies may receive insufficient
attention, leading to \textit{low recall}, while noisy activations near
boundaries often generate \textit{false positives}. These behaviors correspond
directly to the theoretical sensitivity to precision and recall discussed in
Appendix~\ref{app:TheoreticalAnalysis}. In particular, insufficient recall
violates the requirement in Eq.~\eqref{eq:condition_rho}, preventing LoTTS from
reaching enough defective patches to yield meaningful improvement. Conversely,
poor precision inflates the false-positive harm term
$\rho(\tfrac{1}{\pi}-1)h_\ell\overline{\gamma}$, which must remain smaller than
$q\overline{\delta}$ for each selected patch to provide positive expected
benefit, as formalized in Eq.~\eqref{eq:pi_threshold}.

\myparagraph{Global defects and violation of sparsity.}
A second category of failures arises from \textit{global semantic or geometric
errors}, such as incorrect perspective, inconsistent lighting, or cross-object
misalignment. Because LoTTS modifies only a limited set of selected patches, it
lacks the ability to enact coordinated, image-wide corrections, and therefore
cannot address defects that require globally coherent changes. Such cases
fundamentally violate the sparse-defect assumption $s \ll M$ that underpins
Theorem~\ref{thm:main} and Corollary~\ref{cor:equal}. When defects behave as if
they are dense or spatially widespread, the global-harm amplification factor
$(\tfrac{M}{s}-1)h_g\overline{\gamma}$ diminishes, thereby eliminating the
structural advantage that makes localized refinement more compute-efficient than
global resampling. This explains why global TTS can occasionally succeed in
scenes with strong global inconsistencies, despite its disadvantages in
sparse-defect settings.

\myparagraph{Inputs at quality extremes.}
LoTTS also encounters difficulties when the input image lies at the extremes of
the quality spectrum. In very low-quality images, the effective set of defects
behaves as though $s \approx M$, which violates the sparsity regime and limits
the potential for safe and effective localized improvement. Conversely, when the
input image is already near-perfect, the expected repair gain
$\overline{\delta}$ becomes extremely small, reducing the net improvement margin
and making the refinement process susceptible to small perturbation artifacts.
Both scenarios follow directly from Theorem~\ref{thm:main}, which predicts that
LoTTS loses its advantage either when the intrinsic repair benefit collapses or
when the sparsity assumption no longer holds.

\myparagraph{Failure summary.}
Across these categories, the observed empirical failures align closely with the
theoretical failure conditions formalized in Corollary~\ref{cor:failure}. When
defects are dense or global, the global-harm asymmetry disappears and LoTTS no
longer enjoys a structural advantage. When mask precision is extremely low,
false-positive edits impose excessive local harm that overwhelms the expected
benefit. And when the local repair mechanism yields only a limited improvement
relative to $\theta_g$, localized refinement lacks sufficient corrective
strength. These patterns demonstrate that LoTTS's limitations arise not from
unpredictable behavior, but from predictable breakdowns of the assumptions
required for theoretical dominance.

\subsection{Future Directions}
\label{app:fail_2}

\myparagraph{Restrict computation strictly within the mask.}
A promising direction is to integrate algorithms whose computation is 
explicitly confined to the mask region, such as mask-restricted resampling.  
By design, these methods allocate compute exclusively to the regions that 
require refinement, effectively realizing an extreme form of cost asymmetry 
($C_\ell \ll C_g$).  
According to Corollary~\ref{cor:lowcost}, this places the method directly in 
a regime where localized refinement is provably more compute-efficient than 
global resampling.  
Such mask-restricted algorithms therefore offer a principled pathway toward 
significantly improving test-time scaling efficiency: the same compute 
budget yields far more effective updates simply because no computation is 
wasted on clean regions.

\myparagraph{Baseline beyond Best-of-$N$.}
Although our current implementation adopts a Best-of-$N$ sampling scheme 
within the masked region, this strategy represents only the simplest form 
of local search. A more powerful alternative is to integrate global search 
algorithms, such as beam search, that branch along the denoising trajectory, 
but do so \emph{only within the masked area}.  
Such mask-restricted branching preserves the locality and compute efficiency 
of LoTTS while enabling a substantially richer exploration of refinement 
hypotheses than Best-of-$N$.  
This opens the door to ``global search applied locally,'' where complex 
trajectory-level decisions are made without ever incurring the cost or harm 
of full-image resampling.

\begin{table}[t]
\centering
\small
\caption{Notation used in the theoretical analysis.}
\label{tab:notation_theory}
\vspace{-2mm}
\begin{tabular*}{0.95\textwidth}{l @{\extracolsep{\fill}} p{0.7\textwidth}}
\toprule
\textbf{Symbol} & \textbf{Description} \\
\midrule
$M$ & total number of non-overlapping image patches \\
$s$ & number of defective patches ($s \ll M$) \\
$\mathcal{D}$ & set of defective patches, with $|\mathcal{D}|=s$ \\
$\widehat{\mathcal{D}}$ & set of patches selected by LoTTS for resampling \\
$\pi$ & precision: fraction of selected patches that are truly defective \\
$\rho$ & recall: fraction of defective patches successfully selected \\
$x$ & generated image; $x_j$ denotes patch $j$ \\
$r(x)$ & perceptual quality functional of image $x$ \\
$r_j(x_j)$ & patch-level quality score \\
$w_j$ & importance weight of patch $j$ in quality decomposition \\
$\delta_j$ & expected gain if defective patch $j$ is repaired \\
$\gamma_j$ & expected loss if clean patch $j$ is harmed \\
$\overline{\delta}$ & average weighted repair gain over defective patches \\
$\overline{\gamma}$ & average weighted harm over clean patches \\
$\theta_g$ & probability of repairing a defective patch under global resampling \\
$q$ & probability of repairing a defective patch under localized resampling \\
$h_g$ & probability of harming a clean patch under global resampling \\
$h_\ell$ & probability of harming a clean patch under localized resampling \\
$C_g$ & compute cost of one global resampling trial \\
$C_\ell$ & compute cost of one localized resampling trial \\
$B$ & total compute budget (e.g., measured in NFEs) \\
\bottomrule
\end{tabular*}
\end{table}

\myparagraph{Early-stop mechanisms for low-potential cases.}
Another practical extension is to incorporate an early-stop criterion that 
detects when an image has little potential for localized improvement.  
In many cases, the mask may identify only marginal or low-confidence regions, 
or the estimated patch-level quality signals may indicate that no meaningful 
refinement can be achieved within the masked area.  
Continuing localized resampling in such situations not only yields negligible 
benefit but also risks introducing unnecessary perturbations.  
An early-stop mechanism, triggered by low mask coverage, low defect confidence, 
or minimal improvement predicted from intermediate diffusion states, would 
allow the system to gracefully abstain from refinement when appropriate.  
This ensures that compute is spent only when LoTTS is theoretically capable 
of providing positive expected gain, further improving test-time efficiency 
and robustness.

\myparagraph{Improving mask-generation strategies.}
The current mask is constructed from relative internal inconsistencies, which
can miss globally subtle or low-contrast defects. A more principled alternative
is to incorporate absolute quality cues, such as no-reference IQA metrics, to
better identify patches with genuine improvement potential. In addition, richer
mask-generation strategies based on multi-signal fusion (e.g., trajectory
disagreement, latent-gradient saliency) may significantly improve both precision
and recall. By stabilizing the mask and grounding it in multiple complementary
signals, these approaches can move the mask into the regime where LoTTS is
theoretically guaranteed to yield positive expected
gain, as shown in Eq.~\eqref{eq:pi_threshold}, thereby enhancing robustness and overall
refinement effectiveness.

\section{Theoretical Analysis}
\label{app:TheoreticalAnalysis}

\subsection{Problem Setup and Preliminaries}
\label{app:TheoreticalAnalysis_1}

\myparagraph{Overview.}
Our theoretical analysis proceeds in three steps. 
(1) We first decompose global image quality into patch-level contributions
via an additive decomposition, which allows local refinements to be analyzed independently. 
(2) We then characterize the stochastic behavior of the mask via precision and recall, 
which determines how often LoTTS touches defective or clean patches. 
(3) Combining these two components, we compute the expected gain of global and local TTS 
and derive the condition under which localized refinement is more compute-efficient.

\myparagraph{Notation.}
We analyze the conditions under which localized TTS achieves higher quality or efficiency than global TTS. For clarity, Tab.~\ref{tab:notation_theory} summarizes the variables used throughout this section.  
This ensures every symbol in the derivation has an explicit meaning.

\subsubsection{Image Quality Decomposition}

Before comparing global and local strategies, we first relate global quality $r(x)$ to patch-level scores.  
We adopt a standard \emph{additive decomposition} assumption:
\begin{equation}
\label{eq:additive}
r(x) \ =\ \sum_{j=1}^M w_j\, r_j(x_j),
\end{equation}
where $w_j \ge 0$ are weights reflecting the importance of each patch.  
This decomposition holds exactly when the quality metric is patch-separable, as is the case for PSNR, patch-level LPIPS, and regional FID computed over local crops.
It allows us to express global gain as a sum of weighted patch-level gains.

\subsubsection{Mask Selection Statistics}

We next connect the mask’s precision/recall to expected true/false selections.  
This step is necessary to compute how many patches are truly improved versus unnecessarily modified.
\begin{tcolorbox}[colback=gray!10, colframe=gray!30, boxrule=0.3pt]
\begin{lemma}[Expected TP/FP Statistics]\label{lem:pr}
Let $\mathcal{D}$ be the set of defective patches with $|\mathcal{D}| = s$, and let $\widehat{\mathcal{D}}$ be the mask-selected set.  
Define $\mathrm{TP}=|\mathcal{D}\cap\widehat{\mathcal{D}}|$ and $\mathrm{FP}=|\widehat{\mathcal{D}}\setminus \mathcal{D}|$.
If the mask has recall $\rho$ and precision $\pi$, then
\[
\begin{aligned}
\mathbb{E}[\mathrm{TP}] &= \rho s,\\
\mathbb{E}[|\widehat{\mathcal{D}}|] &= \frac{\rho s}{\pi},\\
\mathbb{E}[\mathrm{FP}] &= \rho s\!\left(\frac{1}{\pi}-1\right).
\end{aligned}
\]
\end{lemma}
\end{tcolorbox}

\begin{proof}
By the definition of recall $\rho=\frac{\mathbb{E}[\mathrm{TP}]}{s}$, we immediately have:
\begin{equation}
\mathbb{E}[\mathrm{TP}]=\rho\, s.
\end{equation}
By the (operational) definition of precision $\pi=\frac{\mathbb{E}[\mathrm{TP}]}{\mathbb{E}[|\widehat{\mathcal{D}}|]}$, we get:
\begin{equation}
\mathbb{E}[|\widehat{\mathcal{D}}|]=\frac{\mathbb{E}[\mathrm{TP}]}{\pi}
=\frac{\rho\, s}{\pi}.
\end{equation}
Finally, using linearity of expectation and $\mathrm{FP}=|\widehat{\mathcal{D}}|-\mathrm{TP}$,
\begin{equation}
\mathbb{E}[\mathrm{FP}]
=\mathbb{E}[|\widehat{\mathcal{D}}|]-\mathbb{E}[\mathrm{TP}]
=\frac{\rho\, s}{\pi}-\rho\, s
=\rho\, s\bigl(\tfrac{1}{\pi}-1\bigr).
\end{equation}
\end{proof}

\myparagraph{Discussion.}
\begin{itemize}[leftmargin=*]
\item \textbf{Why use }\(\pi=\frac{\mathbb{E}[\mathrm{TP}]}{\mathbb{E}[|\widehat{\mathcal{D}}|]}\)\textbf{?} \\ 
Precision is often defined as \(\mathbb{E}\!\left[\frac{\mathrm{TP}}{|\widehat{\mathcal{D}}|}\mathbf{1}\{|\widehat{\mathcal{D}}|>0\}\right]\).  
To avoid division-by-zero and keep algebra tractable, we adopt the \emph{ratio-of-expectations} form, which is standard in compute-budget analyses.  
If one insists on the expectation-of-ratio definition, Jensen-type arguments yield the bound
\(\mathbb{E}[|\widehat{\mathcal{D}}|]\ge \frac{\mathbb{E}[\mathrm{TP}]}{\pi}\), so our equalities become tight upper/lower bounds; the conclusions below only change by harmless inequalities.

\item \textbf{Edge cases.}  \\
\(\pi=1\) implies no false positives, hence \(\mathbb{E}[\mathrm{FP}]=0\).  
\(\rho=1\) means all \(s\) defects are captured in expectation, i.e., \(\mathbb{E}[\mathrm{TP}]=s\).  
We exclude the degenerate \(\pi=0\) case since precision \(0\) implies \(\mathbb{E}[\mathrm{TP}]=0\) or \(\mathbb{E}[|\widehat{\mathcal{D}}|]=\infty\).

\item \textbf{What randomness is averaged over?}  \\
Expectations are taken over the randomness of mask construction (and, if applicable, sampling noise).  
In practice, the mask may be stochastic due to thresholding on noisy attention maps,
variation across diffusion trajectories, or random perturbations used during test-time evaluation.
No independence assumptions are required; linearity of expectation suffices.

\end{itemize}

\subsection{Main Theorem}
\label{app:TheoreticalAnalysis_2}

This section derives the expected improvement of a single global or localized resampling step, and then extends it to a fixed compute budget. These expressions form the basis of the dominance condition analyzed in the next subsection.

\myparagraph{Expected Gain per Resampling Step.}
We begin by analyzing the expected gain of a \emph{single} resampling trial.  
Let $x$ denote the image before the update and $x^{+}$ the image after the update, and define:
\begin{equation}
\Delta r := r(x^{+}) - r(x).
\end{equation}
To relate global quality to patch-level behavior, we use the additive decomposition Eq.~\eqref{eq:additive}:
\begin{equation}
r(x) \ =\ \sum_{j=1}^M w_j\, r_j(x_j),
\end{equation}
where $w_j\ge 0$ reflects the importance of patch $j$.  
Substituting the decomposition for both $x$ and $x^{+}$ yields:
\begin{equation}
\Delta r\ =\ \sum_{j=1}^{M} w_j \big(r_j(x_j^{+}) - r_j(x_j)\big).
\end{equation}
For each defective patch $j\in \mathcal{D}$, let $\delta_j\ge 0$ denote the expected improvement when the patch is successfully repaired;  
for each clean patch $j\notin \mathcal{D}$, let $\gamma_j\ge 0$ denote the expected loss when the patch is harmed.  
Define the weighted averages:
\begin{equation}
\overline{\delta} := \frac{1}{s}\sum_{j\in \mathcal{D}} w_j \delta_j,
\qquad
\overline{\gamma} := \frac{1}{M-s}\sum_{j\notin \mathcal{D}} w_j \gamma_j,
\end{equation}
where $s=|\mathcal{D}|\ll M$ is the number of defective patches.  
These quantities do not require independence across patches; they capture only the expected per-patch effect of one resampling step.
For the local harm term derived below, we assume that false-positive patches
are representative of clean patches in terms of $\gamma_j$; if
mask-selected clean patches have systematically different harm,
$\overline{\gamma}$ should be replaced by the corresponding conditional
average.

\myparagraph{Global resampling.}
In global test-time scaling, \emph{all} patches are resampled. 
Each defective patch $j\in \mathcal{D}$ is repaired with probability $\theta_g$, and each clean patch $j\notin \mathcal{D}$ is harmed with probability $h_g$.  
Aggregating over all patches gives the per-trial expected improvement:
\begin{equation}
\label{eq:gain_global}
\mathbb{E}[\Delta r]_{\mathrm{global}}
\ =\ s\,\theta_g\,\overline{\delta}
\;-\; (M-s)\,h_g\,\overline{\gamma}.
\end{equation}
The benefit scales with the $s$ defective patches, while the harm scales with the much larger number $(M-s)$ of clean patches.

\myparagraph{Localized resampling (LoTTS).}
In localized test-time scaling, only patches selected by the mask $\widehat{\mathcal{D}}$ are resampled: potentially repairing true positives $\mathcal{D}\cap\widehat{\mathcal{D}}$ and harming false positives $\widehat{\mathcal{D}}\setminus \mathcal{D}$.
Let recall $\rho$ and precision $\pi$ be defined as in Lemma~\ref{lem:pr}.  
Let $q$ be the probability of repairing a defective selected patch, and $h_\ell$ the probability of harming a selected clean patch.  
Using Lemma~\ref{lem:pr}, the expected numbers of selected defective and clean patches are $\rho s$ and $\rho s(\tfrac{1}{\pi}-1)$, respectively, leading to:
\begin{equation}
\label{eq:gain_local}
\mathbb{E}[\Delta r]_{\mathrm{local}}
\ =\
\rho\,s\,q\,\overline{\delta}
\;-\;
\rho\,s\bigl(\tfrac{1}{\pi}-1\bigr)\,h_\ell\,\overline{\gamma}.
\end{equation}

\myparagraph{Interpretation.}
Global resampling can repair all $s$ defective patches, but pays a harm penalty scaling with the large number $(M-s)$ of clean patches.  
LoTTS repairs only $\rho s$ defective patches in expectation, but its harm penalty grows only with the false positives, of order $\rho s(\tfrac{1}{\pi}-1)$ rather than $M$.  
This asymmetry is the essence of the \emph{sparse-defect advantage}: when $s\ll M$, localized refinement focuses compute on problematic regions while avoiding widespread degradation.

\myparagraph{Expected Gain per Compute Unit.}
To compare the two strategies under a fixed compute budget $B$, we normalize their per-trial gains by their respective costs $C_g$ and $C_\ell$.  
Since a total of $B/C_g$ global trials or $B/C_\ell$ local trials can be performed, we obtain:
\begin{tcolorbox}[colback=gray!10, colframe=gray!30, boxrule=0.3pt]
\begin{theorem}[Expected Quality Gains]\label{thm:main}
For a total compute budget $B$, let $C_g$ and $C_\ell$ denote the cost of a single global and localized trial, and let $\overline{\delta}$ and $\overline{\gamma}$
denote the average per–patch repair gain and harm penalty.  
Under the additive quality decomposition Eq.~\eqref{eq:additive},
with each trial independently resampling from the \emph{same} reference
image\footnote{In practice, all candidates are generated in parallel from
the same base image and scored by a verifier that selects the best one
(Best-of-$N$).  The linear scaling $\tfrac{B}{C}\Delta^{(1)}$ therefore
serves as a \emph{compute-efficiency proxy}: the strategy with higher
$\Delta^{(1)}/C$ produces higher-quality candidates per unit of compute,
and thus also yields a better Best-of-$N$ outcome under any fixed
budget.}, the expected quality
improvements of Global TTS and LoTTS are:
\[
\begin{aligned}
\mathbb{E}[\Delta r]_{\mathrm{global},B}
&\;=\;
\frac{B}{C_g}
\bigl[\,s\,\theta_g\,\overline{\delta}
\;-\;(M-s)\,h_g\,\overline{\gamma}\,\bigr], \\[3pt]
\mathbb{E}[\Delta r]_{\mathrm{local},B}
&\;=\;
\frac{B}{C_\ell}
\bigl[\,\rho s\,q\,\overline{\delta}
\;-\;\rho s\!\big(\tfrac{1}{\pi}-1\big) h_\ell \overline{\gamma}\,\bigr].
\end{aligned}
\]
\end{theorem}
\end{tcolorbox}

\myparagraph{Interpretation.}
Both methods yield an expected quality change equal to  
(average repair gain) minus (average harm penalty),  
scaled by the number of trials affordable under budget $B$.
These compute-normalized expressions provide the basis for the dominance condition analyzed in the next subsection.

\subsection{Derived Corollaries}
\label{app:TheoreticalAnalysis_3}
Having established the expected-gain expressions, we now determine when localized resampling (LoTTS) achieves a higher compute-normalized
expected improvement than global test-time scaling.

\myparagraph{General Corollaries.}
The following corollary gives the general condition under which LoTTS achieves higher compute-normalized expected improvement than global TTS.

\begin{tcolorbox}[colback=gray!10, colframe=gray!30, boxrule=0.3pt]
  \begin{corollary}[General Case]\label{cor:dominance}
  LoTTS achieves higher compute-normalized expected improvement than Global TTS if and only if
  \[
  \begin{gathered}
  \frac{\rho}{C_\ell}\!\left(q\,\overline{\delta}
    -(\tfrac{1}{\pi}-1)\,h_\ell\,\overline{\gamma}\right)
  \\[4pt]
  >\;\;
  \frac{1}{C_g}\!\left(\theta_g\,\overline{\delta}
    -(\tfrac{M}{s}-1)\,h_g\,\overline{\gamma}\right)
  \end{gathered}
  \]
  \end{corollary}
\end{tcolorbox}

\myparagraph{Interpretation.}
The left-hand side represents the compute-normalized expected improvement contributed by the patches selected by LoTTS: recall $\rho$ determines how many defective patches are reached, $q\overline{\delta}$ is the average benefit of repairing a defect, and $(\tfrac{1}{\pi}-1)h_\ell\overline{\gamma}$ accounts for harm from false positives. The right-hand side describes the corresponding global gain, where the potential harm is amplified by $(M-s)$ clean patches rather than by the (much smaller) number of false positives.

The dominance inequality tells us when LoTTS is better, but it does not yet answer the key practical question: 
\emph{how accurate must the mask be for LoTTS to actually win?}
This subsection makes this explicit by deriving the required 
\emph{recall} $\rho$ and \emph{precision} $\pi$ thresholds.

\begin{tcolorbox}[colback=gray!10, colframe=gray!30, boxrule=0.3pt]
\begin{corollary}[Dominance Under Cost Asymmetry]\label{cor:lowcost}
Let $\Delta_\ell^{(1)}$ and $\Delta_g^{(1)}$ denote the per-trial expected gains of localized and global resampling, respectively (i.e., the bracketed expressions in Theorem~\ref{thm:main}).
From Theorem~\ref{thm:main}, LoTTS dominates global TTS iff:
\begin{equation}\label{eq:lowcost_exact}
\frac{\Delta_\ell^{\left(1\right)}}{C_\ell}
\;>\;
\frac{\Delta_g^{\left(1\right)}}{C_g}.
\end{equation}
In the regime where
\[
C_\ell \ll C_g,
\qquad
\Delta_\ell^{\left(1\right)} = \Theta(\Delta_g^{\left(1\right)})>0,
\]
condition \eqref{eq:lowcost_exact} simplifies to:
\[
\frac{\Delta_\ell^{\left(1\right)}}{C_\ell}
\;\gg\;
\frac{\Delta_g^{\left(1\right)}}{C_g}.
\]
Thus LoTTS achieves strictly larger compute-normalized gain under any fixed
compute budget~$B$.
\end{corollary}
\end{tcolorbox}

\myparagraph{Required recall.}
Solving the inequality in Corollary~\ref{cor:dominance} for $\rho$ yields:
\begin{align}
\label{eq:condition_rho}
& \rho\ >\ \frac{C_\ell}{C_g}\,
\frac{\theta_g\overline{\delta}-\big(\tfrac{M}{s}-1\big)h_g\overline{\gamma}}
     {q\overline{\delta}-(\tfrac{1}{\pi}-1)h_\ell\overline{\gamma}} \nonumber\\
&\text{provided}\quad
q\overline{\delta}-(\tfrac{1}{\pi}-1)h_\ell\overline{\gamma} > 0.
\end{align}
The denominator is the expected net benefit of selecting a single patch; it must
be positive for localized refinement to be useful.

\myparagraph{Required precision.}
A convenient sufficient condition for the denominator of Eq.~\eqref{eq:condition_rho} to be positive is:
\begin{equation}
\label{eq:pi_threshold}
q\overline{\delta}-(\tfrac{1}{\pi}-1)h_\ell\overline{\gamma} \;>\; 0
\;\;\Longleftrightarrow\;\;
\pi \;>\; \frac{1}{1+ \tfrac{q\overline{\delta}}{h_\ell\overline{\gamma}}}.
\end{equation}
This threshold is easier to satisfy when true repairs provide large gains
($q\overline{\delta}$ high) or when false-positive edits cause limited harm
($h_\ell\overline{\gamma}$ low).  In either case, even imperfect masks yield
positive expected gain per selected patch.

\myparagraph{Specialized Corollaries.}
We now present simplified dominance conditions obtained from 
Theorem~\ref{thm:main} under practically relevant regimes.
Each corollary corresponds directly to one of the scenarios in the Q\&A 
of Section~\ref{app:TheoreticalAnalysis_4}.

\begin{proof}

\myparagraph{Step 1: Substitute the expressions from Theorem~\ref{thm:main}.}
The expected improvements under total budget $B$ are
\[
\mathbb{E}[\Delta r]_{\mathrm{local},B}
=\frac{B}{C_\ell}\,\Delta_\ell^{\left(1\right)},
\qquad
\mathbb{E}[\Delta r]_{\mathrm{global},B}
=\frac{B}{C_g}\,\Delta_g^{\left(1\right)}.
\]

\myparagraph{Step 2: Translate dominance into an inequality.}
LoTTS dominates global TTS exactly when
\[
\mathbb{E}[\Delta r]_{\mathrm{local},B}
\;>\;
\mathbb{E}[\Delta r]_{\mathrm{global},B},
\]
which becomes:
\[
\frac{B}{C_\ell}\,\Delta_\ell^{\left(1\right)}
\;>\;
\frac{B}{C_g}\,\Delta_g^{\left(1\right)}.
\tag{$\star$}\label{eq:lowcost_raw}
\]

\myparagraph{Step 3: Cancel the common positive factor $B$.}
Since $B>0$, inequality \eqref{eq:lowcost_raw} is equivalent to:
\[
\frac{\Delta_\ell^{\left(1\right)}}{C_\ell}
\;>\;
\frac{\Delta_g^{\left(1\right)}}{C_g},
\]
which is the exact dominance condition in the corollary.

\myparagraph{Step 4: Apply the cost-asymmetry assumptions.}
Assume
\[
C_\ell \ll C_g,
\qquad
\Delta_\ell^{\left(1\right)} \approx \Delta_g^{\left(1\right)} > 0.
\]
The first assumption implies:
\[
\frac{1}{C_\ell} \gg \frac{1}{C_g},
\]
and the second implies that $\Delta_\ell^{\left(1\right)}$ and $\Delta_g^{\left(1\right)}$
are of comparable magnitude.

\myparagraph{Step 5: Compare the compute-normalized benefits.}
Using these relations,
\[
\frac{\Delta_\ell^{(1)}}{C_\ell}
=
\Delta_\ell^{(1)}\!\cdot\!\frac{1}{C_\ell}
\approx
\Delta_g^{(1)}\!\cdot\!\frac{1}{C_\ell}
\;\gg\;
\Delta_g^{(1)}\!\cdot\!\frac{1}{C_g}
=
\frac{\Delta_g^{(1)}}{C_g}.
\]
Thus:
\[
\frac{\Delta_\ell^{\left(1\right)}}{C_\ell}
\;\gg\;
\frac{\Delta_g^{\left(1\right)}}{C_g}.
\]

\myparagraph{Step 6: Conclude dominance for any fixed budget $B$.}
Substituting this back into~\eqref{eq:lowcost_raw} shows that:
\[
\mathbb{E}[\Delta r]_{\mathrm{local},B}
>
\mathbb{E}[\Delta r]_{\mathrm{global},B},
\]
for any fixed $B$, establishing the corollary.
\end{proof}

\myparagraph{Interpretation.}
When local sampling is substantially cheaper, compute becomes the decisive 
advantage: LoTTS performs many more trials within the same budget, so even 
moderate $(\rho,\pi,q)$ are sufficient for dominance.

\begin{tcolorbox}[colback=gray!10, colframe=gray!30, boxrule=0.3pt]
\begin{corollary}[Dominance Under Equal-Cost Sparse Regime]\label{cor:equal}
Assume $C_\ell = C_g$.  
Then Theorem~\ref{thm:main} gives the exact condition
\begin{gather}\label{eq:equal_exact}
(\rho q - \theta_g)\overline{\delta}
\;>\;
\rho\big(\tfrac{1}{\pi}-1\big)h_\ell\overline{\gamma}
-
\big(\tfrac{M}{s}-1\big)h_g\overline{\gamma}.
\end{gather}
In the sparse-defect regime $s\ll M$ and with benign local edits $h_\ell\ll h_g$,
the dominant terms yield the approximation
\[
\rho q
\;\gtrsim\;
\theta_g
-
\frac{M}{s}\,\frac{h_g\overline{\gamma}}{\overline{\delta}}.
\]
Thus, under equal compute, dominance is governed by the relative scaling of the
harm terms.
\end{corollary}
\end{tcolorbox}

\begin{proof}
We begin from the exact dominance inequality in Theorem~\ref{thm:main}:
\begin{align}
\label{eq:thm_raw}
&\frac{B}{C_\ell}\bigl(
\rho q \overline{\delta}
-
\rho(\tfrac{1}{\pi}-1) h_\ell \overline{\gamma}
\bigr) \notag\\
&\quad>\;
\frac{B}{C_g}\bigl(
\theta_g \overline{\delta}
-
(\tfrac{M}{s}-1) h_g \overline{\gamma}
\bigr).
\end{align}

\myparagraph{Step 1: Specialize to $C_\ell = C_g$.}
Under the equal-cost assumption $C_\ell=C_g$, the prefactor $B/C_\ell$ is
strictly positive and common to both sides.  
Therefore we may cancel it from \eqref{eq:thm_raw}, yielding the exact dominance condition:
\begin{align}
\label{eq:equal_raw_long}
&\rho q \overline{\delta}
-
\rho(\tfrac{1}{\pi}\!-\!1)\, h_\ell \overline{\gamma} \notag\\
&\quad>\;
\theta_g \overline{\delta}
-
(\tfrac{M}{s}\!-\!1)\, h_g \overline{\gamma}.
\end{align}

\myparagraph{Step 2: Collect all repair terms on the left and all harm terms on the right.}
Rearranging Eq.~\eqref{eq:equal_raw_long} gives:
\begin{align}
(\rho q - \theta_g)\overline{\delta}
&>\;
\rho(\tfrac{1}{\pi}-1) h_\ell \overline{\gamma}
\;-\;
(\tfrac{M}{s}-1) h_g \overline{\gamma}.
\label{eq:equal_rearranged}
\end{align}
Eq.~\eqref{eq:equal_rearranged} is algebraically equivalent to the original
dominance condition and valid without approximation.

\myparagraph{Step 3: Apply sparse-defect assumption.}
When $s \ll M$, we have:
\[
\tfrac{M}{s}-1 = \Theta(\tfrac{M}{s}),
\]
so the global-harm reduction on the right-hand side becomes:
\[
-(\tfrac{M}{s}-1) h_g \overline{\gamma}
\approx
-\frac{M}{s} h_g \overline{\gamma},
\]
which is a large negative term that dominates the right-hand side because it scales with the number of
clean patches.

\myparagraph{Step 4: Apply benign-local-edit assumption.}
If $h_\ell \ll h_g$, then the local-harm term:
$\rho(\tfrac{1}{\pi}-1) h_\ell \overline{\gamma}$
is negligible compared to $\frac{M}{s} h_g \overline{\gamma}$, since
(i) it does not scale with $M$, and  
(ii) $h_\ell \ll h_g$ suppresses its magnitude.
Thus, in the joint regime $(s\ll M)$ and $(h_\ell \ll h_g)$, the right-hand side
of Eq.~\eqref{eq:equal_rearranged} satisfies the asymptotic approximation:
\[
\rho(\tfrac{1}{\pi}-1) h_\ell \overline{\gamma}
-
(\tfrac{M}{s}-1) h_g \overline{\gamma}
\;\approx\;
-\frac{M}{s} h_g \overline{\gamma}.
\]

\myparagraph{Step 5: Divide both sides by $\overline{\delta}$.}
Since $\overline{\delta}>0$, we divide Eq.~\eqref{eq:equal_rearranged} to obtain:
\[
\rho q - \theta_g
\;\gtrsim\;
-\frac{M}{s}\frac{h_g\overline{\gamma}}{\overline{\delta}},
\]
or equivalently:
\[
\rho q 
\;\gtrsim\;
\theta_g 
-
\frac{M}{s}\frac{h_g\overline{\gamma}}{\overline{\delta}}.
\]
This is exactly the simplified condition stated in the corollary.
\end{proof}

\myparagraph{Interpretation.}
When compute cost is equal, the repair gains of the two methods are comparable; 
the decisive asymmetry lies in the harm terms.  
Global harm scales with all $(M-s)$ clean patches, whereas local harm scales 
only with false positives.  
In sparse regimes this makes the global harm term much larger, allowing LoTTS 
to dominate even with moderate $(\rho,\pi,q)$.

\begin{tcolorbox}[colback=gray!10, colframe=gray!30, boxrule=0.3pt]
\begin{corollary}[Saturation of Best-of-$N$ Global Resampling]\label{cor:bon}
We analyze Best-of-$N$ under an optimistic \emph{per-patch oracle} model
that independently selects the best outcome for each patch across $N$ draws.
This upper-bounds the true Best-of-$N$ repair rate (which selects a single
image), making the saturation conclusion conservative.
For global Best-of-$N$,
\[
C_g(N)=N\,C_g(1),\qquad
\theta_g(N)=1-\bigl(1-\theta_g(1)\bigr)^{\!N}.
\]
The compute-normalized improvement becomes
\[
\frac{s}{NC_g\!\left(1\right)}
\bigl(
\theta_g(N)\overline{\delta}
-
(\tfrac{M}{s}-1)h_g\overline{\gamma}
\bigr).
\]
Since $\theta_g(N)$ increases sublinearly in $N$ while the harm term remains
unchanged, global improvement saturates and cannot overcome the harm term by
increasing $N$.
\end{corollary}
\end{tcolorbox}

\begin{proof}

\myparagraph{Step 1: Define the Best-of-$N$ procedure and its cost.}
A Best-of-$N$ trial draws $N$ independent global resampling samples and
selects the best one.  
Since one global draw costs $C_g\left(1\right)$, performing $N$ draws costs:
\[
C_g(N)=N\,C_g\left(1\right).
\]

\myparagraph{Step 2: Compute the defect repair probability.}
Let $\theta_g\left(1\right)$ be the probability that a \emph{single} global draw
repairs a defect.  
Under independence across draws, the probability that none of the $N$
draws repairs the defect is $(1-\theta_g\left(1\right))^N$.  
Thus the probability that at least one draw repairs it is:
\[
\theta_g(N)
=
1-\big(1-\theta_g\left(1\right)\big)^N.
\]

\myparagraph{Step 3: Substitute into the global gain expression.}
Plugging $C_g(N)$ and $\theta_g(N)$ into the global expression in
Theorem~\ref{thm:main}, the expected improvement under budget $B$ becomes:
\[
\mathbb{E}[\Delta r]_{\mathrm{global},B}
=
\frac{Bs}{C_g(N)}
\bigl(
\theta_g(N)\overline{\delta}
-
(\tfrac{M}{s}-1)h_g\overline{\gamma}
\bigr).
\]

\myparagraph{Step 4: Normalize by compute.}
Dividing both sides by $B$ yields:
\begin{align*}
&\frac{s}{C_g(N)}
\bigl(
\theta_g(N)\overline{\delta}
-
(\tfrac{M}{s}-1)h_g\overline{\gamma}
\bigr) \\
={}\;
&\frac{s}{N C_g(1)}
\bigl(
\theta_g(N)\overline{\delta}
-
(\tfrac{M}{s}-1)h_g\overline{\gamma}
\bigr),
\end{align*}
which is exactly the boxed expression in the corollary.

\myparagraph{Step 5: Show saturation as $N$ increases.}
The function
$\theta_g\left(N\right)=1-\left(1-\theta_g\left(1\right)\right)^N$
is increasing and concave in $N$.  
Indeed, its incremental increase satisfies:
\[
\theta_g(N+1)-\theta_g(N)
=
(1-\theta_g\left(1\right))^N\,\theta_g\left(1\right),
\]
which is positive but strictly decreasing in $N$.  
Thus $\theta_g(N)$ approaches $1$ with diminishing returns.

\myparagraph{Step 6: Conclude saturation of compute-normalized global gain.}
The compute-normalized gain contains the prefactor $s/(N C_g\left(1\right))$, which
decreases linearly in $N$, while the harm term
\[
(\tfrac{M}{s}-1)h_g\overline{\gamma}
\]
does not depend on $N$ at all.  
Therefore, as $N$ increases:
\begin{itemize}
\item the gain term increases sublinearly (diminishing returns),  
\item the harm term remains fixed,  
\item the factor $s/(N C_g\left(1\right))$ shrinks linearly.
\end{itemize}
Consequently, the compute-normalized global improvement saturates and may
eventually decline with increasing $N$.  
Unlike Best-of-$N$, LoTTS avoids global harm structurally, giving it an
advantage that increasing $N$ cannot overcome.

\end{proof}

\myparagraph{Interpretation.}
Increasing $N$ improves the chance of repairing a defect but does not reduce 
global harm.  LoTTS avoids global harm structurally, giving it an advantage
that Best-of-$N$ cannot eliminate.


\begin{tcolorbox}[colback=gray!10, colframe=gray!30, boxrule=0.3pt]
\begin{corollary}[Failure Conditions]\label{cor:failure}
Assume $C_\ell = C_g$.
LoTTS fails to dominate whenever \emph{any one} of the following conditions holds. 
These are exactly the scenarios where the dominance inequality in 
Theorem~\ref{thm:main} reverses:
\begin{itemize}
    \item \textbf{Defects become dense} ($s\!\approx\! M$), eliminating the sparse-defect 
    advantage that normally suppresses global harm.

    \item \textbf{Mask precision becomes extremely low}, causing the false-positive harm term 
    $\rho(\tfrac{1}{\pi}-1)h_\ell\overline{\gamma}$ to dominate and outweigh the repair gains.

    \item \textbf{Local repair probability is significantly smaller than the global one} 
    ($q\!\ll\!\theta_g$), reducing the expected improvement contributed by localized refinement.
\end{itemize}
\end{corollary}
\end{tcolorbox}

\begin{proof}
\myparagraph{Step 1: Express dominance and its negation.}
By Theorem~\ref{thm:main}, LoTTS dominates exactly when
\[
\mathbb{E}[\Delta r]_{\mathrm{local},B}
>
\mathbb{E}[\Delta r]_{\mathrm{global},B}.
\]
Thus LoTTS fails to dominate if and only if
\[
\mathbb{E}[\Delta r]_{\mathrm{local},B}
\le
\mathbb{E}[\Delta r]_{\mathrm{global},B}.
\tag{$\star$}\label{eq:failure_def}
\]

\myparagraph{Step 2: Substitute the expressions from Theorem~\ref{thm:main}.}
Using the explicit formulas for the expected improvements and the assumption
$C_\ell=C_g$, the common positive factor $Bs/C_g$ cancels, yielding:
\[
\rho q\,\overline{\delta}
-
\rho\big(\tfrac{1}{\pi}-1\big)h_\ell\overline{\gamma}
\;\le\;
\theta_g\,\overline{\delta}
-
\big(\tfrac{M}{s}-1\big)h_g\overline{\gamma}.
\]

\myparagraph{Step 3: Group gain terms and harm terms.}
Rearranging the inequality gives:
\[
(\rho q - \theta_g)\,\overline{\delta}
\;\le\;
\rho\big(\tfrac{1}{\pi}-1\big)h_\ell\overline{\gamma}
-
\big(\tfrac{M}{s}-1\big)h_g\overline{\gamma}.
\tag{$\dagger$}\label{eq:failure_raw_corr}
\]
Thus LoTTS fails precisely when the net local repair advantage is no larger than the
local false-positive harm minus the global harm.

\myparagraph{Step 4: Interpret when the inequality can realistically hold.}

\begin{itemize}
\item \emph{Dense defects ($s\approx M$).}  
When $s\approx M$, the factor $\tfrac{M}{s}-1$ becomes small.  
Thus the global harm term
\(
(\tfrac{M}{s}-1)h_g\overline{\gamma}
\)
shrinks, eliminating the sparse-defect advantage.  
Global resampling no longer risks harming many clean patches, so the
right-hand side can easily exceed the left-hand side.

\item \emph{Extremely low precision ($\pi\ll1$).}  
If precision is very low, $(\tfrac{1}{\pi}-1)$ becomes large.  
Then the local harm term
\(
\rho(\tfrac{1}{\pi}-1)h_\ell\overline{\gamma}
\)
may dominate even when $h_\ell$ is small, causing LoTTS to apply many
harmful edits to clean patches and making inequality~\eqref{eq:failure_raw_corr}
hold.

\item \emph{Poor local repair probability ($q\ll\theta_g$).}  
If $q$ is much smaller than $\theta_g$, then $\rho q - \theta_g$ is 
negative or strongly negative, making the left-hand side small even before
considering harm.  
In such cases global resampling offers significantly greater repair gain,
and LoTTS cannot compensate.
\end{itemize}

\myparagraph{Step 5: Conclude the characterization of failure.}
Each of the scenarios above corresponds to a way the right-hand side of
\eqref{eq:failure_raw_corr} can exceed the left-hand side:  
(i) global harm becomes small,  
(ii) local false-positive harm becomes large,  
(iii) local repair gain becomes too weak.  
These are precisely the regimes where
\[
\mathbb{E}[\Delta r]_{\mathrm{local},B}
\le
\mathbb{E}[\Delta r]_{\mathrm{global},B},
\]
i.e.\ LoTTS fails to dominate.

\end{proof}

\myparagraph{Interpretation.}
These corner cases correspond exactly to the settings in which LoTTS's 
structural advantages, locality and sparse harm, no longer hold.

\subsection{Practical Q\&A}
\label{app:TheoreticalAnalysis_4}
Theorem~\ref{thm:main} and Corollary~\ref{cor:dominance} specify when LoTTS
is expected to outperform Global TTS.  
Below we summarize the main implications in a concise question--answer format,
emphasizing why LoTTS is reliable and when it provides a \emph{provably} better 
compute–quality tradeoff.
\smallskip

\myparagraph{Q1:} 
\textbf{Under what conditions is LoTTS \textit{guaranteed} to outperform Global TTS?}

\myparagraph{A1:}
As in Corollary~\ref{cor:lowcost}, LoTTS is guaranteed to be better whenever localized updates are 
significantly cheaper than global ones ($C_\ell \ll C_g$).  
Every unit of compute allocated to LoTTS focuses exclusively on potentially
defective regions, avoiding the large global harm term on $(M-s)$ clean patches.
Thus even \emph{moderate} precision and recall are sufficient for LoTTS to win.

\begin{tcolorbox}[
    colback=violet!8,
    colframe=violet!80!black,
    boxrule=1pt,
    title={\textbf{\textcolor{white}{Takeaway}}},
    coltitle=white,
    fonttitle=\bfseries,
    colbacktitle=violet!80!black,
    enhanced
]
LoTTS becomes strictly more compute-efficient once its \textbf{per-trial} resampling cost is substantially \textbf{lower} than global resampling.
\end{tcolorbox}
\medskip

\myparagraph{Q2:} 
\textbf{In our implementation} ($C_\ell \approx C_g$)\textbf{, why does LoTTS still win?}

\myparagraph{A2:}
As in Corollary~\ref{cor:equal}, when local and global trials cost the same, 
the comparison becomes purely statistical.  In this case, the dominance 
condition in Corollary~\ref{cor:dominance} becomes:
\[
  \rho\!\left[q\,\overline{\delta}
  -\big(\tfrac{1}{\pi}-1\big)h_\ell\,\overline{\gamma}\right]
  \;>\;
  \theta_g\,\overline{\delta}
  -\big(\tfrac{M}{s}-1\big)h_g\,\overline{\gamma}.
\]
LoTTS wins whenever its \emph{recall-weighted} net benefit per selected patch 
exceeds the global benefit minus the massive global harm term.

\begin{tcolorbox}[
    colback=violet!8,
    colframe=violet!80!black,
    boxrule=1pt,
    title={\textbf{\textcolor{white}{Takeaway}}},
    coltitle=white,
    fonttitle=\bfseries,
    colbacktitle=violet!80!black,
    enhanced
]
With equal compute, LoTTS still wins because its updates are localized, whereas the \textbf{global harm term} is too large to yield favorable net benefit.
\end{tcolorbox}
\medskip

\myparagraph{Q3:} 
\textbf{Why does the sparse-defect regime strongly favor LoTTS?}

\myparagraph{A3:}
As shown by the sparse-regime analysis following Corollary~\ref{cor:equal}, 
the sparse-defect regime strongly favors LoTTS because global resampling 
risks harming every clean patch.
If $s \ll M$, the global-harm term:
\[
\big(\tfrac{M}{s}-1\big) h_g \overline{\gamma}
\]
is extremely large. In contrast, LoTTS only touches $\rho s$ defective patches and 
$\rho s(1/\pi-1)$ mistakenly selected ones.

\begin{tcolorbox}[
    colback=violet!8,
    colframe=violet!80!black,
    boxrule=1pt,
    title={\textbf{\textcolor{white}{Takeaway}}},
    coltitle=white,
    fonttitle=\bfseries,
    colbacktitle=violet!80!black,
    enhanced
]
Sparse defects strongly favor LoTTS, because \textbf{local harm does not scale} with the number of clean patches.
\end{tcolorbox}
\medskip

\myparagraph{Q4:} 
\textbf{How accurate must the mask be in avoiding clean patches? (\textit{precision})}

\myparagraph{A4:}
As in Eq.~\eqref{eq:pi_threshold}, LoTTS only requires that each selected patch provides positive expected net gain:
\[
\pi > \bigl(1+\tfrac{q\overline{\delta}}{h_\ell\overline{\gamma}}\bigr)^{\!-1}.
\]

\begin{tcolorbox}[
    colback=violet!8,
    colframe=violet!80!black,
    boxrule=1pt,
    title={\textbf{\textcolor{white}{Takeaway}}},
    coltitle=white,
    fonttitle=\bfseries,
    colbacktitle=violet!80!black,
    enhanced
]
Only \textbf{modest precision} is needed for selected patches to yield positive net gain.
\end{tcolorbox}
\medskip

\myparagraph{Q5:} 
\textbf{How many defective regions must the mask capture? (\textit{recall})}

\myparagraph{A5:}
As in Eq.~\eqref{eq:condition_rho}, when $C_\ell \approx C_g$, LoTTS needs:
\[
\rho \;>\;
\frac{\theta_g\overline{\delta}
-\big(\tfrac{M}{s}-1\big)h_g\overline{\gamma}}
     {q\overline{\delta}
     -(\tfrac{1}{\pi}-1)h_\ell\overline{\gamma}}.
\]

\begin{tcolorbox}[
    colback=violet!8,
    colframe=violet!80!black,
    boxrule=1pt,
    title={\textbf{\textcolor{white}{Takeaway}}},
    coltitle=white,
    fonttitle=\bfseries,
    colbacktitle=violet!80!black,
    enhanced
]
LoTTS \textbf{does not} require perfect defect coverage to outperform global TTS.
\end{tcolorbox}
\medskip

\myparagraph{Q6:} 
\textbf{When might LoTTS fail to outperform global resampling?}

\myparagraph{A6:}
As in Corollary~\ref{cor:failure}, LoTTS may underperform only in regimes where its core advantages disappear:
\begin{itemize}
\item defects are dense ($s\approx M$), eliminating the sparse-defect advantage;
\item precision is too low to avoid excessive harm;
\item local repair probability $q$ is dramatically worse than $\theta_g$.
\end{itemize}

\begin{tcolorbox}[
    colback=violet!8,
    colframe=violet!80!black,
    boxrule=1pt,
    title={\textbf{\textcolor{white}{Takeaway}}},
    coltitle=white,
    fonttitle=\bfseries,
    colbacktitle=violet!80!black,
    enhanced
]
LoTTS underperforms only in \textbf{corner cases} where sparsity, mask precision, or local repair reliability breaks down.
\end{tcolorbox}
\medskip

\myparagraph{Q7:} 
\textbf{What is the practical bottom line?}

\myparagraph{A7:}
In typical real-world scenarios, where defects are sparse and diffusion updates are locally benign, a mask with \emph{modest} precision and recall already satisfies the dominance condition.
Thus LoTTS provides a provably better compute–quality tradeoff than Global TTS,
even when both operate under the same compute budget.

\begin{tcolorbox}[
    colback=violet!8,
    colframe=violet!80!black,
    boxrule=1pt,
    title={\textbf{\textcolor{white}{Takeaway}}},
    coltitle=white,
    fonttitle=\bfseries,
    colbacktitle=violet!80!black,
    enhanced
]
In \textbf{realistic} settings, theoretical guarantees ensure LoTTS provides a superior compute–quality advantage.
\end{tcolorbox}

\end{document}